\documentclass{article}
\usepackage{ijcai26}

\usepackage{times}
\usepackage{soul}
\usepackage{url}
\usepackage[hidelinks]{hyperref}
\usepackage[utf8]{inputenc}
\usepackage[small]{caption}
\usepackage{graphicx}
\usepackage{amsmath}
\usepackage{amsthm}
\usepackage{booktabs}
\usepackage{algorithm}
\usepackage{algorithmic}
\usepackage[switch]{lineno}

\usepackage{graphicx}
\usepackage{hyperref}
\usepackage{url}
\usepackage{tabularx}
\usepackage{adjustbox}
\usepackage{multirow}
\usepackage{wrapfig} 
\usepackage{float}
\usepackage{tablefootnote}
\usepackage{titletoc}
\usepackage{amsmath,amssymb,amsfonts}
\usepackage{booktabs}
\usepackage{color,xcolor}

\definecolor{teal}{RGB}{0,128,128}
\definecolor{crimson}{RGB}{220,20,60}
\definecolor{deep_blue}{RGB}{0,0,255}
\definecolor{green}{RGB}{192,216,174}
\usepackage{stfloats}
\usepackage{colortbl}
\usepackage{subcaption}

\usepackage{tabularx}
\usepackage{array} 

\newcolumntype{Y}{>{\centering\arraybackslash}X}

\newcommand{\algcomment}[1]{\textcolor{gray!80!black}{\small\texttt{\# #1}}}

\newtheorem{theorem}{Theorem}[section]

\newtheorem{definition}[theorem]{Definition}

\nolinenumbers

\title{Limited Reference, Reliable Generation: \\Rule-Guided Tabular Data Generation with Dual-Granularity Filtering}

\author{
    Author Name
    \affiliations
    Affiliation
    \emails
    email@example.com
}

\author{
Mingxuan Jiang$^{1,\dagger}$
\and
Keyang Chen$^{1,\dagger}$
\and
Yongxin Wang$^1$
\and
Yongsheng Zhao$^1$
\and
Ziyue Dai$^1$
\\
Yicun Liu$^1$
\and
Zeping Li$^1$
\and
Qiuyang Zhang$^1$
\And
Hongyi Nie$^{3,*}$
\and
Hongbin Zhu$^2$
\\
Sen Liu$^{2,*}$
\and
Guangnan Ye$^2$
\and
Hongfeng Chai$^2$\\
\affiliations
$^1$School of Computer Science, Fudan University, Shanghai, 200438, China\\
$^2$Institute of Financial Technology, Fudan University, Shanghai, 200438, China\\
$^3$Northwestern Polytechnical University, Xi'an, 710129, China\\
\emails
senliu@fudan.edu.cn, hy\_nie@foxmail.com
}

\begin{document}

\maketitle
\footnotetext[1]{$^\dagger$Mingxuan Jiang and Keyang Chen contributed equally.}
\footnotetext[2]{$^*$Corresponding authors.}

\maketitle

\begin{abstract}
Synthetic tabular data generation is increasingly essential in machine learning, supporting downstream applications when real-world, high-quality tabular data is insufficient.
Existing tabular generation approaches, such as generative adversarial networks (GANs) and fine-tuned Large Language Models (LLMs), typically require sufficient reference data, limiting their effectiveness in domain-specific datasets with scarce records.
While prompt-based LLMs offer flexibility without parameter tuning, they often generate \emph{distributionally drifted data with localized redundancy}, leading to degradation in downstream task performance.
To overcome these issues, we propose \textit{\textbf{ReFine}}, a framework that (i) extracts symbolic \emph{if–then} rules from interpretable models and embeds them into prompts to explicitly guide the generation process toward the domain-specific distribution, and (ii) applies a dual-granularity filtering that mitigates over-sampling patterns while preserving rare but informative samples to reduce localized redundancy. 
Extensive experiments on diverse benchmarks demonstrate that \textit{ReFine} provides robust downstream utility, achieving a top-tier average rank across datasets and data regimes, with an average relative improvement of \textbf{7.48\%} in extreme low-data regimes.
\end{abstract}

\section{Introduction}
Tabular data serves as a foundational modality in machine learning, underpinning critical applications in domains such as healthcare, finance, and scientific research~\cite{benjelloun2020google}.
However, the collection of large-scale, high-quality tabular datasets is often hindered by strict privacy regulations and the prohibitive costs of expert-driven annotation \cite{gdpr,miceli2020between}.
Such limitations severely restrict effective model training in many tabular applications, thereby motivating the use of synthetic data generation~\cite{shankar2024s}.

\begin{figure}[t]
    \centering
    \includegraphics[width=\linewidth]{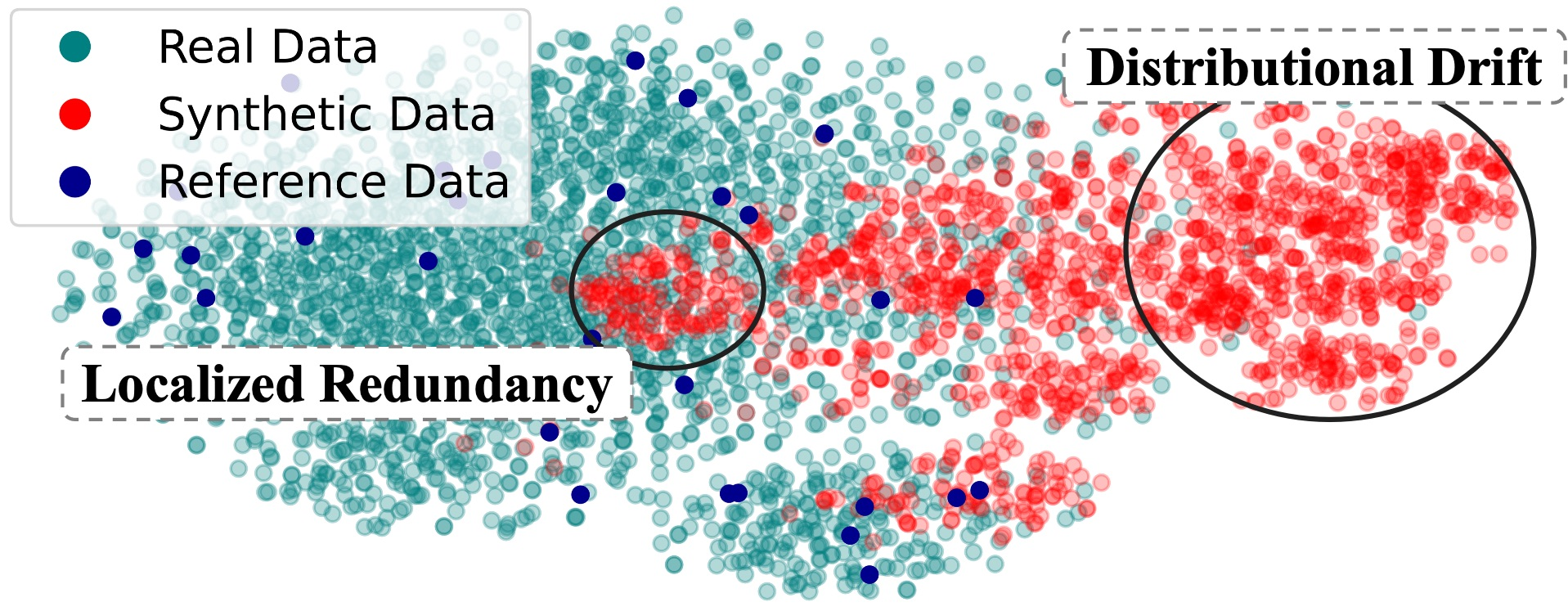}
    \caption{Two key challenges in prompt-based LLM tabular data generation in low-data regimes: (i) \textbf{Distributional Drift}: LLMs often generate biased samples by relying on spurious patterns from pretraining rather than real data distribution. (ii) \textbf{Localized Redundancy}: Synthetic samples tend to concentrate in limited regions of the feature space due to repeated use of identical prompts.}
    \label{motivation}
\end{figure} 

Previous mainstream methods for tabular data generation are based on non-LLM generative models such as VAEs \cite{ctgan}, GANs \cite{zhao2021ctab} and diffusion models \cite{tabddpm}.
Within the LLM-based paradigm, fine-tuning approaches have demonstrated notable performance~\cite{great}.
Both approaches share a fundamental prerequisite: access to sufficient \textit{reference data}, which is used as the foundation for learning underlying distributions.
In practice, this prerequisite is often violated in high-stakes domains, where extremely strict privacy regulations and the rarity of critical events make large-scale, high-quality tabular datasets difficult to obtain~\cite{ji2014differential}.
For example, in rare disease diagnosis, available datasets may contain only a few dozen records, a scenario commonly referred to as a \textit{low-data regime}~\cite{cllm}, which poses severe challenges for data-driven modeling~\cite{wang2024challenges}.
Consequently, in low-data regimes where only a handful of samples are available, existing generative methods fail to capture underlying distributions and thus struggle to produce high-quality synthetic data~\cite{bommareddy2022review,zhang2024metadiff}.
In contrast, prompt-based methods exploit in-context learning to synthesize data without training, offering an alternative for low-data regimes \cite{cllm,epic}.

However, prompt-based methods face two challenges in low-data regimes:
(I) \textbf{Distributional Drift}:
Prompt-based methods often over-rely on LLMs’ pretrained knowledge, which poorly captures dataset-specific distribution \cite{zhong2024opportunities}. Consequently, generated samples tend to follow spurious patterns inherited from pretraining corpora rather than the true patterns of the target dataset \cite{sahoo2024systematic}.
This mismatch leads to synthetic distributions that drifted from the \emph{real data}.
Figure~\ref{motivation} illustrates this issue: when conditioned on a small \emph{reference set} (\textcolor{deep_blue}{blue}), the LLM produces \emph{synthetic data} (\textcolor{crimson}{red}) that drift away from the manifold of \emph{real data} (\textcolor{teal}{cyan}). Many samples populate low-density or unsupported regions, highlighting the LLM’s inability to reconstruct the target joint distribution under data scarcity. 
(II) \textbf{Localized Redundancy}:
Prompt-based generation tends to overproduce high-frequency attribute combinations present in the reference data, while rare but informative patterns are rarely synthesized~\cite{amatriain2024prompt,zevallos2023frequency}. This overconcentration, which we term localized redundancy, is further amplified when identical prompt templates are reused to generate data in multiple batches. As a result, synthetic samples cluster around a few dominant modes~\cite{epic,cllm}, forming high-density regions (\textcolor{crimson}{red}) that contrast sharply with the broader and more balanced distribution of real data (\textcolor{teal}{cyan}), as shown in Figure~\ref{motivation}.

To systematically address the two key challenges of prompt-based tabular generation in low-data regimes, we propose \textbf{\textit{ReFine}} (\underline{R}ule-Guided G\underline{e}neration and Dual-Granularity \underline{F}\underline{i}lt\underline{e}ri\underline{n}g), a framework comprising two components.
To mitigate the distribution of synthetic data often drifted by LLMs in low-data regimes (Challenge I), we introduce \textbf{Rules-Guided Generation}, which extracts symbolic \emph{if–then} formulas from interpretable tree-based models. These association rules are embedded into prompts to guide the LLM toward the distribution of the real data.
To mitigate localized redundancy that persists despite prompt-based generation (Challenge II), we propose \textbf{Dual-Granularity Filtering}. This component informs a two-level filtering process: chunk-level pruning of dominant high-density modes, and instance-level refinement to retain low-density but informative samples. Experiments on diverse benchmarks show that \textit{ReFine} attains the best \emph{Total Rank}.
Our key contributions can be summarized as follows:

\begin{itemize}
\item We identify two key challenges of LLM prompt-based methods in tabular data generation in low-data regimes: (i) distributional drift of the synthetic data; and (ii) localized redundancy in the synthetic data.

\item To address the two challenges, we propose \textbf{\textit{ReFine}}, a framework that constructs \emph{association rules} to guide LLM for tabular data generation, and applies \emph{dual-granularity} filtering to reduce localized redundancy.

\item Experiments show that \textbf{\textit{ReFine}} achieves strong downstream utility, attaining the best \emph{Total Rank} (2.3) and a \textbf{7.48\%} average relative improvement in extreme low-data regime (30--90 samples). Ablations further confirm the complementary benefits of \emph{Rules-Guided Generation} and \emph{Dual-Granularity Filtering}.
\end{itemize}

\section{Related Work}
\subsection{Non-LLM Tabular Generation Method} 

\textbf{Generative Model-based generation.} Many works on tabular data synthesis relied on GANs, diffusion models, and score-based models. 
Among classical methods, CTGAN~\cite{ctgan} extends GANs to handle mixed-type variables but suffers from mode collapse and requires heavy preprocessing. 
TabDDPM~\cite{tabddpm} applies diffusion for continuous attributes, yet its iterative denoising is computationally costly and unstable with scarce samples. 
TABSYN~\cite{tabsyn} integrates diffusion with a VAE backbone to better support mixed-type data, but still depends on sufficient training density to avoid spurious correlations. TabPFN~\cite{hollmann2023tabpfn} is a transformer-based prior-data fitted network pretrained on large collections of synthetic tabular datasets to approximate Bayesian prediction via in-context learning, and it outputs predictions for the full test set in a single forward pass. 
The updated TabPFN study further extends the approach beyond small-scale classification and presents it as a tabular foundation model, additionally demonstrating capabilities such as density estimation and data generation~\cite{hollmann2025tabpfn}.
Overall, while effective in abundant-data settings, these models degrade sharply in low-data regimes.

\noindent\textbf{Constraint-based tabular generation.} Another line of work enforces domain validity through hard logical constraints. 
Early methods embed linear constraints into generative models via \emph{Constraint Layers} \cite{howreal}. 
More recently, \emph{Disjunctive Refinement Layers (DRL)} extend this idea to quantifier-free real linear arithmetic (QFLRA), enabling non-convex and disjunctive feasible regions \cite{drl}. 
While effective for ensuring semantic validity, such methods face two key drawbacks: they rely on exhaustively specified domain rules, which is rarely feasible in practice, and hard constraints restrict the generation space, thereby limiting diversity. 

\subsection{LLM-based Tabular Generation Method} 
LLMs have recently gained attention for tabular data generation, exploiting pretrained knowledge that makes them well-suited for structured data tasks.
Existing approaches fall into two categories: \emph{fine-tuning} and \emph{prompt-based methods}.

\noindent\textbf{Fine-Tuning Methods. } 
Fine-tuning methods adapt LLM parameters to tabular formats and domain constraints.
For instance, GReaT~\cite{great} fine-tunes GPT-2.5 on tabular corpora, while HARMONIC~\cite{wangharmonic} introduces instruction signals derived from nearest-neighbor relationships.
Although effective with abundant reference data, these methods risk severe overfitting when reference data is small, as parameter updates dominate the limited supervision.

\noindent\textbf{Prompt-based Methods. } 
Prompt-based methods leverage the in-context learning ability of LLMs, enabling them to generate tabular data by conditioning on a few labeled examples embedded directly in the prompt\cite{ling2024mallm}. Without modifying model parameters, these methods use prompt design to guide the generation process.
EPIC improves representation balance across classes by formatting grouped data and crafting class-aware prompts~\cite{epic}. CLLM enhances data quality in low-resource scenarios by combining prompt design with a curation step that filters samples based on model confidence and uncertainty estimates~\cite{cllm}. 
However, prompt-based methods struggle to capture logical dependencies and often suffer from distributional imbalance due to repeated use of identical prompts, limiting their effectiveness in low-data regimes.



\section{Methodology}

\begin{figure*}[t]
    \centering
    \includegraphics[width=\linewidth]{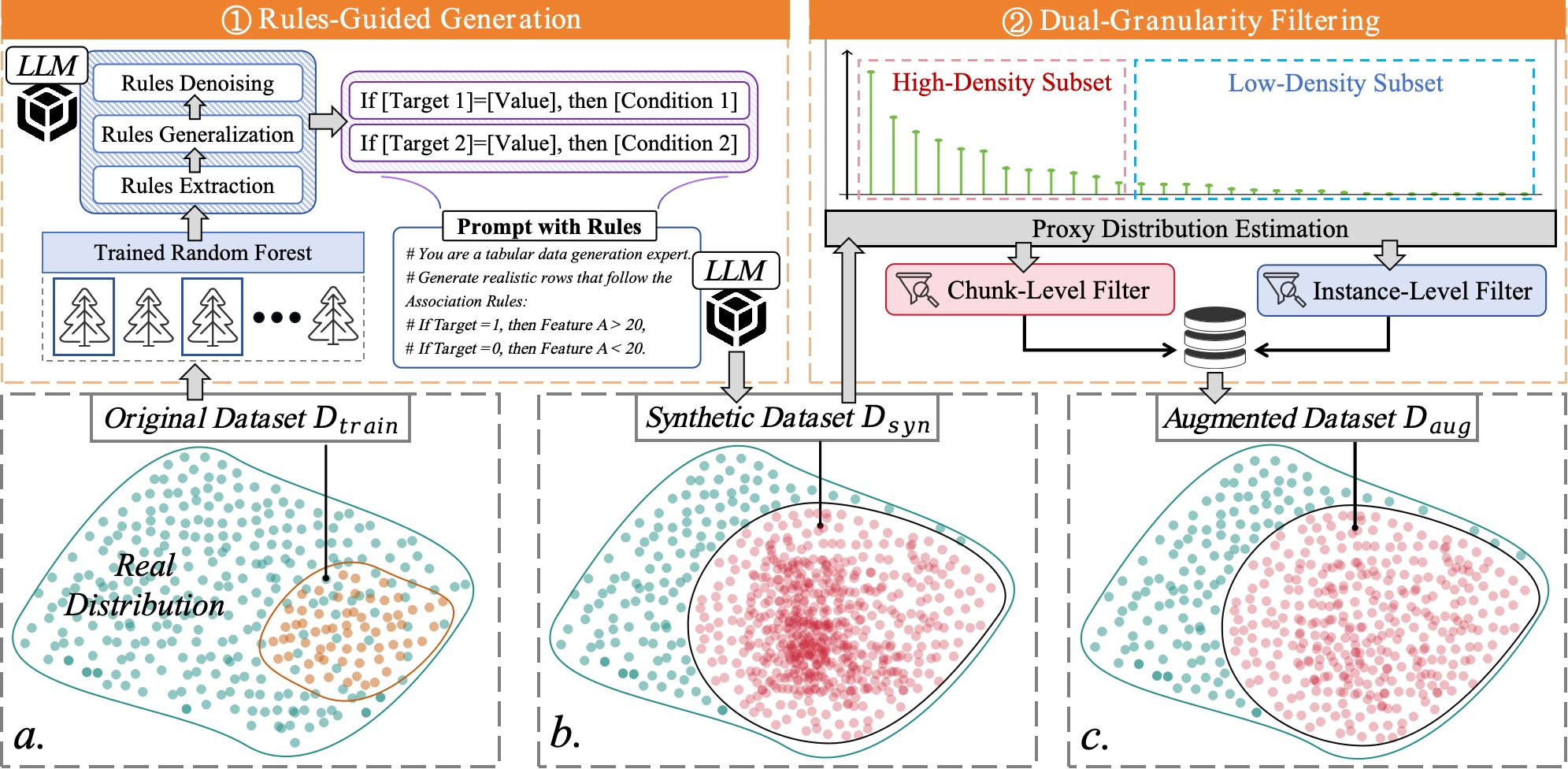}
    \caption{
    Overall framework of \textbf{\textit{ReFine}} (\textit{upper panel}), which consists of two components: \textbf{(1) Rules-Guided Generation}, which leverage rules to guide LLM generation toward the underlying \textit{Target Distribution}; and \textbf{(2) Dual-Granularity Filtering}, which suppresses overrepresented patterns and preserves informative, low-density instances. \textit{(a.)} \(D_{\text{train}}\) sparsely covers the \emph{real distribution} in low-data regimes.
    \textit{(b.)} Using rules extracted from \(D_{\text{train}}\), \textit{Guided Rules} produces \(D_{\text{syn}}\) by steering LLM generation toward domain-consistent patterns.
    \textit{(c.)} \textit{Dual-Granularity Filtering} refines \(D_{\text{syn}}\) into \(D_{\text{aug}}\) by pruning overrepresented patterns and retaining informative samples.}
    \label{pipeline}
\end{figure*}

Prompt-based tabular generation in low-data regimes suffers from two key issues: \textbf{\textit{Distributional Drift}} and \textbf{\textit{Localized Redundancy}}. 
To tackle these challenges, we unify our intuition and methodological pipeline in Figure~\ref{pipeline}.
Here, the \textit{Real Distribution} (\textcolor{teal}{cyan}) denotes the target data manifold, while the \textit{Original Small Dataset} (\textcolor{orange}{orange}) provides limited supervision.
To mitigate \textbf{\textit{Distributional Drift}}, we propose \textbf{(1) Rules-Guided Generation}, which extracts \textit{Association Rules} (\textcolor{red}{red}) from tree-based models to explicitly capture key feature dependencies. These rules are embedded into prompts, guiding the LLM toward generation that better align with the real distribution.
However, static prompting still induces \textbf{\textit{Localized Redundancy}}. To address this, we introduce \textbf{(2) Dual-Granularity Filtering}, which prunes dominant high-density chunks while retaining informative but rare samples at the instance level.
Together, these two components expand the effective generation space and enhance the fidelity and diversity of the augmented dataset for downstream learning. In what follows, we first formally define the tabular generation task under low-data regimes. Then, we describe our two components for addressing the two primary challenges.

\subsection{Problem Setup}
\begin{definition}
\textbf{Tabular Data Generation in Low-data Regimes.}
Let \(X \subseteq \mathbb{R}^{d}\) be the \(d\)-dimensional feature space and \(Y\) the label space (discrete for classification, real-valued for regression).  
Let \(D_{\text{train}}=\{(x_i,y_i)\}_{i=1}^{N}\) be a small labeled dataset independently and identically drawn from the unknown real distribution \(p_R(X,Y)\), where \(x_i\in X\) and \(y_i\in Y\), with \(N\) limited to only a few examples.
The task defines a generation function \(\mathcal{G}\) that maps \(D_{\text{train}}\) to a synthetic dataset:
\[
D_{\text{syn}} = \mathcal{G}(D_{\text{train}}) = \{(\tilde{x}_j, \tilde{y}_j)\}_{j=1}^{M}.
\]
where \((\tilde{x}_j,\tilde{y}_j)\) is a synthetic sample, and \(M\) is the number of synthetic samples, typically satisfying \(M \gg N\).  
The task goal is to generate \(D_{\text{syn}}\) such that a model trained on it achieves strong predictive performance when evaluated on a held-out real test set \(D_{\text{test}} \sim p_R\). 
\end{definition}

\subsection{Component I: Rules-Guided Generation}
We mitigate \textit{Distributional Drift} with \textbf{Rules-Guided Generation}, which extracts association rules from limited reference data $D_{\mathrm{train}}$ as structural priors to guide generation.
Unlike other modalities, tabular data often lack inherent structure and contain many irrelevant features, making it difficult for an LLM to capture feature dependencies from limited data \cite{Fang2024}. 
To provide informative guidance, we define \textit{association rules} as \textit{if–then} formulas extracted from decision paths in tree-based models, capturing supervised dependencies between features and labels. Unlike classical association rule mining based on co-occurrence statistics, our rules reflect predictive patterns learned via training. 
We apply the following steps to generate the synthetic dataset $D_{\mathrm{syn}}$.

\textbf{1) Rules Extraction From Top-Performing Trees.} 
We train a Random Forest (RF) on $D_{\mathrm{train}}$ and rank trees based on in-sample accuracy, selecting the top-$k$ trees (e.g., $k$=3). Each selected tree provides a set of \textit{if–else} decision rules~\cite{azad2025novel,khan2024review}, yielding diverse local patterns under low-data regimes.

\textbf{2) Rules Generalization and Denoising.} 
\label{sec:formula-extraction}
To construct reliable \textit{association rules}, we apply:

\textit{(a) Rule Generalization.}  
We prune and consolidate decision paths across top-performing trees, retaining only their core (i.e., high-support, low-depth) branches that reflect stable feature–label dependencies. This produces \textit{if–then} association rules that generalize beyond individual training instances. Treated as conditional templates with the label as premise, these rules enable inverse reasoning and steer generation toward broader yet distributionally consistent samples.

\textit{(b) Rule Denoising.} 
\label{sec:rule-denoising}
To reduce variance introduced by symbolic extraction and decoding noise~\cite{he2024advancing}, we apply self-consistency techniques from reasoning tasks~\cite{wangself}: we perform multiple generations with different seeds and retain only the most frequently occurring rules. This ensures the resulting rule set is stable and reliable. We provide the full prompt templates used in Appendix~\ref{app:prompts}.

\textbf{3) Tabular Data Generation via Rules-Guided Prompt.} 
Association rules obtained in Component I are converted into structured prompts encoding their \emph{if–then} formulas. These prompts constrain the LLM’s decoding space to enforce meaningful distributional patterns, yielding the synthetic dataset $D_{\text{syn}}$ that offers distribution-consistent diversity.

\subsection{Component II: Dual-Granularity Filtering}
We propose \textbf{Dual-Granularity Filtering} to mitigate \emph{localized redundancy} and filter low-quality samples in $\mathcal{D}_{\mathrm{syn}}$ using learning-dynamics signals from a reference model $\mathcal{M}$ trained only on $\mathcal{D}_{\mathrm{train}}$. Here $\mathcal{M}$ follows the CLLM curator (same architecture and training recipe) and is used solely as a signal extractor rather than a new model contribution.
The procedure has two stages: (i) \emph{chunk-level filtering} prunes over-represented synthetic modes in high-density regions, and (ii) \emph{instance-level filtering} removes unreliable samples in low-density regions. To make this filtering effective, we first estimate where and how redundancy occurs in \(D_{\text{syn}}\).

\textbf{1) Proxy-Based Density Estimation. }
We assign each synthetic sample $x_i\in\mathcal{D}_{\mathrm{syn}}$ to its nearest real \emph{anchor} $s_j\in\mathcal{D}_{\mathrm{train}}$ under the mixed-type distance $\mathrm{DCR}(\cdot,\cdot)$ \cite{great}. Let $\{s_j\}_{j=1}^{N}$ denote anchors with $N=|\mathcal{D}_{\mathrm{train}}|$. This induces a proxy density over anchors,
\begin{equation}
\begin{aligned}
p_j
&= \frac{1}{\lvert \mathcal{D}_{\mathrm{syn}} \rvert}
   \Bigl\lvert \Bigl\{\, x \in \mathcal{D}_{\mathrm{syn}}
   \;\Big|\;
   j = \arg\min_{k \in [N]} \mathrm{DCR}(x, s_k) \Bigr\} \Bigr\rvert, \\
&\hspace{2.1em} j \in [N].
\end{aligned}
\end{equation}
where $p_j$ is the fraction of synthetic samples mapped to anchor $s_j$ and $\sum_j p_j=1$. We summarize concentration with the Gini coefficient $G(p)$: larger $G(p)$ indicates that synthetic mass concentrates on fewer anchors. In practice, we use the anchor assignments (and the resulting anchor-level densities $p_j$) to identify dense vs.\ sparse regions; $G(p)$ is used as a global scalar that modulates filtering strength.

\textbf{2) Chunk-Level Filtering.}
\label{sec:high/low}
To mitigate localized redundancy in high-density subset, we first group synthetic samples into chunks of size $S$, denoted as $\mathcal{C}_r$. Each chunk is evaluated by aggregating per-sample signals from the reference model $\mathcal{M}_{\text{cur}}$  (i.e., $\mathcal{M}_{\mathrm{cur}} \equiv \mathcal{M}$).

To capture learning-dynamics signals from the $\mathcal{M}_{\text{cur}}$, we snapshot its parameters at $T$ checkpoints during a \emph{single} training run on $\mathcal{D}_{\mathrm{train}}$. For each synthetic sample, we then summarize how the $\mathcal{M}_{\text{cur}}$'s predictions evolve over these checkpoints, i.e., a prediction trajectory. In classification, this trajectory is the sequence of gold-label probabilities; in regression, it is the sequence of scalar predictions, from which we derive trajectory-based error and dispersion measures. We treat these task-specific trajectory statistics as per-sample reliability signals and aggregate them within each chunk to obtain a chunk score, which computed as:
\begin{equation}
\mathrm{Score}(\mathcal{C}_r) = \frac{1}{|\mathcal{C}_r|} \sum_{(x_i, y_i) \in \mathcal{C}_r} \frac{1}{T} \sum_{t=1}^T \mathbb{I}_{\mathrm{corr}}(x_i, y_i; \mathcal{M}_{cur,t}),
\end{equation}
where $\mathbb{I}_{\mathrm{corr}}$ evaluates sample correctness as above.

We rank all chunks by their score and retain a fraction determined by the $\mathcal{D}_{\mathrm{train}}$'s concentration level. Specifically, we compute the retention ratio $\mathrm{ratio}_{\mathrm{ret}}$ as:
\begin{equation}
\mathrm{ratio}_{\mathrm{ret}} = \mathrm{clip}\left( A \cdot \ln \left( \max(G(p), \varepsilon) \right) + B,\ 0,\ 1 \right),
\label{eq:ratio2}
\end{equation}
where $\varepsilon$ is a small constant for numerical stability, and $\mathrm{clip}(\cdot, 0, 1)$ restricts the output to a valid fraction. We retain the top $\lfloor \mathrm{ratio}_{\mathrm{ret}} \cdot N \rfloor$ chunks and keep all samples within these retained chunks.

\textbf{3) Instance-Level Filtering. } In low-density subset, we identify and discard unreliable synthetic samples by evaluating their prediction consistency across training. Specifically, we collect the output trajectories of a reference model $\mathcal{M}_{\text{cur}}$ at $T$ checkpoints, and extract two reliability scores: (i) \textbf{Confidence}, measuring how consistently the model supports the gold label; (ii) \textbf{Uncertainty}, capturing the variability of predictions across checkpoints. For classification, confidence is the average predicted probability assigned to the gold label, and uncertainty is its standard deviation. For regression, we use the mean prediction error and output variance, respectively. Thresholds are adaptively set based on the mean and variance of each score over the sparse region, and scaled by the global redundancy score $G(p)$. Only samples with high confidence and low uncertainty are retained.

After applying both filters, we merge the retained samples to form the augmented synthetic set $\mathcal{D}_{\mathrm{aug}}$. 
We then select the chunk size $S$ by training a separate reference model on $\mathcal{D}_{\mathrm{aug}}$ for each candidate $S$ and choosing the one that maximizes performance on $\mathcal{D}_{\mathrm{train}}$. 
Full details of Component II are provided in Appendix~\ref{app:com2}.

\section{Experiment}

\begin{table*}[t]
\scriptsize
\centering
\begin{adjustbox}{max width=\textwidth}
\begin{tabularx}{\textwidth}{c|Y|Y|Y|Y|Y|Y|Y|Y|Y}
\toprule
\multicolumn{2}{c|}{\textbf{Original Data}} &
\multicolumn{3}{c|}{\textbf{Non-LLM Methods}} &
\multicolumn{5}{c}{\textbf{LLM-Based Methods}} \\
\cmidrule(lr){1-2} \cmidrule(lr){3-5} \cmidrule(lr){6-10}
Datasets & Real &\textsc{DRL} & \textsc{TabSyn} &\textsc{Tabpfn(Gen)} &\textsc{EPIC} & \textsc{CLLM} & I \(\setminus\) II & II \(\setminus\) I & I+II \\
\midrule
\cellcolor{blue!10}Disease ($N$=30) & 93.68$\pm$1.4 & 39.49$\pm$0.31 & 54.79$\pm$2.7 & 61.38$\pm$0.45 & 32.01$\pm$8.2 & 61.89$\pm$2.1 & 59.61$\pm$1.5 & 62.07$\pm$1.7 & \textbf{70.22$\pm$0.83} \\
\cellcolor{blue!10}Game ($N$=30)    & 86.0$\pm$0.73  & 33.54$\pm$6.33 & 44.13$\pm$1.4 & \textbf{70.43$\pm$1.81} & 13.93$\pm$1.2 & 54.12$\pm$2.7 & 56.44$\pm$2.2 & 45.0$\pm$1.3 & 59.13$\pm$2.8 \\
\cellcolor{blue!10}GPA ($N$=30)     & 47.29$\pm$0.90 & 14.46$\pm$1.54 & 19.22$\pm$2.5 & 41.40$\pm$0.92 & 32.03$\pm$2.8 & 40.17$\pm$0.46 & 41.21$\pm$0.97 & 35.88$\pm$1.7 & \textbf{43.14$\pm$0.88} \\
\cellcolor{blue!10}Student ($N$=30) & 0.67$\pm$0.07  & -0.01$\pm$0.01 & 0.14$\pm$0.04 & 0.37$\pm$0.01 & 0.35$\pm$0.05 & -0.11$\pm$0.12 & 0.37$\pm$0.02 & 0.08$\pm$0.06 & \textbf{0.38$\pm$0.02} \\
\cellcolor{red!10}Adult ($N$=30)   & 76.92$\pm$0.68 & 46.59$\pm$2.44 & 62.83$\pm$3.2 & \underline{63.88$\pm$5.35} & 61.94$\pm$3.3 & 73.11$\pm$0.65 & 73.04$\pm$0.59 & 73.15$\pm$0.18 & \textbf{73.91$\pm$0.37} \\
\cellcolor{red!10}Heart ($N$=30)  & 86.71$\pm$1.7  & 38.66$\pm$2.29 & 79.40$\pm$1.3 & \underline{61.42$\pm$2.00} & \textbf{81.20$\pm$1.3} & 80.47$\pm$0.84 & 75.17$\pm$4.7 & 80.00$\pm$0.24 & 80.14$\pm$0.33 \\
\midrule
\textbf{Avg Rank} & - & 7.5 & 5.8 &	\underline{3.6} & 5.7 &	4.0 & 3.9	&4.0 &\textbf{1.5}\\
\midrule
\cellcolor{blue!10}Disease ($N$=60) & 93.68$\pm$1.4 & 38.25$\pm$0.22 & 65.34$\pm$9.3 & 71.88$\pm$0.62 & 44.05$\pm$7.8 & 66.86$\pm$1.1 & \textbf{75.65$\pm$2.5} & 62.58$\pm$2.2 & 72.41$\pm$0.77 \\
\cellcolor{blue!10}Game ($N$=60)    & 86.0$\pm$0.73  & 30.10$\pm$1.07 & 61.10$\pm$1.1 & \textbf{78.28$\pm$0.99} & 13.16$\pm$0 & 67.54$\pm$1.0 & 61.39$\pm$2.6 & 59.81$\pm$1.1 & 70.87$\pm$0.42 \\
\cellcolor{blue!10}GPA ($N$=60)     & 47.29$\pm$0.90 & 15.64$\pm$1.88 & 28.48$\pm$1.8 & 47.81$\pm$0.61 & 32.19$\pm$3.5 & 33.17$\pm$1.1 & 44.34$\pm$0.91 & 21.57$\pm$1.8 & \textbf{44.58$\pm$0.56} \\
\cellcolor{blue!10}Student ($N$=60) & 0.67$\pm$0.07  & -0.04$\pm$0.08 & -0.14$\pm$0.14 & \textbf{0.46$\pm$0.01} & -0.63$\pm$0.19 & -0.48$\pm$0.19 & 0.27$\pm$0.03 & -0.15$\pm$0.04 & 0.34$\pm$0.07 \\
\cellcolor{red!10}Adult ($N$=60)    & 76.92$\pm$0.68 & 49.02$\pm$1.34 & 63.58$\pm$1.6 & \underline{64.69$\pm$3.40} & 62.78$\pm$1.5 & \textbf{73.94$\pm$0.54} & 73.28$\pm$0.85 & 72.85$\pm$0.50 & 71.48$\pm$0.60 \\
\cellcolor{red!10}Heart ($N$=60)    & 86.71$\pm$1.7  & 38.14$\pm$2.85 & 81.35$\pm$1.2 & \underline{63.98$\pm$1.78} & 77.55$\pm$1.6 & 80.33$\pm$0.85 & \textbf{81.37$\pm$0.96} & 78.15$\pm$0.53 & 80.36$\pm$0.03 \\
\midrule
\textbf{Avg Rank} & - & 7.2 & 4.8 & 3.2 & 6.8 & 3.8 & \textbf{2.3} & 5.5 & \textbf{2.3} \\
\midrule
\cellcolor{blue!10}Disease ($N$=120) & 93.68$\pm$1.4 & 73.92$\pm$3.30 & 62.37$\pm$1.25 & 78.85$\pm$0.40 & 55.37$\pm$10.36 & 78.30$\pm$1.44 & \underline{78.97$\pm$3.13} & 61.82$\pm$2.57 & \textbf{81.96$\pm$0.64} \\
\cellcolor{blue!10}Game ($N$=120)    & 86.0$\pm$0.73  & 31.04$\pm$2.48 & \underline{63.50$\pm$0.78} & \textbf{73.93$\pm$0.41} & 48.99$\pm$2.41 & 61.32$\pm$1.16 & 45.48$\pm$1.62 & 54.76$\pm$2.01 & 61.67$\pm$1.40 \\
\cellcolor{blue!10}GPA ($N$=120)     & 47.29$\pm$0.90 & 10.63$\pm$1.75 & 38.57$\pm$2.30 & 44.78$\pm$0.71 & 46.61$\pm$3.7 & 45.20$\pm$2.1 & 40.42$\pm$1.02 & \underline{46.95$\pm$0.5} & \textbf{47.77$\pm$0.81} \\
\cellcolor{blue!10}Student ($N$=120) & 0.67$\pm$0.07  & 0.01$\pm$0.03 & \underline{0.41$\pm$0.03} & \textbf{0.47$\pm$0.03} & 0.29$\pm$0.07 & 0.22$\pm$0.03 & 0.21$\pm$0.04 & 0.17$\pm$0.06 & 0.29$\pm$0.04 \\
\cellcolor{red!10}Adult ($N$=120)    & 76.92$\pm$0.68 & 39.22$\pm$6.96 & 68.87$\pm$0.41 & \underline{68.42$\pm$1.04} & 68.35$\pm$2.10 & 71.73$\pm$0.82 & \textbf{73.36$\pm$0.67} & 73.19$\pm$0.65 & 72.93$\pm$0.74 \\
\cellcolor{red!10}Heart ($N$=120)    & 86.71$\pm$1.7  & 38.29$\pm$0.31 & 83.80$\pm$0.51 & \underline{69.27$\pm$1.19} & 71.33$\pm$6.33 & 75.24$\pm$0.72 & \textbf{84.97$\pm$1.08} & 75.02$\pm$0.97 & 75.93$\pm$0.84 \\
\midrule
\textbf{Avg Rank} & - & 7.5 & 4.0 & \underline{3.8} & 5.6 & 4.2 & \underline{3.8} & 4.7 & \textbf{2.4} \\
\midrule
\cellcolor{blue!10}Disease ($N$=200) & 93.68$\pm$1.4 & 51.08$\pm$1.71 & 64.53$\pm$1.49 & \textbf{78.07$\pm$0.84} & 55.30$\pm$6.14 & 73.72$\pm$1.98 & 68.58$\pm$2.13 & 71.96$\pm$2.86 & \underline{74.25$\pm$1.38} \\
\cellcolor{blue!10}Game ($N$=200)    & 86.0$\pm$0.73  & 32.98$\pm$2.47 & \textbf{77.92$\pm$0.69} & \underline{69.18$\pm$2.03} & 55.19$\pm$6.76 & 51.19$\pm$2.06 & 44.68$\pm$3.78 & 61.56$\pm$0.68 & 61.98$\pm$1.52 \\
\cellcolor{blue!10}GPA ($N$=200)     & 47.29$\pm$0.90 & 14.53$\pm$1.94 & 41.59$\pm$0.86 & \textbf{46.26$\pm$0.9} & 40.88$\pm$0.94 & 36.40$\pm$3.18 & 43.92$\pm$1.42 & 38.73$\pm$0.00 & \underline{44.68$\pm$0.51} \\
\cellcolor{blue!10}Student ($N$=200) & 0.67$\pm$0.07  & -0.02$\pm$0.02 & 0.39$\pm$0.11 & 0.50$\pm$0.01 & \underline{0.51$\pm$0.04} & 0.24$\pm$0.06 & 0.50$\pm$0.08 & 0.20$\pm$0.07 & \textbf{0.52$\pm$0.02} \\
\cellcolor{red!10}Adult ($N$=200)    & 76.92$\pm$0.68 & 46.19$\pm$6.68 & 69.43$\pm$0.71 & 69.21$\pm$0.66 & 70.95$\pm$1.34 & 73.40$\pm$0.75 & 72.42$\pm$0.88 & \underline{73.66$\pm$0.71} & \textbf{74.50$\pm$0.30} \\
\cellcolor{red!10}Heart ($N$=200)    & 86.71$\pm$1.7  & 47.80$\pm$6.27 & 74.33$\pm$2.60 & 66.09$\pm$1.51 & 74.74$\pm$2.10 & 76.17$\pm$2.01 & \underline{79.95$\pm$0.11} & \textbf{80.11$\pm$0.91} & 77.84$\pm$1.01 \\
\midrule
\textbf{Avg Rank} & - & 8.0 & 4.7 & \underline{3.6} & 4.8 & 4.8 & 4.1 & 4.0 & \textbf{2.0} \\
\midrule
\cellcolor{blue!10}Disease ($N$=500) & 93.68$\pm$1.4 & \underline{75.27$\pm$2.39} & 69.93$\pm$1.41 & \textbf{78.52$\pm$0.26} & 65.26$\pm$3.15 & 67.56$\pm$2.83 & 61.41$\pm$3.90 & 68.55$\pm$2.44 & 72.48$\pm$1.97 \\
\cellcolor{blue!10}Game ($N$=500)    & 86.0$\pm$0.73  & 25.69$\pm$3.03 & \textbf{80.33$\pm$1.26} & \underline{79.60$\pm$0.89} & 61.38$\pm$5.18 & 53.88$\pm$1.77 & 69.37$\pm$2.33 & 65.33$\pm$1.19 & 75.95$\pm$1.57 \\
\cellcolor{blue!10}GPA ($N$=500)     & 47.29$\pm$0.90 & 7.57$\pm$1.44 & 40.17$\pm$1.20 & \textbf{43.13$\pm$0.74} & 36.35$\pm$0.95 & 30.14$\pm$2.30 & 37.90$\pm$1.94 & 35.93$\pm$0.98 & \underline{40.30$\pm$0.65} \\
\cellcolor{blue!10}Student ($N$=500) & 0.67$\pm$0.07  & -0.04$\pm$0.03 & \textbf{0.57$\pm$0.01} & 0.56$\pm$0.01 & \underline{0.54$\pm$0.02} & -0.28$\pm$0.08 & 0.46$\pm$0.04 & 0.03$\pm$0.22 & 0.52$\pm$0.01 \\
\cellcolor{red!10}Adult ($N$=500)    & 76.92$\pm$0.68 & 49.33$\pm$3.06 & 70.11$\pm$0.78 & 71.02$\pm$0.92 & 68.06$\pm$1.37 & \underline{73.33$\pm$0.97} & \textbf{74.11$\pm$0.32} & 71.15$\pm$0.77 & 70.11$\pm$0.31 \\
\cellcolor{red!10}Heart ($N$=500)    & 86.71$\pm$1.7  & 27.73$\pm$0.13 & 77.26$\pm$2.46 & 63.92$\pm$2.90 & \textbf{81.71$\pm$1.81} & 76.01$\pm$0.21 & 75.20$\pm$0.13 & 74.20$\pm$0.90 & \underline{80.31$\pm$1.17} \\
\midrule
\textbf{Avg Rank} & - & 6.8 & \underline{2.9} & \textbf{2.8} & 4.8 & 5.7 & 4.5 & 5.2 & 3.3 \\
\midrule
\textbf{Total Rank} & - & 7.4 & 4.3 &  \underline{3.4} & 5.5 & 4.5 & 3.7  &4.7 & \textbf{2.3}\\

\bottomrule
\end{tabularx}
\end{adjustbox}
\caption{Main results on benchmark datasets. We report \textbf{$F_{1}$ score} for classification tasks and \textbf{R\textsuperscript{2}} for regression tasks. \colorbox{blue!10}{Unseen datasets} (i.e., not included in the LLM's memory for LLM-based Generator) are highlighted in blue, and \colorbox{red!10}{seen datasets} are highlighted in red. \textbf{I$\backslash$II} denotes \textit{Rules-Guided Generation} without \textit{Dual-Granularity Filtering}, and \textbf{II$\backslash$I} denotes \textit{Dual-Granularity Filtering} without \textit{Rules-Guided Generation}. The best result in each row is shown in \textbf{bold}, and the second-best result is \underline{underlined}.}
\label{tab:main}
\end{table*}

In this section, we evaluate \textbf{\textit{ReFine}} on downstream tasks and analyze how its two components address the key challenges:

\textbf{1) Low-data Effectiveness:} \emph{How well does \textit{ReFine} improve downstream performance in low-data regimes?}
Section~\ref{sec:main_results} presents comprehensive results across low-data regimes, demonstrating strong downstream utility and competitive average ranks.

\textbf{2) Mitigating Distributional Drift (Challenge I)}: \emph{How do rules enhance data quality?} Section~\ref{com1} shows that rule-guided generation achieves rule-compliance rates consistent with real data, thereby aligning synthetic samples more faithfully with the target distribution.  

\textbf{3) Reducing Localized Redundancy (Challenge II)}: \emph{How does filtering balance the distribution?} Section~\ref{sec:com2} shows that a reliable redundancy metric together with dual-granularity filtering prevents overconcentration, leading to more useful synthetic datasets.


\subsection{Experiment Settings}
\noindent\textbf{Datasets:} To avoid potential \textit{data contamination}, where strong performance on popular benchmarks such as \textit{Adult}, \textit{Heart}, and \textit{Housing} may arise from LLM memorization rather than genuine generalization~\cite{xu2024benchmark}, we use \textit{tabmemcheck}~\cite{bordt2024colm}, a recently proposed tool for detecting memorization in tabular data. Specifically, we apply two of its tests: the \textit{Feature Names} and \textit{Header Test} to eight candidate datasets. Based on these tests, only \textit{adult} and \textit{heart} are classified as \colorbox{red!10}{seen} datasets; the remaining four show no evidence of memorization and are classified as \colorbox{blue!10}{unseen}; full results are in Appendix~\ref{appendix:dataset}.

\noindent\textbf{Baselines:} 
We compare our method against a diverse set of competitive baselines:

\noindent (1) Generative models for tabular data:
(i) \textsc{TabSyn}~\cite{tabsyn}, a score-based generative model that achieves strong performance in sufficient-data settings; and (ii) \textsc{TabPFN}~\cite{hollmann2025tabpfn}, a transformer-based tabular foundation model pretrained on millions of synthetic datasets that delivers accurate predictions and supports generative capabilities on small to medium tabular datasets.

\noindent (2) LLM-based baselines:
(i) \textsc{EPIC}~\cite{epic}, a prompting-based method that automates dataset construction through instruction-driven generation; and (ii) \textsc{CLLM}~\cite{cllm}, which enhances LLM-generated data via instance-level curation.

\noindent (3) Constraint-based baseline:
\textsc{DRL}~\cite{drl}, a constraint-driven generator that synthesizes data by solving logical constraints. Note that \textsc{DRL} builds on a GAN-based formulation, which is subsumed by and outperformed by more recent generative methods; we include it to represent constraint-guided synthetic generation.

\noindent\textbf{Experimental Setup.}
We evaluate all methods under low-data regimes with
\(N \in \{30, 60, 90, 120, 160, 200, 300, 500\}\).
For brevity, the main paper reports results for five representative settings, with complete results provided in Appendix~\ref{appendix:setup}.

For each dataset and value of \(N\), we create three random train--test splits using different seeds.
Following the Machine Learning Efficiency (MLE) protocol, each method generates 2{,}000 synthetic samples per split, from which 1{,}000 instances are randomly subsampled ten times to reduce variance.
Downstream performance is evaluated using XGBoost~\cite{chen2016xgboost}, and reported as average \textbf{F1 score} (classification) or \textbf{R\textsuperscript{2}} (regression).

Additional evaluations with TabPFN are deferred to Appendix~\ref{app:exp_main}, the privacy evaluation is reported in Appendix~\ref{app:dcr}, and computational cost details are provided in Appendix~\ref{app:cost}.

\begin{figure}[t]
    \centering
    \includegraphics[width=0.5\textwidth]{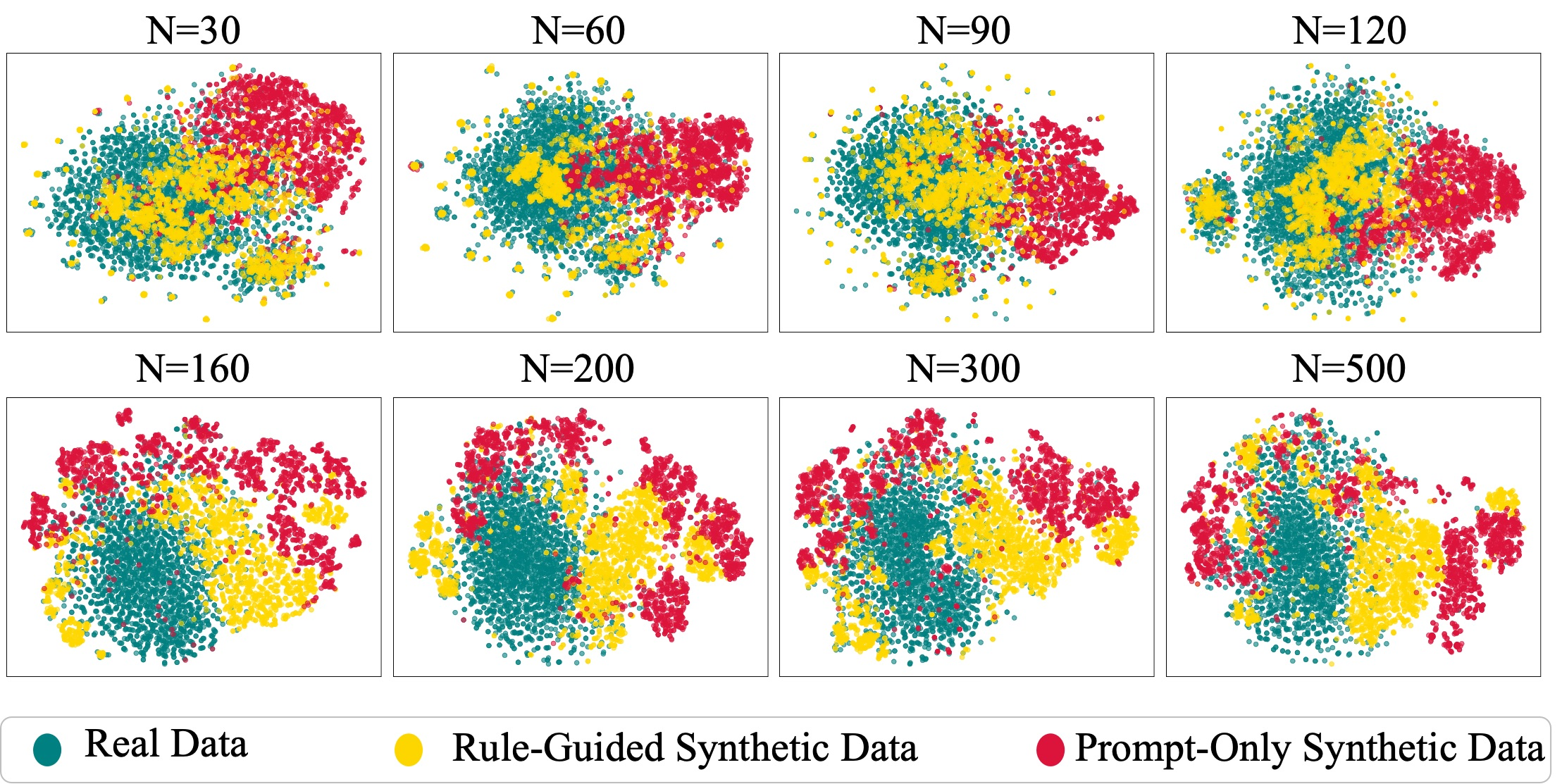}
    \caption{t-SNE visualization of \emph{Real Data}, \emph{Rule-Guided} synthetic data, and \emph{Prompt-only} synthetic data on the \textit{Disease} dataset across varying training sizes ($N{=}30/60/90/120/160/200/300/500$).}
    \label{fig:tsne}
\end{figure} 

\subsection{Low-data effectiveness}
\label{sec:main_results}
\emph{How well does \textit{ReFine} improve downstream performance in low-data regimes?} Table~\ref{tab:main} shows downstream performance under various low-data conditions. \textbf{\textit{ReFine}} (I+II) achieves the best average rank across datasets for ($N \in {30, 60, 120, 200}$) under the XGBoost evaluator, and remains top-tier at ($N = 500$). When aggregating across all regimes, \textit{ReFine} achieves the best Total Rank (\textbf{2.3}), reflecting its consistent performance across datasets and training sizes. This is further supported by results from the TabPFN evaluator in 
Table~\ref{app:tabpfn} (Appendix \ref{app:exp_main}), where \textit{ReFine} also achieves the best total rank and maintains competitiveness as (N) increases. Notably, in the most data-scarce regime ($N \in [30, 90]$), \textit{ReFine} delivers an average relative gain of \textbf{7.48\%} over the strongest baseline (\textsc{TabPFN(GEN)}), demonstrating the effectiveness of rule grounding and redundancy-aware filtering in limited-data scenarios.

Across all reported \(N\), the full framework consistently improves upon its single-component variants in average rank, indicating that the two components provide complementary benefits: \emph{Rules-Guided Generation} promotes distributional faithfulness, whereas \emph{Dual-Granularity Filtering} suppresses localized redundancy and oversampling artifacts.
As \(N\) grows, stronger non-LLM generators become increasingly competitive: \textsc{TabSyn} and \textsc{TabPFN(GEN)} substantially narrow the gap and can dominate in the high-\(N\) regime, consistent with their improved distribution modeling when sufficient reference data is available.
By contrast, \textsc{DRL} remains comparatively weak and shows limited scaling with \(N\), suggesting that hard constraint satisfaction can reduce sample diversity and hinder effective exploitation of additional data.

The seen/unseen split further exposes a robustness gap among several LLM-based baselines: while they achieve strong performance on \colorbox{red!10}{seen datasets}, their utility drops substantially on \colorbox{blue!10}{unseen datasets}.
This behavior is consistent with the \emph{data contamination} concern raised by our dataset selection protocol, in which part of the observed gains may be attributable to pretraining-time overlap or memorization rather than distributional generalization.
In contrast, \textbf{\textit{ReFine}} exhibits markedly smaller degradation from seen to unseen datasets, indicating improved robustness to distribution shift.

\begin{table}[t]
\centering
\renewcommand{\arraystretch}{1.15}
\begin{adjustbox}{max width=\columnwidth}
\begin{tabular}{lcc|cc|cc}
\toprule
 & \multicolumn{1}{c}{\textbf{$D_{\text{test}}$}} 
 & \multicolumn{1}{c|}{\textbf{$D_{\text{train}}$}} 
 & \multicolumn{2}{c|}{\textbf{Rule Guided $D_{\text{syn}}$}} 
 & \multicolumn{2}{c}{\textbf{Prompt-only $D_{\text{syn}}$}} \\
\cmidrule(lr){2-2}\cmidrule(lr){3-3}\cmidrule(lr){4-5}\cmidrule(lr){6-7}
 & \textbf{RCR(\%)} & \textbf{RCR(\%)} & \textbf{RCR(\%)} & \textbf{MLE} & \textbf{RCR(\%)} & \textbf{MLE} \\
\midrule
Disease ($N$=30)  & 21.4 & 33.3 & 16.3 & \textbf{59.61} & 12.2 & 48.70 \\
GPA ($N$=30)      & 64.5 & 70.0 & 56.7 & \textbf{41.21} & 33.8 & 26.85 \\
Student ($N$=30)  & 33.9 & 36.7 & 46.4 & \textbf{0.37}  & 15.9 & -0.11 \\
\midrule
Disease ($N$=60)  & 25.5 & 31.7 & 30.9 & \textbf{75.65} & 71.1 & 54.32 \\
GPA ($N$=60)      & 51.5 & 61.7 & 49.1 & \textbf{44.34} & 22.5 & 15.44 \\
Student ($N$=60)  & 80.2 & 13.3 & 16.1 & \textbf{0.27}  & 15.6 & -0.49 \\
\midrule
Disease ($N$=120)& 28.7 & 38.3 & 30.7 & \textbf{76.56} & 52.8 & 55.52 \\
GPA ($N$=120)& 52.6 & 53.3 & 40.8 & 40.42 & 31.6 & \textbf{44.13} \\
Student ($N$=120)& 57.7 & 54.2 & 67.7 & \textbf{0.21}  & 17.2 & -0.46 \\
\midrule
Disease ($N$=200)& 35.5& 24.4& 38.4& \textbf{68.58}& 17.1& 61.11\\
GPA ($N$=200)& 51.0& 57.7& 64.5& \textbf{43.92}& 48.6& 31.67\\
Student ($N$=200)& 31.0& 29.4& 48.3& \textbf{0.50}& 12.4& -0.14\\
\midrule
 Disease ($N$=500)& 33.4& 22.4& 44.9& \textbf{61.41}& 34.4&58.76\\
 GPA ($N$=500)& 25.2& 27.0& 36.1& \textbf{37.90}& 38.4&27.07\\
 Student ($N$=500)& 40.2& 41.9& 50.3& \textbf{0.46}& 18.7&-1.19\\
\hline
\end{tabular}
\end{adjustbox}
\caption{Comparison of \emph{rule-guided} data vs.\ \emph{prompt-only} data across three datasets and varying training sizes ($N{=}30/60/120/200/500$). The better results are in bold.}
\label{tab:rcr_results}
\end{table}

\subsection{Mitigating Distributional Drift}
\label{com1}

\emph{How do rules enhance data quality?} To assess the extent to which generated samples adhere to the extracted \emph{association rules}, we define the \emph{Rule Compliance Rate} (RCR). Given a dataset $S$ and a set of association rules $\mathcal{R}$, RCR is the percentage of samples in $S$ that satisfy all rules in $\mathcal{R}$. 
We compare \textit{rule-guided} and \textit{prompt-only} generation across three representative datasets and five low-data regimes ($N{=}30/60/120/200/500$), evaluating both rule adherence (RCR) and downstream utility.
Table~\ref{tab:rcr_results} shows that under extreme low-data regimes ($N{=}30$), rule-guided generation significantly compensates for the lack of distributional evidence: compared to prompt-only generation, it produces samples with both higher rule adherence and stronger downstream utility (MLE). As $N$ increases ($N{=}60/120$), the difference becomes most evident in \emph{distributional alignment}. For instance, in the Disease dataset, prompt-only data attains superficially higher RCR but deviates from the real distribution and yields much lower utility. By contrast, rule-guided data achieves RCR values closer to real data and substantially improves utility. This indicates that high but distorted

RCR is not reliable, and that the true advantage of rules lies in correcting such bias and aligning the synthetic distribution with the target one. The t-SNE visualizations (Figure~\ref{fig:tsne}) further support this finding: rule-guided samples form distributions that remain close to the real manifold, whereas prompt-only samples drift into unsupported regions. Overall, rules effectively mitigate \textit{distributional drift} and enhance the utility of synthetic data across regimes.

We conduct further studies on \emph{Component I} to assess both robustness and design effectiveness (Appendix~\ref{app:com1}). Results show that while performance remains stable across different top-$k$ values, structured rule formats (i.e., “if-then”) and self-consistency denoising yield clear improvements, validating the effectiveness of our design. 

\subsection{Reducing Localized Redundancy}
\label{sec:com2}

\begin{table}[t]
\centering
\resizebox{\linewidth}{!}{
\begin{tabular}{l|cc|cc||c|c|c}
\toprule
& \multicolumn{4}{c||}{\textbf{Redundancy Metric}} & \multicolumn{3}{c}{\textbf{Different Granularity}} \\
\cmidrule(lr){2-5} \cmidrule(lr){6-8}
& \multicolumn{2}{c}{Gini} & \multicolumn{2}{c||}{Entropy} & Instance & Chunk & Dual \\
\cmidrule(lr){2-3} \cmidrule(lr){4-5}
& Value & MLE & Value & MLE & Only & Only & Granularity \\
\midrule
Disease ($N$=30)& 0.40 & \textbf{70.22} & 0.13 & 68.87 & 69.29 & 63.88 & \textbf{70.22} \\
GPA ($N$=30)& 0.23 & \textbf{43.14} & 0.06 & 39.88 & 39.39 & 41.02 & \textbf{43.14} \\
Student ($N$=30)& 0.58 & \textbf{0.38}  & 0.31 & 0.36  & \textbf{0.39} & 0.35 & 0.38 \\
\midrule
Disease ($N$=60)& 0.68 & \textbf{72.41} & 0.30 & 72.03 & 71.77 & 68.81 & \textbf{72.41} \\
GPA ($N$=60)& 0.23 & \textbf{44.58} & 0.03 & 38.53 & 43.25 & \textbf{45.04} & 44.58 \\
Student ($N$=60)& 0.61 & \textbf{0.34}  & 0.20 & 0.31  & 0.32  & 0.19 & \textbf{0.34} \\
\midrule
Disease ($N$=120)& 0.35 & \textbf{81.96} & 0.14 & 78.97 & 80.48 & 69.74 & \textbf{81.96} \\
GPA ($N$=120)& 0.29 & \textbf{47.77} & 0.08 & 45.12 & 42.99 & 40.77 & \textbf{47.77} \\
Student ($N$=120)& 0.31 & \textbf{0.29} & 0.09 & 0.33  & 0.05  & 0.45 & \textbf{0.29} \\
\midrule
Disease ($N$=200)& 0.66& \textbf{74.25}& 0.32& 65.47& 72.45& 65.03& \textbf{74.25}\\
GPA ($N$=200)& 0.19& 44.68& 0.1& \textbf{45.1}& 43.6& 43.07& \textbf{44.68}\\
Student ($N$=200)& 0.47& \textbf{0.52}& 0.14& 0.49& 0.50& 0.26& \textbf{0.52}\\
\midrule
 Disease ($N$=500)& 0.48& \textbf{72.48}& 0.39& 59.87& 60.97& 64.87&\textbf{72.48}\\
 GPA ($N$=500)& 0.34& \textbf{40.30}& 0.15& 39.97& \textbf{45.24}& 39.23&40.30\\
 Student ($N$=500)& 0.5& 0.52& 0.31& \textbf{0.53}& 0.51& 0.22&\textbf{0.52}\\
 \bottomrule
\end{tabular}
}
\caption{Evaluation of different redundancy metrics (Gini vs. Entropy) and filtering granularities (Instance-only, Chunk-only, Dual) within \emph{Component II}.}
\label{tab:dual}
\end{table}

\begin{figure}[t]
\centering
    \includegraphics[width=0.5\textwidth]{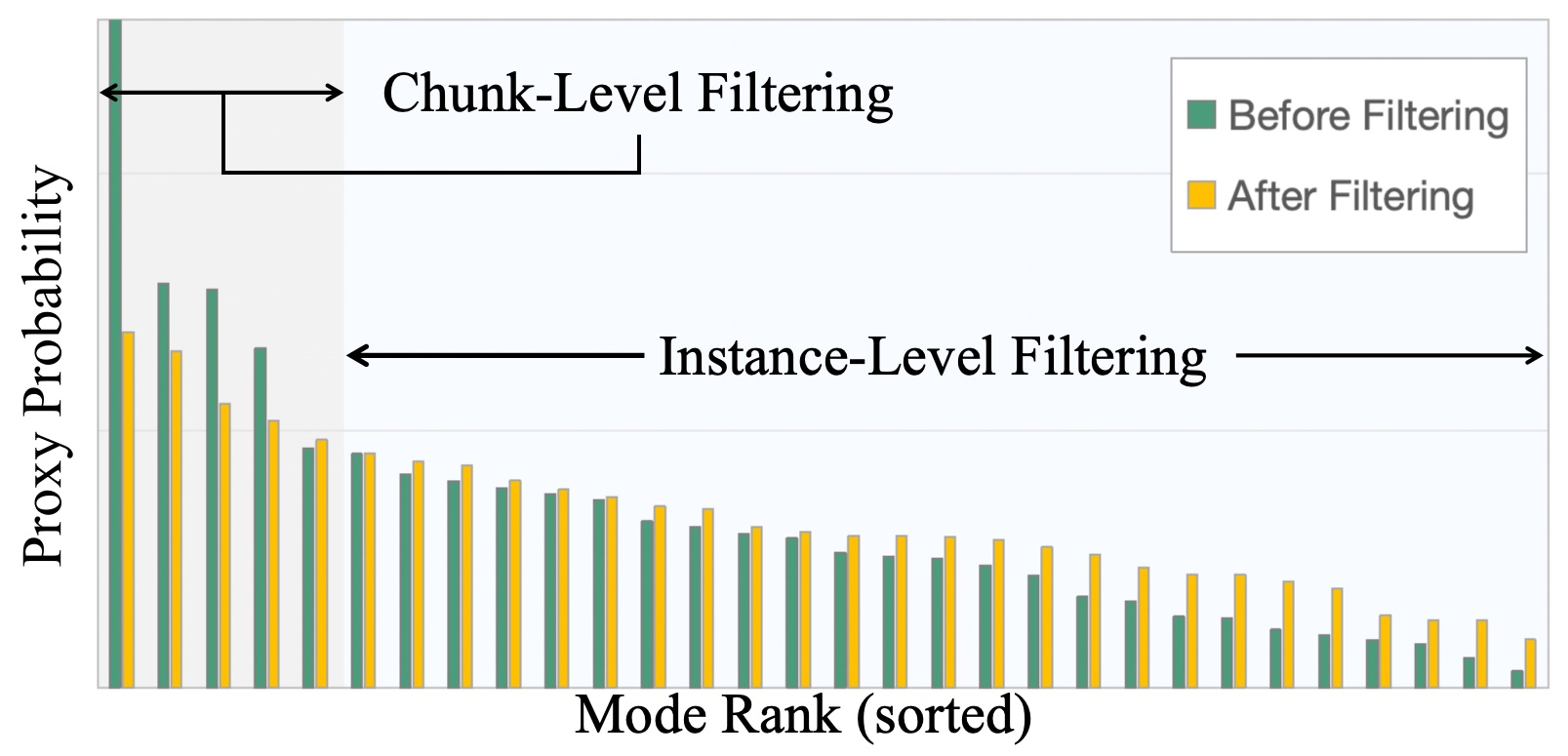}
    \caption{
    Proxy mode probabilities $p$ (sorted in descending order) before and after \emph{dual-granularity filtering}.
    }
\label{fig:mode_dist}
\end{figure}

\emph{How does filtering balance the distribution?}
To answer this question, we compare several \emph{redundancy} metrics and filtering granularities, and assess their impact on downstream utility (MLE) across datasets and data regimes. Table~\ref{tab:dual} yields two key insights.
(i) Not all redundancy metrics are equally effective. Although both Gini and entropy can quantify redundancy, Gini consistently leads to higher MLE across datasets and training sizes ($N$), with the largest gains under data scarcity. This suggests that redundancy in prompt-generated data is dominated by a small number of highly repeated patterns rather than widespread noise. Consistent with this, Gini better captures such concentration by emphasizing inequality, whereas entropy averages across regions and can understate redundancy by assigning relatively more weight to rare, potentially noisy instances.
(ii) Redundancy in LLM-generated data appears at multiple scales, both within individual samples and in batch-level concentration. Table~\ref{tab:dual} shows that dual-granularity filtering consistently achieves higher utility than instance-only or chunk-only variants across datasets and training sizes, indicating that both levels must be addressed. Instance-level filtering removes noisy outliers but does not correct global distribution imbalance. Chunk-level filtering reduces dominant high-density modes but can still leave localized redundancy. Combining both signals improves coverage and downstream utility. Figure~\ref{fig:mode_dist} supports this observation: chunk-level filtering reduces overrepresented modes, while instance-level filtering increases coverage in long-tail regions, leading to a more uniform and informative synthetic distribution.

We further validate the robustness of \emph{Component II} in Appendix~\ref{app:com2}. The Gini-driven pruning function peaks at intermediate retention, striking a balance between diversity and redundancy removal. Notably, Gini values stabilize with as few as 1,000 samples, confirming its suitability as a stable redundancy signal across generations.

\section{Conclusion}

In this paper, we identify two issues of prompt-based LLM tabular data generation in low-data regimes: \emph{distributional drift} and \emph{localized redundancy}, both of which can degrade downstream utility. To address them, we propose \textbf{ReFine}, a two-component framework that injects symbolic \emph{if--then} rules extracted from interpretable models to ground generation in domain structure, and applies dual-granularity filtering to suppress oversampling while preserving rare but informative samples. Across classification and regression benchmarks, ReFine achieves the best overall \emph{Total Rank} and strong average ranks across training regimes, with the largest gains in low-data settings. Further analyses show that the two components are complementary: rule guidance improves distributional faithfulness and dependency capture, while filtering enhances informative coverage and robustness on unseen datasets. Future work will explore adaptive retention and principled objectives for sample selection and refinement.

\clearpage
\bibliographystyle{named}
\bibliography{ijcai_bib}

\clearpage

\appendix
\addcontentsline{toc}{section}{APPENDIX}  

\setcounter{tocdepth}{2}   
\startcontents[appendix]
\printcontents[appendix]{l}{1}{\section*{Contents of Appendix}}
\clearpage

\section{Use of Large Language Models (LLMs)}
In preparing this manuscript, we employed a large language model (LLM) to assist with polishing the writing. Specifically, the LLM was used to improve clarity, grammar, and readability of the text. All substantive ideas, analyses, and conclusions are solely those of the authors.

\section{Limitations}
A remaining limitation is that the chunk-level retention schedule in our dual-granularity filtering is currently specified by a fixed, empirically chosen form. This design may be suboptimal under extreme distribution shifts or highly imbalanced rule coverage. Future work will explore adaptive retention mechanisms and more principled objectives for selecting and refining synthetic samples.

\section{Computational Cost Analysis}
\label{app:cost}

We analyze the computational footprint of \textit{ReFine} and show that it is lightweight and deployment-friendly. 
A key design goal of \textit{ReFine} is to avoid any LLM fine-tuning and keep all auxiliary steps inexpensive, so that the overall cost is comparable to prompt-only generation.

\paragraph{No fine-tuning overhead.}
\textit{ReFine} does not fine-tune the LLM. All improvements come from prompt augmentation (rule grounding) and post-generation filtering. 
Therefore, the dominant cost remains the LLM decoding cost, which is identical to prompt-only baselines under the same generation budget.

\paragraph{API latency for LLM generation.}
In practice, the wall-clock time of LLM-based generation is dominated by API latency (network overhead and decoding throughput), which depends on deployment conditions (e.g., rate limits, batching, and concurrency) and is therefore not directly comparable across environments.
Importantly, \textit{ReFine} uses the same generation budget as prompt-only baselines, so its generation-time cost is essentially identical; the only additional API usage comes from a small number of short rule-consolidation calls.

\paragraph{Rule extraction and consolidation.}
Rule mining is performed with small interpretable tree models trained on the low-data split. 
In practice, training and path extraction complete in under one second on a CPU per dataset/split. 
The subsequent rule consolidation and denoising uses only a small number of short LLM calls (each \(<1\)k tokens), resulting in negligible overhead compared to data generation.

\paragraph{Filtering cost.}
Dual-granularity filtering scales linearly with the number of generated samples.
Instance-level scoring and proxy-density estimation are simple vectorized operations with \(O(M)\) time for \(M\) generated instances.
The only relatively expensive step is chunk-level scoring, which trains a small XGBoost model on the low-data split and records prediction trajectories. 
This step takes approximately 1--2 minutes on CPU in our implementation, and does not require GPUs.

\paragraph{Practical runtime profile and comparison.}
Overall, the full \textit{ReFine} pipeline runs comfortably in CPU-only environments. 
Since generation cost is unchanged from prompt-only baselines, the additional overhead primarily comes from chunk-level scoring and is modest relative to the total experimental runtime. 
In contrast, gradient-based tabular generators (e.g., diffusion or GAN training pipelines) typically require iterative optimization and GPU resources, leading to substantially higher computational costs for comparable evaluation suites.

\paragraph{Reproducibility and hyperparameters.}
\textit{ReFine} introduces no LLM fine-tuning and only minimal tunable choices. 
Most thresholds are computed from closed-form statistics, and the chunk size is selected automatically via surprisal minimization, which keeps the method easy to deploy and highly reproducible.

\section{Dataset Contamination Test}

\label{appendix:dataset}
\subsection{\texttt{tabmemcheck}}
To rigorously evaluate the generalization capabilities of prompt-based tabular data generation, we assess potential training set contamination in popular benchmark datasets. Following concerns raised in recent studies~\cite{xu2024benchmark}, we employ \texttt{tabmemcheck}~\cite{bordt2024colm}, a diagnostic tool for detecting dataset memorization in LLMs. Specifically, we use two tests:
\begin{itemize}
\item Feature Names Test: Given a few sample rows, the LLM is prompted to infer corresponding \textit{Column Names}.

\item Header Test: The LLM is prompted to reconstruct the header (column names) and the first \textit{few data rows} of the dataset in CSV format.
\end{itemize}

\subsection{Example: \texttt{Heart} Dataset (GPT-4o)}

\begin{figure*}[h]
    \centering
    \includegraphics[width=\linewidth]{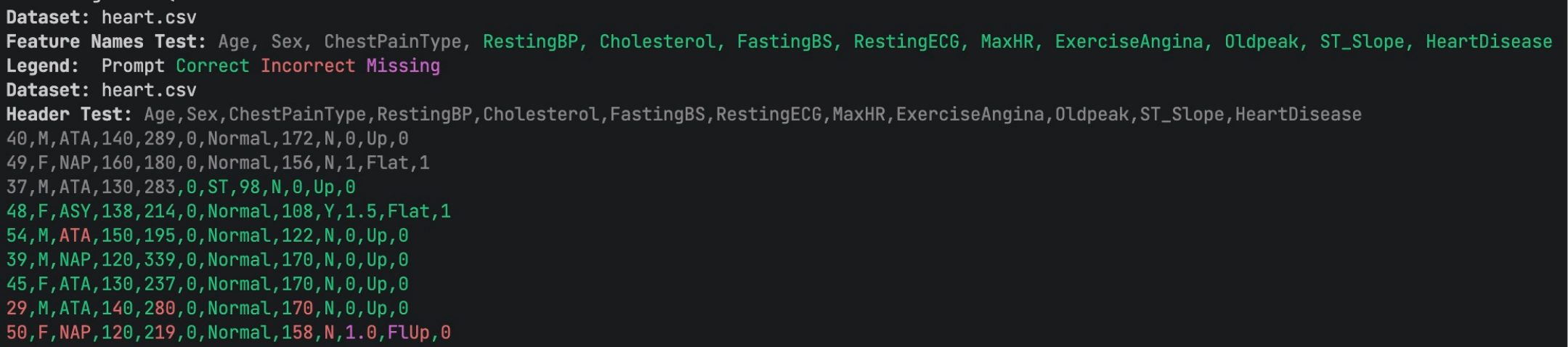}
    \caption{\textbf{GPT-4o reproduces the \texttt{heart} dataset.} Field names and values are copied precisely, indicating contamination.}
    \label{fig:heart_results}
\end{figure*}

As shown in Figure~\ref{fig:heart_results}, GPT-4o accurately recovers the feature names and generates realistic rows for the \texttt{heart} dataset, despite limited prompting. Fields such as \texttt{ST\_Slope} and \texttt{RestingECG} are generated verbatim, indicating memorization.

This demonstrates that \textbf{strong LLMs may replicate datasets}, compromising fair evaluation.

\subsection{Example: \texttt{Adult} Dataset (Qwen2.5-32b)}

\begin{figure*}[tb]
    \centering
    \includegraphics[width=\linewidth]{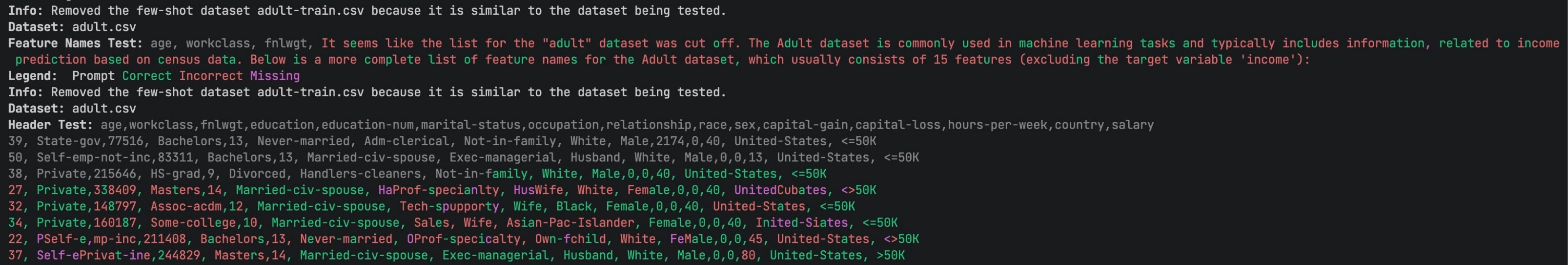}
    \caption{\textbf{Qwen2.5-32b recognizes the \texttt{adult} dataset.} While exact headers are missing, the model describes the dataset and its usage.}
    \label{fig:adult_results}
\end{figure*}

In contrast, Qwen2.5-32b fails to regenerate exact feature names but instead \textbf{identifies the dataset by name and describes its contents and use cases} (as shown in Figure~\ref{fig:adult_results}). This demonstrates semantic-level exposure rather than strict memorization.

Such dataset-wide familiarity still violates the assumption of data independence and motivates filtering benchmarks during evaluation.

\begin{table*}[b]
\centering
\begin{adjustbox}{max width=\textwidth}
\begin{tabular}{l|cc|cc|cc|cc}
\toprule
& \multicolumn{2}{c}{GPT-4o-0806} & \multicolumn{2}{c}{GPT-3.5-turbo-1106} & \multicolumn{2}{c}{Qwen2.5-32b-Instruct} & \multicolumn{2}{c}{Qwen2.5-14b-Instruct} \\
\cmidrule(lr){2-3} \cmidrule(lr){4-5} \cmidrule(lr){6-7} \cmidrule(lr){8-9}
 & Feature Names & Header & Feature Names & Header & Feature Names & Header & Feature Names & Header \\
\midrule
Other Datasets & 0\% & 0\% & 0\% & 0\% & 0\% & 0\% & 0\% & 0\% \\
Adult & 86.67\% & 75.0\% & 86.67\% & 100.0\% & 0\%\(*\) & 25.0\% & 0\%\(*\) & 80.0\% \\
Heart & 100.0\% & 55.56\% & 25\% & 22.22\% & 0\% & 11.11\% & 0\% & 12.50\% \\
\bottomrule
\multicolumn{9}{l}{$^{\mathrm{*}}$Although no correct column names were produced, the LLM identified ``\textit{the Adult dataset from the UCI Machine Learning Repository}''}
\end{tabular}
\end{adjustbox}
\caption{\textbf{Feature Names and Header Test} Results. \textbf{Feature Names Test}: The LLM infers column names from sample rows. \textbf{Header Test}: The LLM completes the header and first few rows of the CSV. Percentages indicate accuracy for each dataset and LLM.}
\label{tab:datasets}
\end{table*}

\subsection{Implications and Dataset Filtering}

These examples reveal that both closed-source (GPT-4o) and open-source (Qwen2.5) models can memorize benchmarks—either by reproducing format or semantically recognizing dataset identity. 
Table~\ref{tab:datasets} summarizes model accuracy across these tests. High scores on these tasks suggest that the LLM has memorized structural or content-level information about the dataset, potentially inflating downstream performance.

\section{Experimental Setup Details}
\label{appendix:setup}

\begin{table*}[b]
  \centering
  \caption{Hyper-parameter settings for non-llm based baseline methods.}
  \label{tab:hyperparams}
  \begin{tabular}{lccccc}
    \toprule
     Method / Phase  & Epochs & Batch size & Optimizer & Initial LR & Additional settings \\
    \midrule
    DRL (CTGAN-based)  & 150 & 24 & Adam & $2\times10^{-4}$ & --- \\
    \midrule
    TabSyn – VAE phase  & 3000 & 4096 & Adam & $10^{-3}$ & --- \\
    TabSyn – DDPM phase 
        & up to 10001\tablefootnote{Early stopping: if current loss $\ge$ best loss for 500 consecutive epochs.} 
        & 4096 
        & Adam 
        & $10^{-3}$ 
        & LR scheduler\tablefootnote{\texttt{ReduceLROnPlateau} (PyTorch) applied during the DDPM stage.} \\
    \bottomrule
  \end{tabular}
  \label{tab:hyperparams}
\end{table*}

\subsection{Our Method}
\textbf{Model Configuration.} 
We use GPT-3.5-Turbo-1106 as the backend model for both association rule extraction and data generation. Each baseline is configured to generate roughly 2,000 synthetic samples per dataset.

\textbf{Rule-Guided Generation.} 
We set \(k=3\) to select top-performing trees from the Random Forest for rule extraction; for GPA at \(N=500\), the resulting decision structure exceeds the prompt length budget, so we use \(k=1\). We apply self-consistency by aggregating 5 independently sampled generations for rule consolidation and denoising.

\textbf{Dual-Granularity Filtering.} 
We use XGBoost as the reference model \(\mathcal{M}\) for evaluating chunk-level informativeness, and search chunk sizes \( S \in \{20, 25, \ldots, 60\} \) during tuning.

\textbf{Evaluation.} For each run, we train XGBoost on 1,000 synthetic samples under 10 random seeds and evaluate on the corresponding real dataset. Classification tasks are measured with F1 score; regression tasks with \(R^2\).

\subsection{Hyperparameter Settings for Baselines}

For \textbf{non-LLM based method}, we follow the standard training configurations reported in prior work. The detailed hyperparameter settings are summarized in Table~\ref{tab:hyperparams}. 

For \textbf{LLM-based methods}, we standardize the generation temperature across all models.
We use a fixed sampling temperature of 0.9 for all LLM prompt-based approaches, including baselines such as \textsc{CLLM} and \textsc{EPIC} as well as our method. Unless otherwise specified, all LLM generations use this temperature. The GReaT's hyperparameter settings are summarized in Table~\ref{tab:hyperparams}.

\section{Further Study on Component I}
\label{app:com1}
We conduct a series of supplementary analyses to unpack this question. In Section \ref{app:1.1}, we test the stability of extracted rules by varying the number of top-$k$ trees and find that rule quality remains robust across settings. In Section \ref{app:1.2}, we compare rule formats and show that explicit \emph{if–then} clauses provide clearer guidance than natural-language paraphrases. In Section \ref{app:1.3}, we examine rule denoising strategies and demonstrate that self-consistency aggregation yields more reliable rules than single-pass or CoT prompting. In Section \ref{app:1.4}, we validate transferability across different LLM backbones, confirming that rule guidance consistently improves synthetic data quality regardless of model scale. In Section \ref{app:1.5}, a case study illustrates how noisy tree paths are distilled into compact and interpretable rules through merging and denoising, highlighting both robustness and interpretability. Together, these studies confirm that rules enhance data quality by providing stable, precise, and broadly applicable generation guidance.

\begin{table*}[htbp]
\centering
\renewcommand{\arraystretch}{1.2}
\begin{adjustbox}{max width=\textwidth}
\begin{tabular}{l|cc|cc|cc|cc|cc}
\toprule
 & \multicolumn{2}{c|}{$k=1$} & \multicolumn{2}{c|}{$k=2$} & \multicolumn{2}{c|}{$k=3$} & \multicolumn{2}{c|}{$k=5$} & \multicolumn{2}{c}{$k=10$} \\
& $D_{\text{test}}$ & $D_{\text{train}}$ 
& $D_{\text{test}}$ & $D_{\text{train}}$ 
& $D_{\text{test}}$ & $D_{\text{train}}$ 
& $D_{\text{test}}$ & $D_{\text{train}}$ 
& $D_{\text{test}}$ & $D_{\text{train}}$ \\
\midrule
Disease (N=30)  & 63.44 & 83.33 & 36.38 & 53.33 & 29.79 & 40.00 & 44.61 & 50.00 & 72.99 & 63.33 \\
GPA (N=30)      & 62.45 & 70.00 & 62.45 & 70.00 & 58.17 & 66.67 & 62.57 & 70.00 & 63.76 & 60.00 \\
Student (N=30)  & 46.12 & 56.67 & 28.83 & 50.00 & 26.53 & 50.00 & 50.66 & 73.33 & 36.45 & 30.00 \\
\midrule
Disease (N=60)  & 60.67 & 76.67 & 34.92 & 35.00 & 43.04 & 53.33 & 26.41 & 28.33 & 50.93 & 56.67 \\
GPA (N=60)      & 41.90 & 36.67 & 42.02 & 56.67 & 58.02 & 70.00 & 49.01 & 65.00 & 49.70 & 58.33 \\
Student (N=60)  & 17.72 & 13.33 & 54.77 & 48.33 & 34.46 & 36.67 & 43.07 & 35.00 & 57.95 & 61.67 \\
\midrule
Disease (N=120) & 51.21 & 63.33 & 27.16 & 41.67 & 30.26 & 39.17 & 28.63 & 38.33 & 28.78 & 38.33 \\
GPA (N=120)     & 40.98 & 44.17 & 48.50 & 48.33 & 60.74 & 41.67 & 54.52 & 39.11 & 64.73 & 54.62 \\
Student (N=120) & 29.57 & 31.67 & 65.82 & 63.33 & 42.58 & 43.33 & 32.36 & 34.17 & 28.97 & 34.17 \\
\midrule
Disease (N=200) & 87.99 & 86.50 & 25.45 & 34.50 &25.35&34.00&25.14&34.50&24.32&33.50 \\
GPA (N=200)     &  49.91&42.50&63.50&59.00&62.41&57.50&63.69&56.00&58.39&51.00\\
Student (N=200) & 40.47&41.00&22.73&23.50&28.16&24.50&25.55&26.50&17.84&20.00 \\
\midrule
Disease (N=500) &23.71&35.20&27.59&36.20&22.68&32.20&19.16&30.20&22.56&32.00  \\
GPA (N=500)     & 28.12 & 30.80&-&-&-&-&-&-&-&- \\
Student (N=500) & 38.55&39.00&26.87&26.20&49.86&48.80&40.85&38.40&30.34&30.34 \\
\bottomrule
\end{tabular}
\end{adjustbox}
\caption{Rule Compliance Rate (RCR) Results for different $k$ values ($D_{\text{test}}$ and $D_{\text{train}}$).}
\label{app:topk}
\end{table*}

\subsection{Effect of Top-$k$ Trees in Rule Extraction.}
\label{app:1.1}
To examine the robustness of the extracted rules, we vary the number of top-$k$ trees used for rule extraction (\(k\in\{1,2,3,5,10\}\)).
Across settings, the resulting rule-compliance rates (RCR) are largely stable, indicating that the induced rules capture consistent feature--label dependencies and are not overly sensitive to the specific choice of \(k\).
In practice, \(k=1\) or \(2\) already yields comparable RCR, while increasing \(k\) can lead to longer rule lists.
For some configurations, the aggregated rule text exceeds the prompt length budget, and we mark these cases as ``--'' in Table~\ref{app:topk}.
We adopt \(k=3\) by default, as it offers broader rule coverage than very small \(k\) while remaining prompt-feasible and avoiding the additional overhead of larger \(k\).
This choice provides a favorable trade-off between robustness, coverage, and practical prompting constraints.


\subsection{Different Rule Format Comparison.}
\label{app:1.2}
\begin{figure*}[t]
    \centering
    \includegraphics[width=\textwidth]{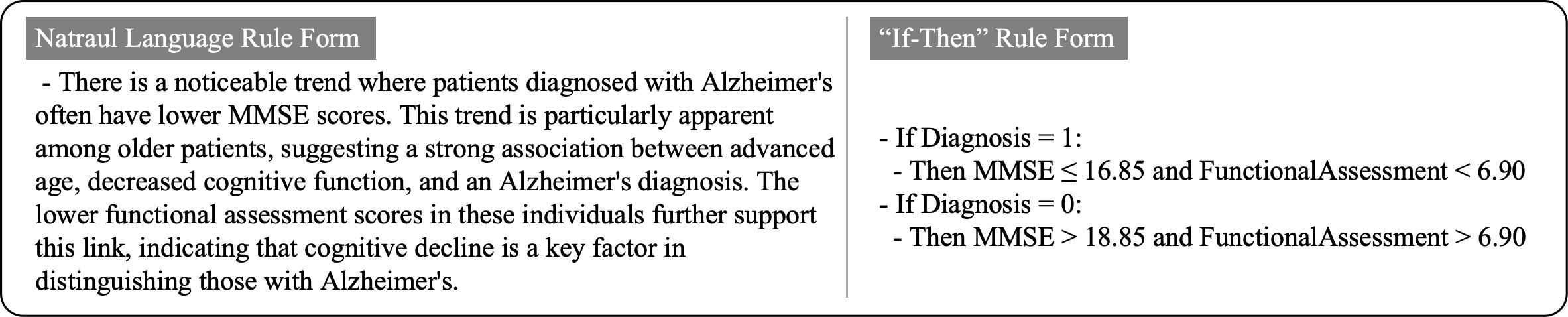}
    \caption{Illustrative ``if–then'' Form and its Natural-Language paraphrase derived from the \textit{Disease} dataset ($N{=}30$).}
    \label{fig:rule-form}
\end{figure*}

To gauge whether the \textit{explicit} ``if--then'' representation is essential to the success of Rule-Guided Generation, we recast every Random-Forest path into two formats: (i) its original ``if--then'' clause and (ii) a concise natural-language paraphrase, which automatically produced by prompting the LLM to restate each path in natural language. These two formats reflect two distinct rule-extraction schemes. A side-by-side example of the two rule forms is shown in Figure~\ref{fig:rule-form}.
As shown in Table~\ref{tab:rule-form}, \emph{if--then} rules outperform both natural-language rules and the No-Rule baseline, demonstrating the benefit of symbolic structure. While both rule-based approaches surpass the baseline, the symbolic form delivers more reliable performance. This advantage stems from two key factors: (i) Random Forests extract concise and faithful feature--label dependencies even in low-data settings, and (ii) the \emph{if--then} format retains explicit numeric boundaries and dependent constraints, unlike natural language, which tends to weaken precision \cite{symboli2024}. 
By guiding generation through explicit symbolic rules, the LLM is directed toward semantically coherent subspaces, resulting in higher-quality samples.

\begin{table*}[t]
\centering
\begin{minipage}{0.48\textwidth}
\centering
\adjustbox{max width=\linewidth}{
\begin{tabular}{l|c|c|c}
\toprule
& \textbf{No Rule} & \textbf{Natural Language} & \textbf{``if-then'' Form} \\
\cmidrule(lr){1-4}
Disease & $49.13 \pm 1.2$ & $57.54 \pm 1.7$ & \textbf{59.61 $\pm$ 1.5} \\
GPA     & $24.80 \pm 4.0$ & $37.78 \pm .60$ & \textbf{41.21 $\pm$ .97} \\
Student & $-0.11 \pm .12$  & $-0.79 \pm .53$  & \textbf{0.37 $\pm$ .02} \\
\bottomrule
\end{tabular}
}
\caption{Performance under different rule representations ($n=30$).}
\end{minipage}%
\hfill
\begin{minipage}{0.48\textwidth}
\centering
\adjustbox{max width=\linewidth}{
\begin{tabular}{l|c|c|c}
\toprule
& \textbf{Single-Pass} & \textbf{COT} & \textbf{Self-Consistency} \\
\cmidrule(lr){1-4}
Disease & $54.26 \pm 3.2$ & $58.46 \pm .98$ & \textbf{59.61 $\pm$ 1.5} \\
GPA     & $24.80 \pm 4.0$ & $38.48 \pm .82$ & \textbf{41.21 $\pm$ .97} \\
Student & $0.35 \pm .06$  & $0.35 \pm .02$  & \textbf{0.37 $\pm$ .02} \\
\bottomrule
\end{tabular}
}
\caption{Performance under different \emph{rule denoising} strategies ($n=30$).}
\label{tab:rule-form}
\end{minipage}
\end{table*}

\begin{table*}[b]
\normalsize
\centering
\begin{adjustbox}{max width=\textwidth}
\begin{tabular}{c|c|c|c|c|c}
\toprule
\multicolumn{2}{c|}{\textbf{Original Data}} & \multirow{2}{*}{\textbf{Rules Generator}} & \multicolumn{3}{c}{\textbf{Data Generator}} \\
\cmidrule(lr){1-2} \cmidrule(lr){4-6}
Datasets & Real Data & & Qwen2.5-14b-Instruct & Qwen2.5-32b-Instruct & GPT-4o-0806 \\
\midrule

\multirow{3}{*}{Disease} & \multirow{3}{*}{$93.68 \pm 1.4$} 
& No Rules             & $59.52 \pm 1.9$  & $65.12 \pm 2.3$ & $62.26 \pm 3.2$ \\
& & Qwen2.5-32b-Instruct & $66.34 \pm 1.4$  & $70.16 \pm .80$ & $67.38 \pm .85$ \\
& & GPT-4o-0806          & \textbf{70.21 $\pm$ 1.9} & $67.74 \pm 1.4$ & $64.04 \pm 2.7$ \\

\midrule
\multirow{3}{*}{GPA} & \multirow{3}{*}{$47.29 \pm .90$} 
& No Rules             & $20.67 \pm .87$  & $30.23 \pm 1.4$ & $27.58 \pm 2.3$ \\
& & Qwen2.5-32b-Instruct & $40.62 \pm 1.4$  & $39.13 \pm 3.4$ & \textbf{47.07 $\pm$ 0.5} \\
& & GPT-4o-0806          & $42.25 \pm 1.4$  & $31.99 \pm 2.9$ & $44.34 \pm 0.6$ \\

\midrule
\multirow{3}{*}{Student} & \multirow{3}{*}{$0.67 \pm .07$} 
& No Rules             & $0.25 \pm .04$   & $0.32 \pm .03$ & $-1.29 \pm .27$ \\
& & Qwen2.5-32b-Instruct & $0.30 \pm .03$   & \textbf{0.33 $\pm$ .03} & $0.25 \pm .01$ \\
& & GPT-4o-0806          & $0.18 \pm .05$   & $-0.38 \pm .16$ & $-0.01 \pm .14$ \\

\bottomrule
\end{tabular}
\end{adjustbox}
\caption{Main results on benchmark datasets ($n$=30). We report \textbf{F1 score} for classification tasks and \textbf{R\textsuperscript{2}} for regression tasks.
The best result in each row is shown in \textbf{bold}.}
\label{tab:model}
\end{table*}

\subsection{Robustness of Rule Denoising Strategy.}
\label{app:1.3}
We evaluate how different Rule Denoising strategies affect data quality. We compare our proposed \emph{Self-Consistency} \emph{Rule Denoising} (described in Section~\ref{sec:rule-denoising}) against two alternatives: (i) \textbf{Single-Pass}, which denoise rules in one step without verification, and (ii) \textbf{Chain-of-Thought (CoT)} prompting, which guides the LLM to reason step-by-step during rule denoising \cite{wei2022chain}.
Results in Table~\ref{tab:rule-form} reveal that both \textbf{Single-Pass} and \textbf{CoT} aggregation yield less consistent performance across datasets. Their reliance on one-shot reasoning makes them vulnerable to local inconsistencies and stochastic behavior in LLM outputs. In contrast, self-consistency aggregation enforces cross-run agreement and filters out unstable logic fragments, leading to more robust rule sets and better downstream fidelity.

\subsection{Rule-Based Methods Generalize Across LLMs}
\label{app:1.4}
To evaluate the robustness of \emph{Rule-Guided Generation} in different LLM backbones, we compare performance using the MLE under varying combinations of rules generators and data generators. The result as shown in Table~\ref{tab:model}, across all tested LLMs which ranging from mid-sized (Qwen2.5-14b) to larger models (Qwen2.5-32b and GPT-4o), the introduction of association rules consistently improves the quality of synthetic data. Notably, even when rules are extracted by weaker models (e.g., Qwen2.5-32b), stronger models like GPT-4o still benefit from them—highlighting that rule-based guidance contributes independently of LLMs capacity. This demonstrates the generality and transferability of our design.

\begin{figure*}[tb]
    \centering
    \includegraphics[width=\linewidth]{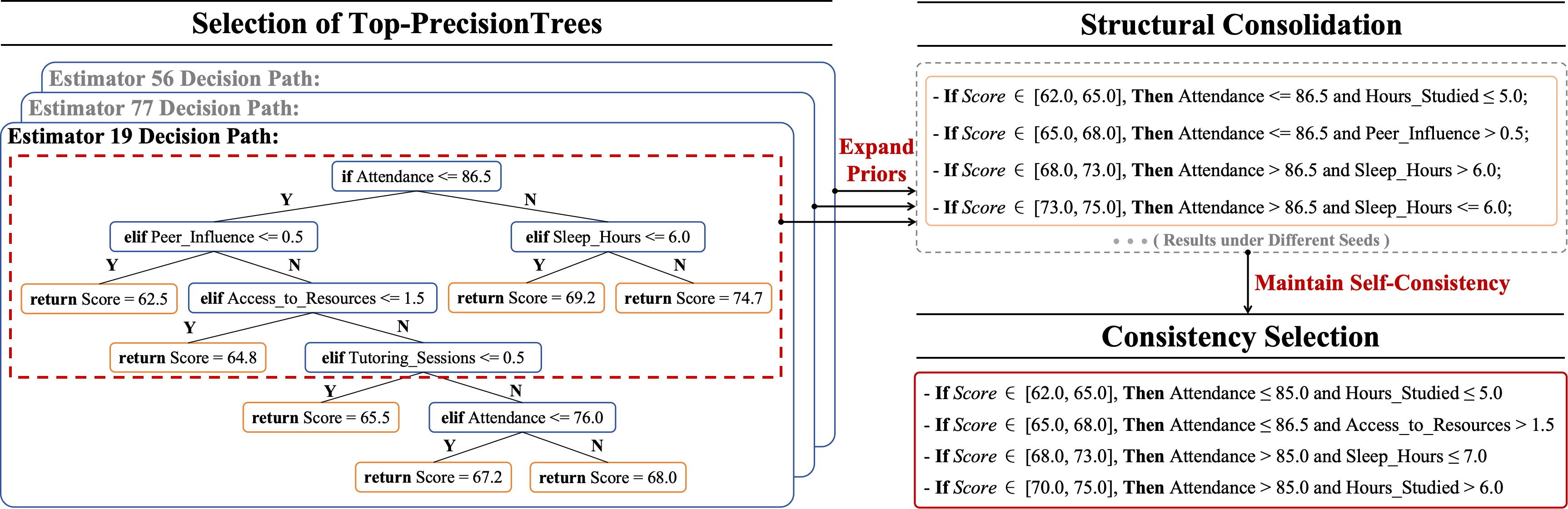}
    \caption{Case study for Component I on the \textit{Student} dataset. Noisy tree paths are denoised into a self-consistent symbolic formulas set and later guides data generation.}
    \label{case1}
\end{figure*}

\subsection{Illustrative Example of Rule Generalization \& Denoising.}
\label{app:1.5}
To illustrate the practical functioning of \textbf{Component~I}, we present a case study demonstrating how symbolic rules are distilled into interpretable \emph{if–then} forms through rule merging and aggregation. Intermediate outcomes are visualized in Figure~\ref{case1}. In this example, decision trees from the trained random forest exhibit conflicting paths—for example, assigning different $Exam\_Score$ values to overlapping input regions such as $Attendance \leq 86.5$ and $Attendance \leq 76.0$. The merging phase addresses these inconsistencies by consolidating noisy rule fragments into coherent patterns, such as those involving $Sleep\_Hours$. 
To improve robustness, the training and generalization process is repeated under multiple random seeds. The denoising step then retains only those patterns that appear consistently across runs, thereby filtering out unstable conditions and enforcing self-consistency.
This case highlights two key strengths of Component I: (1) it distills noisy and fragmented tree logic into compact rules that capture meaningful relationships in low-data regimes; (2) it produces one concise association rule per segment, enhancing both interpretability and generation quality.

\section{Further Study on Component II}
\label{app:com2}

As shown in Figure~\ref{pipeline}, synthetic data generated under low-data regimes often exhibit localized redundancy: over-representation of high-density modes and under-sampling of rare but informative modes. 
Prior methods typically employ instance-level filtering, but lack a global view of the overall distribution \cite{cllm}.
To address this challenge, we introduce \textbf{Dual-Granularity Filtering} as Component~II, which aims to post-process the raw synthetic data \(D_{\text{syn}}\) by selectively filtering samples at two levels of granularity.
The combined use of chunk-level and instance-level filters allows us to correct both global density distortion. \emph{Chunk-level filtering} mitigates oversampling from the high-density subset, while \emph{instance-level filtering} improves the quality of under-sampled instances from the low-density subset.
The complete \emph{Dual-Granularity Filtering} workflow is summarized in Algorithm~\ref{alg:stage2}, which takes the raw synthetic dataset \(D_{\text{syn}}\) as input and outputs a augmented synthetic dataset \(D_{\text{aug}}\) for downstream learning.

To make this filtering effective, we first estimate where and how redundancy occurs in \(D_{\text{syn}}\). This is non-trivial in our low-data and mixed-type tabular setting, where LLM-generated samples lack explicit sampling probabilities~\cite{sehwagrobust,sehwag2021improving,wu2025survey}. 

\subsection{Estimating Redundancy via Proxy-Based Density Analysis.}
To support distribution-aware filtering, we introduce a \emph{proxy-based} method to estimate local sampling frequencies with respect to the training set \(\mathcal{D}_{\mathrm{train}}\).
We use the \emph{Distance to Closest Record} (DCR) as a lightweight, task-agnostic heuristic to measure how each synthetic sample aligns with the training set~\cite{osorio2024privacy,steier2025synthetic}. Specifically, for each synthetic sample \(x_i \in \mathcal{D}_{\mathrm{syn}}\), we assign it to its nearest training point \(s_j \in \mathcal{D}_{\mathrm{train}}\) based on the DCR metric:
\begin{equation}
\text{Cluster}^*(i) = \arg\min_{1 \le j \le n} \mathrm{DCR}(x_i, s_j).
\label{eq:j}
\end{equation}
By counting the number of synthetic data assigned to each seed and normalizing the result, we obtain a discrete \emph{proxy distribution} \(p = (p_1, \dots, p_n)\), where \(p_j\) reflects the local sampling density around seed \(s_j\). 
To quantify localized redundancy, we compute the \emph{Gini coefficient}:
\begin{equation}
G(p) = \frac{1}{n-1} \sum_{j=1}^n (2j - n - 1)\, p_{(j)},
\label{eq:gini}
\end{equation}
where \(p_{(1)} \le \cdots \le p_{(n)}\) are the sorted components of the proxy distribution \(p\). A larger \(G(p)\) indicates greater concentration of synthetic samples around fewer seeds, i.e., higher redundancy from \(\mathcal{D}_{\mathrm{train}}\).

Unlike entropy-based metrics, the Gini coefficient is more sensitive to sharp density disparities, especially in the presence of dominant high-density modes, while being less affected by sparse long-tail noise \cite{biro2020gintropy,xinying2023guide}. This makes it well-suited for detecting localized redundancy in \(\mathcal{D}_{\mathrm{syn}}\).
To enable generalization across settings, we use the global redundancy score $G(p)$ as a unified control signal, and further utilize $G(p)$ to partition the synthetic dataset by estimated sampling density. Specifically, we sort the training seeds in descending order of their proxy frequencies \( p_j \), and identify the smallest index \(K\) such that \( \sum_{j=1}^K p_j \ge G(p) \). Synthetic samples associated with these top-\(K\) seeds form the high-density subset \( \mathcal{D}_{\mathrm{high}} \), while the remaining samples constitute the low-density subset \( \mathcal{D}_{\mathrm{low}} = \mathcal{D}_{\mathrm{syn}} \setminus \mathcal{D}_{\mathrm{high}} \).

\subsection{Dual-Granularity Filtering.}
\label{sec:dual_granularity}
Filtering synthetic samples is challenging due to the variability in localized redundancy and data quality across datasets \cite{chen2024unveiling,goyal2024systematic}. 
Given the partitioning into high-density (\(\mathcal{D}_{\text{high}}\)) and low-density (\(\mathcal{D}_{\text{low}}\)) subsets, we apply distinct filtering strategies at two levels of granularity. \emph{Chunk-level filtering} targets redundancy in \(\mathcal{D}_{\text{high}}\), while \emph{instance-level filtering} focuses on improving quality within \(\mathcal{D}_{\text{low}}\).

\subsubsection{1) Chunk-level Filtering in High-Density Subset.}
\label{app:com2.1}
High-density regions often contain subsets of highly similar samples, resulting from the generation pipeline. Evaluating individual samples in these regions is often noisy and ineffective, as over-sampled patterns tend to dominate locally. To address this, we apply a \emph{chunk-level filtering} strategy that evaluates samples in groups based on their collective utility.

We partitioned \(\mathcal{D}_{\mathrm{high}}\) into non-overlapping chunks \(\{\mathcal{C}_1, \dots, \mathcal{C}_N\}\) of fixed size \(S\). To assess the utility of each chunk, we train a lightweight reference model \(\mathcal{M}_{cur}\) (e.g., XGBoost) on \(\mathcal{D}_{\mathrm{train}}\). For each synthetic pair \((x, y)\), we track the model \(\mathcal{M}_{cur}\)’s prediction over \(T\) epochs and compute a task-dependent reliability signal from its prediction trajectory. 

These per-sample signals are then aggregated at the chunk level to reflect the overall utility.

\paragraph{1.1) Classification.}
In classification settings, we evaluate how consistently the model assigns high probability to the gold label.
Let $\mathbb{P}_{\mathcal{M}_{cur,t}}(y \mid x)$ denote the predicted probability for label $y$ at epoch $t$.
We define the per-sample \emph{correctness} score as
\begin{equation}
\label{eq:score_class}
\mathrm{Correctness}(x, y)
= \frac{1}{T} \sum_{t=1}^{T}
\mathbf{1}\!\left( \mathbb{P}_{\mathcal{M}_{cur,t}}(y \mid x) > 0.5 \right),
\end{equation}
which captures the fraction of training epochs in which the model is confidently correct.
Given a chunk $\mathcal{C}_r = \{(x_1,y_1),\dots,(x_S,y_S)\}$, we compute its chunk-level utility by averaging per-sample correctness:
\begin{equation}
\label{eq:chunk_score_class}
\mathrm{Score}(\mathcal{C}_r)
= \frac{1}{S} \sum_{(x_i, y_i) \in \mathcal{C}_r}
\mathrm{Correctness}(x_i, y_i).
\end{equation}
Chunks with higher scores are prioritized during filtering.

\paragraph{1.2) Regression.}
For regression tasks, per-sample correctness is no longer well-defined.
Instead, we characterize the prediction trajectory using two complementary signals: the average prediction error and the temporal dispersion across epochs.
Let $\hat{y}_t(x) = \mathcal{M}_{cur,t}(x)$ denote the scalar prediction at epoch $t$.
We define
\begin{equation}
\label{eq:reg_err}
\mathrm{Err}(x, y)
= \frac{1}{T} \sum_{t=1}^{T} \left| \hat{y}_t(x) - y \right|,
\qquad
\bar{y}(x)
= \frac{1}{T} \sum_{t=1}^{T} \hat{y}_t(x),
\end{equation}
and the predictive dispersion
\begin{equation}
\label{eq:reg_uncert}
\mathrm{Uncert}(x)
= \sqrt{ \frac{1}{T} \sum_{t=1}^{T}
\left( \hat{y}_t(x) - \bar{y}(x) \right)^2 }.
\end{equation}
The first term measures how well the model fits the sample on average, while the second captures the stability of predictions across epochs.
Chunk-level aggregation proceeds as
\begin{equation}
\label{eq:chunk_score_reg}
\begin{aligned}
\mathrm{Err}(\mathcal{C}_r)
&= \frac{1}{S} \sum_{(x_i, y_i) \in \mathcal{C}_r}
\mathrm{Err}(x_i, y_i), \\
\mathrm{Uncert}(\mathcal{C}_r)
&= \frac{1}{S} \sum_{(x_i, y_i) \in \mathcal{C}_r}
\mathrm{Uncert}(x_i).
\end{aligned}
\end{equation}
Chunks are ranked in ascending order by the pair
$\big(\mathrm{Err}(\mathcal{C}_r),\, \mathrm{Uncert}(\mathcal{C}_r)\big)$,
preferring low-error and low-variance groups.

Static pruning thresholds often fail to generalize across datasets with varying redundancy levels, and can exacerbate distributional imbalance.
To enable density-aware pruning, we adopt a redundancy-adaptive retention schedule driven by the global redundancy score
$G(p)$.

To ensure numerical stability when the Gini coefficient $G(p)$ is small, we apply a lower-bound clipping:
\begin{equation}
G_{\mathrm{safe}} = \max(G(p), \varepsilon),
\end{equation}
where $\varepsilon = 10^{-3}$ is a small constant used to prevent divergence in the logarithmic mapping.

We then compute the pruning ratio via a log-linear transformation:
\begin{equation}
\mathrm{ratio}_{\mathrm{ret}} = \mathrm{clip}(A \cdot \log(G_{\mathrm{safe}}) + B,\ 0,\ 1),
\end{equation}
where $A$ and $B$ are fixed constants, and $\mathrm{clip}(\cdot, 0, 1)$ bounds the output to the valid range $[0,1]$. This ratio determines the fraction of chunks to retain in dense regions.

We choose a logarithmic schedule over a linear one for two reasons.
First, in low-data regimes, estimates of $G(p)$ can be noisy, and a linear mapping may amplify this noise into overly aggressive pruning.
The logarithmic function grows smoothly and monotonically with $G(p)$, increasing pruning strength without abrupt jumps.
Second, in early experiments, we observed that linear rules behaved similarly under low redundancy but became unstable when $G(p)$ was high, often discarding too many samples.
In contrast, the log-based rule provides a more stable control signal across redundancy regimes while remaining simple and tuning-free.

The coefficients $A=0.15$ and $B=0.55$ are fixed across all experiments and are selected via regression on cross-dataset pruning behavior, capturing the general shape of the empirically optimal retention curve.
While more flexible non-parametric schedules could offer finer control, our goal here is to provide a robust and broadly applicable mechanism that generalizes across heterogeneous datasets.

\begin{algorithm}[H]
\caption{Dual-Granularity Filtering}
\label{alg:stage2}
\begin{algorithmic}[1]
\REQUIRE Training data $\mathcal{D}_{\text{train}}$, Synthetic data $\mathcal{D}_{\text{syn}}$, Candidate chunk sizes $\mathcal{S}_{\text{cand}}$
\ENSURE Filtered augmented dataset $\mathcal{D}_{\text{aug}}^*$

\STATE \algcomment{Phase 1: Proxy-Based Redundancy Estimation}
\FOR{$x_i \in \mathcal{D}_{\text{syn}}$}
    \STATE Find nearest seed $s_j^* = \arg\min_{s_j \in \mathcal{D}_{\text{train}}} \mathrm{DCR}(x_i, s_j)$
\ENDFOR
\STATE Construct proxy distribution $p=(p_1, \dots, p_n)$ and\\
global redundancy $G(p)$ 

\STATE \algcomment{Phase 2: Density-Based Partitioning}
\STATE Sort $p$ to find permutation $\pi$ where $p_{\pi(1)} \ge \dots \ge p_{\pi(n)}$
\STATE Find smallest $K$ such that $\sum_{j=1}^K p_{\pi(j)} \ge G(p)$
\STATE Partition $\mathcal{D}_{\text{syn}}$ into $\mathcal{D}_{\text{high}}$ and $\mathcal{D}_{\text{low}}$ (rest)
\STATE \algcomment{Phase 3: Dual-Granularity Filtering}
\STATE Initialize $\mathrm{min\_surprisal} \leftarrow \infty$, $\mathcal{D}_{\text{aug}}^* \leftarrow \emptyset$
\FOR{each chunk size $S \in \mathcal{S}_{\text{cand}}$}
    \STATE \algcomment{Chunk-level Filtering on $\mathcal{D}_{\text{high}}$}
    \STATE Partition $\mathcal{D}_{\text{high}}$ into chunks $\{\mathcal{C}_r\}$ of size $S$
    \STATE Set $G_{\text{safe}} \leftarrow \max\!\big(G(p), 10^{-3}\big)$
    \STATE Set $\mathrm{ratio}_{\text{ret}} \leftarrow \min\!\left(1,\ \max\!\left(0,\ A\log(G_{\text{safe}})+B\right)\right)$
    \STATE Set $N \leftarrow \left\lceil \frac{|\mathcal{D}_{\text{high}}|}{S} \right\rceil$ \algcomment{number of chunks}
    \STATE Set $K \leftarrow \left\lfloor \mathrm{ratio}_{\text{ret}} \cdot N \right\rfloor$
    \STATE Sort chunks $\{\mathcal{C}_r\}_{r=1}^{R}$ by $\mathrm{Score}(\mathcal{C}_r)$ descending
    \STATE $\mathcal{D}_{\text{high\_filtered}} \leftarrow \bigcup_{r=1}^{K} \mathcal{C}_r$


    
    \STATE \algcomment{Instance-level Filtering on $\mathcal{D}_{\text{low}}$}
    \STATE $\mathcal{D}_{\text{low\_filtered}} \leftarrow \{x \in \mathcal{D}_{\text{low}} \mid \mathrm{Conf}(x) \ge \mathrm{Conf}_{\text{thresh}} \land \mathrm{Uncert}(x) \le \mathrm{Uncert}_{\text{thresh}}\}$
    
    \STATE \algcomment{Surprisal Evaluation}
    \STATE $\mathcal{D}_{\text{aug}}(S) \leftarrow \mathcal{D}_{\text{high\_filtered}} \cup \mathcal{D}_{\text{low\_filtered}}$
    \STATE Train reference model ${M}$ on $\mathcal{D}_{\text{aug}}(S)$ and calculate Surprisal $L(S)$ on $\mathcal{D}_{\text{train}}$ 
    \IF{$L(S) < \mathrm{min\_surprisal}$}
        \STATE $\mathrm{min\_surprisal} \leftarrow L(S)$
        \STATE $\mathcal{D}_{\text{aug}}^* \leftarrow \mathcal{D}_{\text{aug}}(S)$
    \ENDIF
\ENDFOR
\STATE \textbf{return} $\mathcal{D}_{\text{aug}}^*$
\end{algorithmic}
\end{algorithm}

To further assess the empirical behavior and robustness of this redundancy-driven schedule, we conduct two supporting analyses below.
Section~1.3 examines the relationship between $G(p)$, the resulting pruning ratio, and downstream performance.
Section~1.4 studies the stability of $G(p)$ under varying synthetic data generation sizes, validating its role as a reliable control signal.

\begin{figure}[h]
    \centering
    \includegraphics[width=\columnwidth]{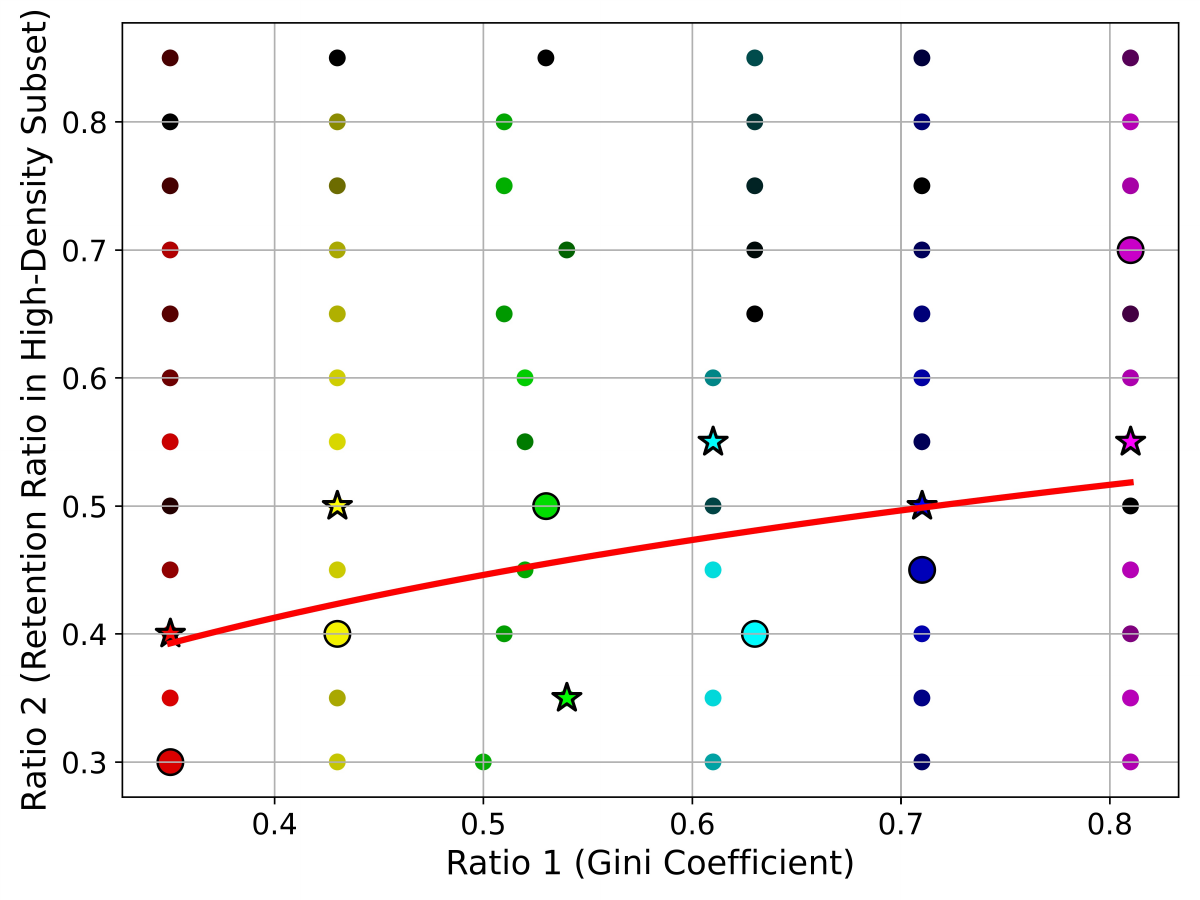}
    \caption{
    Scatter plot of \textit{$G(p)$} (Gini coefficient) versus \textit{$ratio_{ret}$}.
    Point color indicates downstream model performance after filtering, with lighter colors representing higher performance and darker colors indicating lower performance.
    In each group, the star (\(\star\)) marks the best-performing point and the circle (\(\circ\)) marks the second-best.
    }
    \label{fig:ratio_gini}
\end{figure}

\paragraph{1.3) Empirical behavior and robustness.}
Using the log-scaled mapping in Equation~\eqref{eq:ratio2}, we vary $G(p)$ across its empirical range and record the resulting pruning ratio as well as downstream performance (Figure~\ref{fig:ratio_gini}). The results show that performance peaks at intermediate retention: the Gini-driven schedule prunes aggressively only when redundancy is severe, while preserving diversity in more balanced settings. We also observe that $G(p)$ values across datasets tend to lie in a relatively narrow and stable range, which contributes to the robustness of the fitted schedule. Accordingly, the coefficients $(A,B)$, obtained via cross-dataset regression, generalize well without additional tuning.

\begin{figure}[t]
    \centering
    \includegraphics[width=\columnwidth]{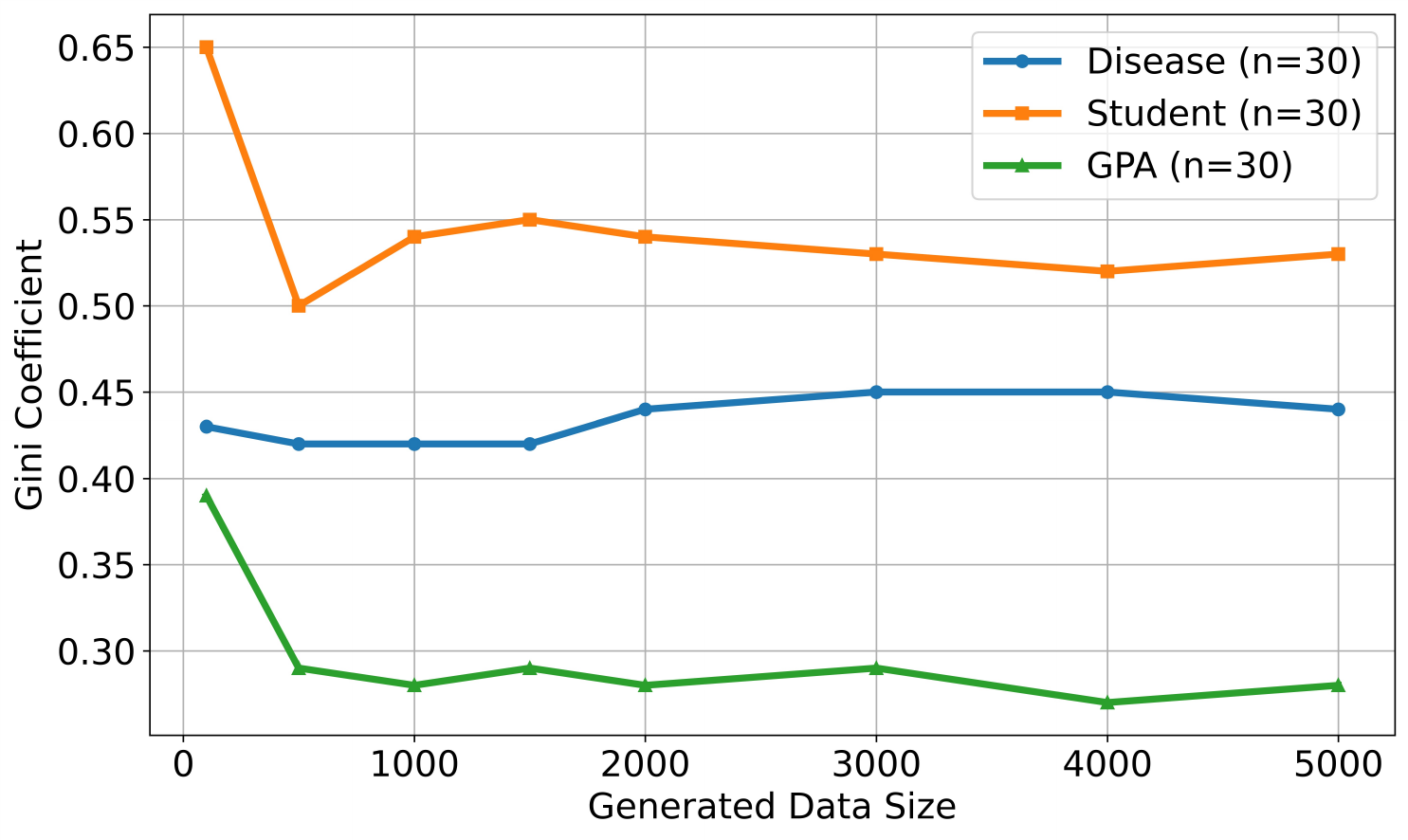}
    \caption{
    Gini coefficient under different synthetic data sizes, illustrating how data coverage varies as the amount of generated data increases.
    }
    \label{fig:gini_coverage}
\end{figure}

\paragraph{1.4) Stability under varying generation sizes.}
We further test whether $G(p)$ is sensitive to the number of generated samples, since an unstable control signal would undermine redundancy-adaptive filtering. Figure~\ref{fig:gini_coverage} shows that the Gini coefficient stabilizes after roughly 1{,}000 generated samples across tasks, with only minor fluctuations thereafter. This suggests that dominant distributional modes emerge early during generation and that $G(p)$ provides a consistent control signal as $|\mathcal{D}_{\mathrm{syn}}|$ increases.

\subsubsection{2) Instance-Level Filtering in Low-Density Subset.}
\label{app:com2.2}
Samples in the low-density subset \(\mathcal{D}_{\mathrm{low}}\) tend to reflect under-sampled regions of the feature space, which are more likely to contain useful variation but also noisy or unreliable outputs. To retain informative samples and discard potential outliers, we apply instance-level filtering guided by model confidence and predictive uncertainty.

For each sample \((x, y) \in \mathcal{D}_{\mathrm{low}}\), we compute two quantities based on the reference model \(\mathcal{M}\): the average confidence \(\mathrm{Conf}(x)\) and the average predictive uncertainty \(\mathrm{Uncert}(x)\), both measured across training epochs.
To make the filtering thresholds adaptive to data characteristics, we modulate them using $G(p)$, such that higher redundancy yields looser acceptance criteria:
\begin{equation}
\label{eq:ratio_thresholds}
\begin{aligned}
\mathrm{Conf}_{\text{thresh}} &= \mu_{\text{conf}} - G(p) \cdot \sigma_{\text{conf}}, \\
\mathrm{Uncert}_{\text{thresh}} &= \mu_{\text{uncert}} + G(p) \cdot \sigma_{\text{uncert}},
\end{aligned}
\end{equation}
where \(\mu\) and \(\sigma\) denote the empirical mean and standard deviation over \(\mathcal{D}_{\mathrm{low}}\). A higher $G(p)$ yields more relaxed thresholds, allowing retention of uncertain but potentially informative samples; a lower $G(p)$ results in stricter filtering to suppress noise.
A sample is retained if it satisfies both of the following conditions:
\[
\mathrm{Conf}(x) \ge \mathrm{Conf}_{\text{thresh}} 
\quad \text{and} \quad 
\mathrm{Uncert}(x) \le \mathrm{Uncert}_{\text{thresh}}.
\]
By coupling ${ratio}_1$ with two instance-level quantities, this mechanism effectively adapts filtering criteria across datasets with varying structural sparsity.

\subsubsection{3) Joint tuning via Surprisal.}
The only remaining hyperparameter is the chunk size \(S\), which determines the granularity of chunk-level evaluation in \(\mathcal{D}_{\mathrm{high}}\). 
To tune the chunk size $S$, we evaluate the quality of filtered synthetic data by measuring how well a downstream model performs when trained on it. Specifically, we train an evaluation model $\mathcal{M}_{\text{eval}}$ on $\mathcal{D}_{\text{train}} \cup \mathcal{D}_{\text{syn}}^{(S)}$ and select $S$ that minimizes its negative log-likelihood on real data:
\begin{equation}
S^\ast = \arg\min_S \left[ -\frac{1}{|\mathcal{D}_{\text{train}}|} \sum_i \log \mathbb{P}_{\mathcal{M}_{\text{eval}}}(y_i \mid x_i) \right].
\end{equation}

Here, $\mathcal{M}_{\text{cur}}$ and $\mathcal{M}_{\text{eval}}$ play distinct roles: the former is fixed and used for filtering, while the latter is retrained to reflect the quality of the filtered data. This separation ensures that $S$ is selected based on downstream generalization, not circular signals.

\section{Extended Experiment on Different Evaluator}
\label{app:exp_main}

\begin{table*}[htbp]
\scriptsize
\centering
\caption{Comparison between Best Non-LLM baselines, LLM-based generators, and \textsc{TabPFNGen}. 
We report \textbf{$F_{1}$ score} for classification tasks and \textbf{R\textsuperscript{2}} for regression tasks. 
\colorbox{blue!10}{Unseen datasets} are highlighted in blue, and \colorbox{red!10}{seen datasets} are highlighted in red.  
All synthetic data are evaluated using the same \textbf{XGBoost} downstream model. 
All values are reported as \textbf{mean ± standard deviation} (computed over multiple runs).  
The best result in each row is shown in \textbf{bold}, and the second-best result is \underline{underlined}.
}
\label{tab:tabpfngen}
\begin{adjustbox}{max width=\textwidth}
\begin{tabularx}{\textwidth}{c|Y|Y|Y|Y|Y|Y|Y|Y|Y}
\toprule
\multicolumn{1}{c|}{\textbf{Datasets}} &
\textbf{Real} &
\textsc{DRL} &
\textsc{TabSyn} &
\textsc{EPIC} &
\textsc{CLLM} &
I \(\setminus\) II &
II \(\setminus\) I &
I+II &
\textsc{TabPFNGen} \\
\midrule
\multicolumn{9}{c}{\textbf{N = 30}} \\
\midrule
\cellcolor{blue!10}Disease  & 93.68$\pm$1.4 &39.49$\pm$0.31& 54.79$\pm$2.7 & 32.01$\pm$8.2 & 61.89$\pm$2.1 & 59.61$\pm$1.5 & \underline{64.44$\pm$1.7} & \textbf{70.22$\pm$0.83} & 61.38$\pm$0.45 \\
\cellcolor{blue!10}Game     & 86.0$\pm$0.73 &33.54$\pm$6.33& 44.13$\pm$1.4 & 13.93$\pm$1.2 & 54.12$\pm$2.7 & 56.44$\pm$2.2 & 45.00$\pm$1.3 & \underline{59.13$\pm$2.8} & \textbf{70.43$\pm$1.81} \\
\cellcolor{blue!10}GPA      & 47.29$\pm$0.90 &14.46$\pm$1.54& 19.22$\pm$2.5 & 32.03$\pm$2.8 & 40.17$\pm$0.46 & 41.21$\pm$0.97 & 35.88$\pm$1.7 & \textbf{43.14$\pm$0.59} & \underline{41.40$\pm$0.92} \\
\cellcolor{blue!10}Student  & 0.67$\pm$0.07 &-0.01$\pm$0.01& 0.14$\pm$0.04 & 0.35$\pm$0.05 & \tiny -0.11$\pm$0.12 & 0.37$\pm$0.02 & 0.08$\pm$0.06 & \textbf{0.38$\pm$0.02} & \underline{0.37$\pm$0.01} \\
\cellcolor{red!10}Adult & 76.92$\pm$0.68 & 46.59$\pm$2.44 & 62.83$\pm$3.2  & 61.94$\pm$3.3 & 73.11$\pm$0.65 & 73.04$\pm$0.59 & \underline{73.15$\pm$0.18} & \textbf{73.91$\pm$0.37}& 63.88$\pm$5.35 \\
\cellcolor{red!10}Heart  & 86.71$\pm$1.7  & 38.66$\pm$2.29 & 79.40$\pm$1.3  & \textbf{81.20$\pm$1.3} & \underline{80.47$\pm$0.84} & 75.17$\pm$4.7 & 80.00$\pm$0.24 & 80.14$\pm$0.33& 61.42$\pm$2.00 \\

\midrule
\textbf{Avg Rank} & - & 7.5 & 5.8 & 5.7 & 4.0 & 3.9 & 4.0 & \textbf{1.5} & 3.6 \\
\midrule
\multicolumn{9}{c}{\textbf{N = 60}} \\
\midrule
\cellcolor{blue!10}Disease & 93.68$\pm$1.4 &38.25$\pm$0.22& 65.34$\pm$9.3 & 44.05$\pm$7.8 & 66.86$\pm$1.1 & \textbf{75.65$\pm$2.5} & 62.58$\pm$2.2 & \underline{72.41$\pm$0.77} & 71.88$\pm$0.62 \\
\cellcolor{blue!10}Game    & 86.0$\pm$0.73 &30.10$\pm$1.07& 61.10$\pm$1.1 & 13.16$\pm$0 & 67.54$\pm$1.0 & 61.39$\pm$2.6 & 59.81$\pm$1.1 & \underline{70.87$\pm$0.42} & \textbf{78.28$\pm$0.99} \\
\cellcolor{blue!10}GPA     & 47.29$\pm$0.90 &15.64$\pm$1.88& 28.48$\pm$1.8 & 32.19$\pm$3.5 & 33.17$\pm$1.1 & 44.34$\pm$0.91 & 21.57$\pm$1.8 & \textbf{44.58$\pm$0.56} & \underline{42.81$\pm$0.61} \\
\cellcolor{blue!10}Student  & 0.67$\pm$0.07 &-0.04$\pm$0.08& \tiny -0.14$\pm$0.14 & \tiny -0.63$\pm$0.19 & \tiny -0.48$\pm$0.19 & 0.27$\pm$0.03 & \tiny -0.15$\pm$0.04 & \underline{0.34$\pm$0.07} & \textbf{0.46$\pm$0.01} \\
\cellcolor{red!10}Adult    & 76.92$\pm$0.68 & 49.02$\pm$1.34 & 63.58$\pm$1.6  & 62.78$\pm$1.5 & \textbf{73.94$\pm$0.54} & \underline{73.28$\pm$0.85 }& 72.85$\pm$0.50 & 71.48$\pm$0.60& 64.69$\pm$3.40 \\
\cellcolor{red!10}Heart    & 86.71$\pm$1.7  & 38.14$\pm$2.85 & 81.35$\pm$1.2 & 77.55$\pm$1.6 & 80.33$\pm$0.85 & \textbf{81.37$\pm$0.96} & 78.15$\pm$0.53 &  \underline{80.36$\pm$0.03}& 63.98$\pm$1.78  \\
\midrule
\textbf{Avg Rank} & - & 7.2 & 4.8 & 6.8 & 3.8 & \textbf{2.3} & 5.5 & 2.5 & 3.0 \\
\midrule
\multicolumn{9}{c}{\textbf{N = 90}} \\
\midrule
\cellcolor{blue!10}Disease  & 93.68$\pm$1.4 &39.04$\pm$0.00& 69.90$\pm$2.0 & 30.03$\pm$3.2 & 74.04$\pm$1.6 & 69.74$\pm$1.5 & 65.47$\pm$1.2 & \textbf{76.45$\pm$1.1} & \underline{75.55$\pm$0.36} \\
\cellcolor{blue!10}Game    & 86.0$\pm$0.73 &27.81$\pm$2.93& \textbf{65.93$\pm$1.2} & 13.16$\pm$0 & 59.97$\pm$2.3 & 59.44$\pm$2.7 & 41.77$\pm$1.6 & \underline{62.17$\pm$2.7} & 60.12$\pm$0.95 \\
\cellcolor{blue!10}GPA     & 47.29$\pm$0.90 &12.24$\pm$0.90& 34.51$\pm$1.3 & 16.28$\pm$1.98 & 41.79$\pm$1.1 & 47.21$\pm$1.3 & 31.56$\pm$0.87 & \underline{48.89$\pm$0.56} & \textbf{49.77$\pm$0.66} \\
\cellcolor{blue!10}Student & 0.67$\pm$0.07 &-0.03$\pm$0.05& \tiny -0.12$\pm$0.10 & 0.32$\pm$0.07 & 0.11$\pm$0.13 & 0.34$\pm$0.09 & 0.27$\pm$0.09 & \textbf{0.47$\pm$0.09} & \underline{0.44$\pm$0.05} \\
\cellcolor{red!10}Adult& 76.92$\pm$0.68 & 47.47$\pm$2.36 & 69.77$\pm$0.68 & 66.73$\pm$1.3 & \underline{74.11$\pm$1.36} & \textbf{78.45$\pm$0.97} & 73.75$\pm$0.51 & 74.10$\pm$0.62 & 62.98$\pm$5.38 \\
\cellcolor{red!10}Heart & 86.71$\pm$1.7 & 39.03$\pm$2.85 & 81.43$\pm$1.3 & 77.65$\pm$1.7 & 81.23$\pm$1.0 & \textbf{82.44$\pm$0.86} & 78.40$\pm$1.4 & 80.00$\pm$0.03 & 53.29$\pm$4.87 \\
\midrule
\textbf{Avg Rank} & - & 7.5 & 4.2 & 6.5 & 3.7 &3.0 & 5.3 & \textbf{2.2} & 3.7 \\
\midrule
\multicolumn{9}{c}{\textbf{N = 120}} \\
\midrule
\cellcolor{blue!10}Disease & 93.68$\pm$1.4 &73.92$\pm$3.30& 62.37$\pm$1.25 & 55.37$\pm$10.36 & 78.30$\pm$1.44 & \underline{78.97$\pm$3.13} & 61.82$\pm$2.57 & \textbf{81.96$\pm$0.64} & 78.85$\pm$0.40 \\
\cellcolor{blue!10}Game  & 86.0$\pm$0.73 &31.04$\pm$2.48& \underline{63.50$\pm$0.78} & 48.99$\pm$2.41 & 61.32$\pm$1.16 & 45.48$\pm$1.62 & 54.76$\pm$2.01 & 61.67$\pm$1.40 & \textbf{73.93$\pm$0.41} \\
\cellcolor{blue!10}GPA  & 47.29$\pm$0.90 &10.63$\pm$1.75& 38.57$\pm$2.30 & 46.61$\pm$3.7 & 45.20$\pm$2.1 & 40.42$\pm$1.02 & \underline{46.95$\pm$0.5} & \textbf{47.77$\pm$0.81} & 44.78$\pm$0.71 \\
\cellcolor{blue!10}Student & 0.67$\pm$0.07 &0.01$\pm$0.03& 0.41$\pm$0.03 & 0.29$\pm$0.07 & 0.22$\pm$0.03 & 0.21$\pm$0.04 & 0.17$\pm$0.06 & \underline{0.29$\pm$0.04} & \textbf{0.47$\pm$0.03} \\
\cellcolor{red!10}Adult  & 76.92$\pm$0.68 & 39.22$\pm$6.96 & 68.87$\pm$0.41  & 68.35$\pm$2.10 & 71.73$\pm$0.82 & \textbf{73.36$\pm$0.67} & \underline{73.19$\pm$0.65} & 72.93$\pm$0.74&68.42$\pm$1.04 \\
\cellcolor{red!10}Heart   & 86.71$\pm$1.7  & 38.29$\pm$0.31 & \underline{83.80$\pm$0.51}  & 71.33$\pm$6.33 & 75.24$\pm$0.72 & \textbf{84.97$\pm$1.08} & 75.02$\pm$0.97 & 75.93$\pm$0.84& 69.27$\pm$1.19 \\
\midrule
\textbf{Avg Rank} & - & 7.5 & 4.0 & 5.6 & 4.2 & 3.8 & 4.7 & \textbf{2.4} & 3.8 \\
\midrule
\multicolumn{9}{c}{\textbf{N = 160}} \\
\midrule
\cellcolor{blue!10}Disease & 93.68$\pm$1.4&68.20$\pm$9.36 & \underline{72.25$\pm$1.33} & 68.75$\pm$2.71 & 69.86$\pm$2.51 & 58.60$\pm$2.09 & 63.60$\pm$1.73 & 71.66$\pm$1.75 & \textbf{76.45$\pm$0.24} \\
\cellcolor{blue!10}Game   & 86.0$\pm$0.73 &27.37$\pm$4.16& \underline{69.91$\pm$0.64} & 49.69$\pm$4.99 & 48.18$\pm$3.40 & 39.78$\pm$1.33 & 54.45$\pm$1.50 & 62.79$\pm$1.13 & \textbf{74.08$\pm$0.35} \\
\cellcolor{blue!10}GPA   & 47.29$\pm$0.90 &12.70$\pm$1.30& 39.37$\pm$0.81 & 29.08$\pm$1.04 & 31.07$\pm$1.83 & 44.49$\pm$1.76 & 41.01$\pm$2.41 & \underline{44.65$\pm$0.63} & \textbf{45.62$\pm$0.75} \\
\cellcolor{blue!10}Student& 0.67$\pm$0.07&-0.01$\pm$0.01 & 0.32$\pm$0.07 & 0.43$\pm$0.10 & 0.22$\pm$0.07 & 0.30$\pm$0.17 & 0.37$\pm$0.12 & \textbf{0.46$\pm$0.13} & \underline{0.43$\pm$0.02} \\
\cellcolor{red!10}Adult & 76.92$\pm$0.68 & 45.37$\pm$4.44 & \underline{71.20$\pm$0.38} & 69.08$\pm$1.16 & \textbf{72.64$\pm$0.62} & 70.51$\pm$1.47 & 70.44$\pm$1.22 & 68.99$\pm$0.87 & 69.21$\pm$0.66 \\
\cellcolor{red!10}Heart & 86.71$\pm$1.7 & 34.11$\pm$7.39 & 67.85$\pm$2.18 & 73.17$\pm$3.32 & \underline{79.47$\pm$1.73} & 78.73$\pm$1.66 & \textbf{79.87$\pm$0.00} & 78.69$\pm$1.24 & 66.09$\pm$1.51 \\
\midrule
\textbf{Avg Rank } & - & 7.7 & 3.7 & 5.1 & 4.3 & 5.0 & 4.0 &3.3 & \textbf{2.9} \\
\midrule
\multicolumn{9}{c}{\textbf{N = 200}} \\
\midrule
\cellcolor{blue!10}Disease & 93.68$\pm$1.4 &51.08$\pm$1.71& 64.53$\pm$1.49 & 55.30$\pm$6.14 & 73.72$\pm$1.98 & 68.58$\pm$2.13 & 71.96$\pm$2.86 & \underline{74.25$\pm$1.38} & \textbf{78.07$\pm$0.84} \\
\cellcolor{blue!10}Game   & 86.0$\pm$0.73 &32.98$\pm$2.47& \textbf{77.92$\pm$0.69} & 55.19$\pm$6.76 & 51.19$\pm$2.06 & 44.68$\pm$3.78 & 61.56$\pm$0.68 & 61.98$\pm$1.52 & \underline{69.18$\pm$2.03} \\
\cellcolor{blue!10}GPA    & 47.29$\pm$0.90& 14.53$\pm$1.94& 41.59$\pm$0.86 & 40.88$\pm$0.94 & 36.40$\pm$3.18 & 43.92$\pm$1.42 & 38.73$\pm$0.00 & \underline{44.68$\pm$0.51} & \textbf{46.26$\pm$0.90} \\
\cellcolor{blue!10}Student & 0.67$\pm$0.07&-0.02$\pm$0.02 & 0.39$\pm$0.11 & \underline{0.51$\pm$0.04} & 0.24$\pm$0.06 & 0.50$\pm$0.08 & 0.20$\pm$0.07 & \textbf{0.52$\pm$0.02} & 0.50$\pm$0.01 \\
\cellcolor{red!10}Adult  & 76.92$\pm$0.68 & 46.19$\pm$6.68 & 69.43$\pm$0.71  & 70.95$\pm$1.34 & 73.40$\pm$0.75 & 72.42$\pm$0.88 & \underline{73.66$\pm$0.71} & \textbf{74.50$\pm$0.30}& 69.21$\pm$0.66 \\
\cellcolor{red!10}Heart   & 86.71$\pm$1.7  & 47.80$\pm$6.27 & 74.33$\pm$2.60  & 74.74$\pm$2.10 & 76.17$\pm$2.01 & \underline{79.95$\pm$0.11} & \textbf{80.11$\pm$0.91} & 77.84$\pm$1.01& 66.09$\pm$1.51 \\
\midrule
\textbf{Avg Rank} & - & 8.0 & 4.7 & 4.8 & 4.8 & 4.1 & 4.0 & \textbf{2.0} & 3.6 \\
\midrule
\multicolumn{9}{c}{\textbf{N = 300}} \\
\midrule
\cellcolor{blue!10}Disease & 93.68$\pm$1.4&67.48$\pm$5.59 & \underline{77.26$\pm$1.18} & 35.37$\pm$2.47 & 67.31$\pm$1.59 & 68.65$\pm$4.11 & 66.02$\pm$1.73 & 73.31$\pm$2.82 & \textbf{78.96$\pm$0.51} \\
\cellcolor{blue!10}Game   & 86.0$\pm$0.73&26.19$\pm$2.19 & \textbf{77.64$\pm$0.56} & 54.23$\pm$4.23 & 46.69$\pm$2.05 & 47.58$\pm$3.79 & 47.25$\pm$2.47 & 61.67$\pm$1.23 & \underline{71.90$\pm$1.61} \\
\cellcolor{blue!10}GPA     & 47.29$\pm$0.90& 15.25$\pm$1.62& 39.86$\pm$1.29 & 37.77$\pm$0.63 & 39.67$\pm$2.14 & 42.98$\pm$0.05 & 43.98$\pm$0.96 & \textbf{46.72$\pm$1.12} & \underline{46.09$\pm$1.44} \\
\cellcolor{blue!10}Student  & 0.67$\pm$0.07&-0.05$\pm$0.02 & \underline{0.54$\pm$0.01} & 0.44$\pm$0.03 & \tiny -0.03$\pm$0.06 & 0.50$\pm$0.05 & \tiny -0.17$\pm$0.06 & \textbf{0.55$\pm$0.01} & \underline{0.54$\pm$0.01} \\
\cellcolor{red!10}Adult& 76.92$\pm$0.68 & 54.08$\pm$3.99 & 70.53$\pm$0.79 & 70.98$\pm$1.05 & 74.09$\pm$0.56 & 73.07$\pm$0.62 & \textbf{74.57$\pm$1.12} & \underline{74.29$\pm$0.67} & 70.01$\pm$1.41 \\
\cellcolor{red!10}Heart  & 86.71$\pm$1.7 & 29.61$\pm$0.78 & 76.21$\pm$2.16 & \textbf{79.60$\pm$3.40} & 76.64$\pm$2.04 & 77.33$\pm$0.23 & \underline{78.33$\pm$0.50} & 76.66$\pm$0.81 & 66.49$\pm$1.87 \\
\midrule
\textbf{Avg Rank } & - & 7.3 & 3.8 & 5.0 & 5.5 & 4.0 & 4.5 & \textbf{2.3} & 3.6 \\
\midrule
\multicolumn{9}{c}{\textbf{N = 500}} \\
\midrule
\cellcolor{blue!10}Disease & 93.68$\pm$1.4& 75.27$\pm$2.39& 69.93$\pm$1.41 & 65.26$\pm$3.15 & 67.56$\pm$2.83 & 61.41$\pm$3.90 & 68.55$\pm$2.44 & \underline{72.48$\pm$1.97} & \textbf{78.52$\pm$0.26} \\
\cellcolor{blue!10}Game     & 86.0$\pm$0.73&25.69$\pm$3.03 & \textbf{80.33$\pm$1.26} & 61.38$\pm$5.18 & 53.88$\pm$1.77 & 69.37$\pm$2.33 & 65.33$\pm$1.19 & 75.95$\pm$1.57 & \underline{79.60$\pm$0.89} \\
\cellcolor{blue!10}GPA    & 47.29$\pm$0.90&7.57$\pm$1.44 & 40.17$\pm$1.20 & 36.35$\pm$0.95 & 30.14$\pm$2.30 & 37.90$\pm$1.94 & 35.93$\pm$0.98 & \underline{40.30$\pm$0.65} & \textbf{43.13$\pm$0.74} \\
\cellcolor{blue!10}Student & 0.67$\pm$0.07&-0.04$\pm$0.03 & \textbf{0.57$\pm$0.01} & 0.54$\pm$0.02 & \tiny -0.28$\pm$0.08 & 0.46$\pm$0.04 & 0.03$\pm$0.22 & 0.52$\pm$0.01 & \underline{0.56$\pm$0.01} \\
\cellcolor{red!10}Adult & 76.92$\pm$0.68 & 49.33$\pm$3.06 & 70.11$\pm$0.78 & 68.06$\pm$1.37 & \underline{73.33$\pm$0.97} & \textbf{74.11$\pm$0.32} & 71.15$\pm$0.77 & 70.11$\pm$0.31  & 71.02$\pm$0.92\\
\cellcolor{red!10}Heart  & 86.71$\pm$1.7  & 27.73$\pm$0.13 & 77.26$\pm$2.46  & \textbf{81.71$\pm$1.81} & 76.01$\pm$0.21 & 75.20$\pm$0.13 & 74.20$\pm$0.90 & \underline{80.31$\pm$1.17}& 63.92$\pm$2.90 \\
\midrule
\textbf{Avg Rank} & - & 6.8 & 2.9 & 4.8 & 5.7 & 4.5 & 5.2 & 3.3 & \textbf{2.8} \\
\midrule
\textbf{Total Rank} & - & 7.4 & 4.2 & 5.5 & 4.5 & 3.8 & 4.7 & \textbf{2.4} &3.4 \\

\bottomrule
\end{tabularx}
\end{adjustbox}
\end{table*}

\begin{table*}[t]
\scriptsize
\centering
\caption{Comparison between Non-LLM baselines, LLM-based generators, and \textsc{TabPFNGen}. 
All results are evaluated with the \textbf{\textsc{TabPFN}} downstream model and reported as mean$\pm$standard deviation.
\colorbox{blue!10}{Unseen datasets} are highlighted in blue, and \colorbox{red!10}{seen datasets} are highlighted in red.
The best result in each row is shown in \textbf{bold}, and the second-best result is \underline{underlined}.
}
\label{app:tabpfn}
\begin{adjustbox}{max width=\textwidth}
{\fontsize{6pt}{7pt}\selectfont
\begin{tabularx}{\textwidth}{c|Y|Y|Y|Y|Y|Y|Y|Y|Y}
\toprule
Datasets & Real &DRL& \textsc{TabSyn} & \textsc{EPIC} & \textsc{CLLM} & I(w/oII) & II(w/oI) & I+II & \textsc{TabPFNGen} \\
\midrule
\multicolumn{9}{c}{\textbf{N = 30}} \\
\midrule
\cellcolor{blue!10}Disease & 96.12$\pm$0.28 &38.90$\pm$0.25& 55.35$\pm$2.26 & 33.69$\pm$9.38 & \underline{68.09$\pm$1.89} & 55.50$\pm$0.40 & 61.34$\pm$1.78 & \textbf{69.45$\pm$0.23} & 56.75$\pm$0.83 \\
\cellcolor{blue!10}Game    & 90.11$\pm$0.31 &27.45$\pm$4.87& 38.31$\pm$2.11 & 37.33$\pm$5.61 & \underline{58.54$\pm$1.72} & 50.80$\pm$2.35 & 51.66$\pm$0.00 & 51.36$\pm$0.81 & \textbf{69.18$\pm$2.03} \\
\cellcolor{blue!10}GPA     & 71.07$\pm$1.48 &4.25$\pm$0.52& 16.37$\pm$1.84 & 30.38$\pm$2.28 & 40.26$\pm$0.50 & \underline{41.84$\pm$1.27} & 36.55$\pm$0.00 & \textbf{42.33$\pm$0.64} & 39.12$\pm$0.39 \\
\cellcolor{blue!10}Student  & 0.73$\pm$0.00  &-0.04$\pm$0.02& 0.18$\pm$0.05 & 0.28$\pm$0.04 & -0.26$\pm$0.14 & 0.40$\pm$0.02 & 0.15$\pm$0.00 & \underline{0.44$\pm$0.01} & \textbf{0.54$\pm$0.01} \\
\cellcolor{red!10}Adult  &78.47$\pm$0.35   &43.16$\pm$0.10& 61.01$\pm$1.76 & 52.02$\pm$2.36& 75.90$\pm$0.52 & \underline{75.98$\pm$0.58} & 75.52$\pm$0.27 & \textbf{76.54$\pm$0.17} & 54.64$\pm$0.66\\
\cellcolor{red!10}Heart  &95.00$\pm$0.60   &36.80$\pm$0.00& 79.05$\pm$2.00 & 67.66$\pm$4.01 & 72.96$\pm$1.10 & \underline{ 80.79$\pm$2.40} & 77.36$\pm$0.99 & \textbf{82.13$\pm$0.72} & 70.38$\pm$0.83 \\
\midrule
\textbf{Avg Rank} & - & 7.7 & 5.3 & 6.5 & 3.8 & 3.2 & 4.2 & \textbf{1.7} & 3.7 \\
\midrule
\multicolumn{9}{c}{\textbf{N = 60}} \\
\midrule
\cellcolor{blue!10}Disease & 96.12$\pm$0.28 &38.30$\pm$0.11& 62.56$\pm$0.94 & 27.33$\pm$1.53 & \textbf{70.84$\pm$2.19} & 62.66$\pm$1.61 & 63.70$\pm$0.65 & 68.50$\pm$1.00 & \underline{69.22$\pm$1.27} \\
\cellcolor{blue!10}Game   & 90.11$\pm$0.31 &22.13$\pm$0.63& 57.03$\pm$1.76 & 50.84$\pm$4.16 & \underline{65.79$\pm$1.49} & 59.88$\pm$2.70 & 54.43$\pm$0.95 & 63.25$\pm$0.00 & \textbf{75.87$\pm$1.18} \\
\cellcolor{blue!10}GPA     & 71.07$\pm$1.48&7.46$\pm$1.92 & 29.09$\pm$0.85 & 38.26$\pm$1.50 & 36.20$\pm$2.81 & 45.12$\pm$0.64 & 27.30$\pm$0.00 & \textbf{48.89$\pm$0.25} & \underline{47.45$\pm$0.87} \\
\cellcolor{blue!10}Student & 0.73$\pm$0.00 &-0.09$\pm$0.04 & 0.11$\pm$0.12 & 0.29$\pm$0.05 & -0.48$\pm$0.11 & 0.27$\pm$0.05 & -0.03$\pm$0.04 & \underline{0.35$\pm$0.02} & \textbf{0.54$\pm$0.01} \\
\cellcolor{red!10}Adult  &  77.89$\pm$0.96 &51.39$\pm$0.83& 64.00$\pm$0.62 & 57.71$\pm$4.20& \textbf{76.57$\pm$0.25} & 74.08$\pm$0.99 & \underline{76.28$\pm$0.19} & 74.13$\pm$0.54 & 63.87$\pm$0.49\\
\cellcolor{red!10}Heart  &   94.97$\pm$0.82&36.94$\pm$0.00& \textbf{82.62$\pm$0.55} & 74.45$\pm$2.70 & 75.01$\pm$1.61 & 79.12$\pm$1.88 & 77.36$\pm$0.99 & \underline{79.92$\pm$0.62} & 69.87$\pm$0.82 \\
\midrule
\textbf{Avg Rank} & - & 7.7 & 4.7 & 5.8 & 3.7 & 3.8 & 4.8 & \textbf{2.3} & 3.2 \\
\midrule
\multicolumn{9}{c}{\textbf{N = 90}} \\
\midrule
\cellcolor{blue!10}Disease & 96.12$\pm$0.28&39.04$\pm$ 0.00 & 71.90$\pm$1.24 & 28.39$\pm$2.57 & \textbf{77.01$\pm$1.10} & 54.44$\pm$2.97 & 67.25$\pm$2.29 & 62.84$\pm$1.06 & \underline{73.04$\pm$2.0} \\
\cellcolor{blue!10}Game   & 90.11$\pm$0.31 &22.13$\pm$0.63& 39.48$\pm$1.05 & 54.31$\pm$2.00 & 66.10$\pm$2.21 & 55.64$\pm$2.01 & 48.52$\pm$1.77 & \underline{67.46$\pm$2.31} & \textbf{72.83$\pm$0.63} \\
\cellcolor{blue!10}GPA    & 71.07$\pm$1.48 &5.93$\pm$0.11& 35.49$\pm$1.54 & 14.58$\pm$1.92 & 46.62$\pm$0.79 & 48.48$\pm$1.55 & 37.89$\pm$1.93 & \underline{49.49$\pm$0.55} & \textbf{52.33$\pm$0.43} \\
\cellcolor{blue!10}Student  & 0.73$\pm$0.00&0.05$\pm$0.08  & 0.21$\pm$0.09 & 0.38$\pm$0.05 & -0.27$\pm$0.15 & 0.48$\pm$0.03 & 0.14$\pm$0.04 & \textbf{0.53$\pm$0.01} & \textbf{0.53$\pm$0.01}\\
\cellcolor{red!10}Adult  &77.97$\pm$0.64   & 43.18$\pm$0.00 & 69.18 $\pm$1.00 & 55.94 $\pm$ 3.62 & 74.73 $\pm$ 0.33 & \underline{75.91 $\pm$ 0.20} & 74.28 $\pm$ 0.34 & \textbf{76.55 $\pm$ 0.31} & 59.63 $\pm$ 0.62 \\
\cellcolor{red!10}Heart  & 95.59$\pm$0.34  & NULL & 78.34 $\pm$ 0.00 & 69.22 $\pm$ 2.50 & 78.25 $\pm$ 1.08 & \underline{80.69 $\pm$1.04} & 77.41 $\pm$ 0.46 & \textbf{80.72 $\pm$ 1.00} & 71.25 $\pm$ 0.71 \\
\midrule
\textbf{Avg Rank} & - & 7.7 & 4.8 & 6.3 & 3.8 & 3.3 & 5.0 & \textbf{2.1} & 2.9 \\
\midrule
\multicolumn{9}{c}{\textbf{N = 120}} \\
\midrule
\cellcolor{blue!10}Disease & 96.12$\pm$0.28&68.40$\pm$5.71 & 64.04$\pm$1.55 & 31.89$\pm$9.27 & \textbf{71.33$\pm$1.31} & 62.33$\pm$4.82 & 63.08$\pm$2.65 & 64.65$\pm$1.95 & \underline{68.87$\pm$2.74} \\
\cellcolor{blue!10}Game   & 90.11$\pm$0.31&23.60$\pm$2.34 & 57.47$\pm$0.96 & 53.76$\pm$1.43 & \underline{74.16$\pm$0.78} & 60.56$\pm$2.40 & 62.54$\pm$0.00 & 69.62$\pm$0.70 & \textbf{74.81$\pm$0.57} \\
\cellcolor{blue!10}GPA     & 71.07$\pm$1.48 &5.63$\pm$0.27& 37.13$\pm$4.26 & 44.32$\pm$1.83 & \underline{47.43$\pm$2.11} & 42.36$\pm$1.23 & 38.67$\pm$2.03 & 41.06$\pm$1.76 & \textbf{57.26$\pm$1.10} \\
\cellcolor{blue!10}Student & 0.73$\pm$0.00&0.0$\pm$0.01  & 0.05$\pm$0.08 & \underline{0.54$\pm$0.02} & -0.50$\pm$0.16 & 0.46$\pm$0.07 & 0.04$\pm$0.03 & \textbf{0.56$\pm$0.02} & 0.53$\pm$0.01 \\
\cellcolor{red!10}Adult  &   77.70$\pm$0.73& 42.33$\pm$1.23 & 68.23 $\pm$0.32 & 64.92 $\pm$ 1.66 & \underline{75.44 $\pm$ 0.52} & 71.14 $\pm$ 1.37 & 75.02 $\pm$ 0.34 & \textbf{76.29 $\pm$ 0.89} & 62.28 $\pm$ 0.70 \\
\cellcolor{red!10}Heart  & 95.17$\pm$0.59  & 38.14 $\pm$ 0.00 & 81.54 $\pm$ 0.00 & \underline{82.33 $\pm$ 0.79} & 71.61 $\pm$ 1.73 & 81.63 $\pm$1.57 & 76.54 $\pm$ 0.48 & \textbf{83.84 $\pm$ 0.81} & 75.56 $\pm$ 0.62 \\
\midrule
\textbf{Avg Rank} & - & 6.8 & 5.3 & 4.7 & 3.7 & 4.5 & 5.1 & \textbf{2.5} & 3.5 \\
\midrule
\multicolumn{9}{c}{\textbf{N = 160}} \\
\midrule
\cellcolor{blue!10}Disease & 96.12$\pm$0.28&68.40$\pm$5.71 & \textbf{74.48$\pm$0.73} & 68.65$\pm$3.92 & 68.76$\pm$3.19 & 58.26$\pm$2.59 & 66.83$\pm$1.18 & 69.35$\pm$1.40 & \underline{73.16$\pm$0.25} \\
\cellcolor{blue!10}Game    & 90.11$\pm$0.31&23.60$\pm$2.34 & 64.91$\pm$1.07 & 52.97$\pm$6.99 & \underline{66.82}$\pm$3.05 & 51.38$\pm$4.03 & 53.43$\pm$2.10 & 58.69$\pm$0.92 & \textbf{74.34$\pm$0.75} \\
\cellcolor{blue!10}GPA   & 71.07$\pm$1.48 &5.63$\pm$0.27& 34.39$\pm$0.84 & 44.14$\pm$3.00 & \underline{45.72$\pm$0.44} & 50.60$\pm$0.64 & 42.47$\pm$2.13 & \textbf{50.97$\pm$0.17} & 41.90$\pm$0.56 \\
\cellcolor{blue!10}Student  & 0.73$\pm$0.00&0.0$\pm$0.01  & 0.36$\pm$0.09 & 0.51$\pm$0.03 & -1.06$\pm$0.15 & \underline{0.56$\pm$0.02} & -0.86$\pm$0.04 & \textbf{0.56$\pm$0.01} & 0.50$\pm$0.02 \\
\cellcolor{red!10}Adult &77.67$\pm$0.55 & 43.45$\pm$0.50 & 71.37$\pm$0.75 & 68.13$\pm$0.74 & 73.73$\pm$0.53 & \underline{75.51$\pm$0.46} & 73.11$\pm$0.36 & \textbf{76.13$\pm$0.19} & 62.69$\pm$0.66 \\
\cellcolor{red!10}Heart &96.07$\pm$0.66 & 31.00$\pm$0.25 & 70.91$\pm$1.39 & 79.46$\pm$1.87 & 78.03$\pm$1.03 & \underline{78.78$\pm$1.90} & 78.51$\pm$0.41 & \textbf{83.56$\pm$0.78} & 71.58$\pm$0.59 \\
\midrule
\textbf{Avg Rank} & - & 7.3 & 4.7 & 4.3 & 4.2 & 3.9 & 5.3 & \textbf{1.9} & 4.3 \\
\midrule
\multicolumn{9}{c}{\textbf{N = 200}} \\
\midrule
\cellcolor{blue!10}Disease & 96.12$\pm$0.28&51.43$\pm$0.26 & 68.05$\pm$1.15 & 62.53$\pm$5.31 & 69.56$\pm$1.75 & 69.97$\pm$2.37 & 71.14$\pm$0.93 & \underline{71.42$\pm$1.02} & \textbf{73.37$\pm$1.46} \\
\cellcolor{blue!10}Game   & 90.11$\pm$0.31 &26.35$\pm$1.96& 64.73$\pm$6.83 & 55.00$\pm$0.75 & \underline{72.15$\pm$1.45} & 49.36$\pm$2.42 & 64.08$\pm$1.33 & 65.07$\pm$2.17 & \textbf{85.32$\pm$0.51} \\
\cellcolor{blue!10}GPA    & 71.07$\pm$1.48&6.35$\pm$1.51 & 37.51$\pm$1.29 & 43.25$\pm$0.90 & 45.75$\pm$0.44 & \underline{47.37$\pm$0.94} & 41.27$\pm$2.75 & \textbf{48.80$\pm$0.58} & 42.32$\pm$0.67 \\
\cellcolor{blue!10}Student & 0.73$\pm$0.00 & -0.037$\pm$0.022  & 0.48$\pm$0.03 & 0.47$\pm$0.11 & 0.05$\pm$0.09 & \underline{0.53$\pm$0.01} & 0.11$\pm$0.00 & \textbf{0.55$\pm$0.02} & \underline{0.53$\pm$0.01} \\
\cellcolor{red!10}Adult & 77.80$\pm$0.67& 45.42$\pm$4.32 & 67.69$\pm$0.86 & 69.42$\pm$1.36 & 74.77$\pm$0.62 & 73.76$\pm$0.62 & \underline{74.87$\pm$0.42} & \textbf{77.28$\pm$0.66} & 63.17$\pm$0.57 \\
\cellcolor{red!10}Heart &96.07$\pm$0.66 & 54.50$\pm$7.18 & 73.01$\pm$1.69 & 80.09$\pm$2.65 & 79.10$\pm$1.31 & 79.68$\pm$0.64 & \underline{81.79$\pm$0.47} & \textbf{83.98$\pm$0.80} & 74.91$\pm$0.63 \\
\midrule
\textbf{Avg Rank} & - & 8.0 & 5.7 & 5.0 & 4.2 & 3.9 & 4.0 & \textbf{1.5} & 3.8 \\
\midrule
\multicolumn{9}{c}{\textbf{N = 300}} \\
\midrule
\cellcolor{blue!10}Disease & 96.12$\pm$0.28&70.78$\pm$1.93 & \textbf{77.54$\pm$0.64} & 36.97$\pm$2.97 & 69.55$\pm$3.10 & 63.23$\pm$4.46 & 63.81$\pm$1.14 & 71.46$\pm$3.67 & \underline{75.89$\pm$0.58} \\
\cellcolor{blue!10}Game   & 90.11$\pm$0.31 &24.58$\pm$1.66& \underline{78.80$\pm$0.97} & 61.16$\pm$6.76 & 66.01$\pm$1.25 & 51.63$\pm$1.22 & 56.64$\pm$1.64 & 65.03$\pm$1.18 & \textbf{86.15$\pm$0.69} \\
\cellcolor{blue!10}GPA    & 71.07$\pm$1.48 &4.15$\pm$2.12& 38.52$\pm$0.69 & 39.44$\pm$1.13 & 44.72$\pm$0.98 & 43.71$\pm$0.74 & \underline{46.75$\pm$1.42} & \textbf{47.10$\pm$0.26} & 44.84$\pm$0.63 \\
\cellcolor{blue!10}Student & 0.73$\pm$0.00  &-0.06$\pm$0.02& 0.48$\pm$0.03 & 0.47$\pm$0.11 & 0.04$\pm$0.10 & 0.53$\pm$0.01 & 0.19$\pm$0.04 & \textbf{0.55$\pm$0.01} & \underline{0.54$\pm$0.01} \\
\cellcolor{red!10}Adult & 77.83$\pm$0.88& 50.72$\pm$3.19 & 71.39$\pm$0.87 & 70.55$\pm$0.77 & 74.94$\pm$0.27 & \underline{76.17$\pm$0.61} & 73.25$\pm$0.16 & \textbf{77.18$\pm$0.45} & 61.90$\pm$0.64 \\
\cellcolor{red!10}Heart &97.11$\pm$0.60 & 29.61$\pm$0.00 & 75.78$\pm$1.81 & \textbf{81.11$\pm$1.34} & 76.72$\pm$1.40 & 75.24$\pm$0.71 & 77.62$\pm$0.00 & \underline{80.22$\pm$0.19} & 74.08$\pm$1.24 \\
\midrule
\textbf{Avg Rank} & - & 7.3 & 4.0 & 5.2 & 4.2 & 5.0 & 4.5 & \textbf{2.0} & 3.8 \\
\midrule
\multicolumn{9}{c}{\textbf{N = 500}} \\
\midrule
\cellcolor{blue!10}Disease & 96.12$\pm$0.28 &74.08$\pm$1.30& 74.57$\pm$0.91 & \underline{75.00$\pm$1.47} & 68.03$\pm$1.73 & 66.29$\pm$1.56 & 64.74$\pm$0.99 & 68.79$\pm$1.04 & \textbf{77.32$\pm$0.73} \\
\cellcolor{blue!10}Game    & 90.11$\pm$0.31&13.63$\pm$1.21 & \underline{80.55$\pm$0.73} & 63.73$\pm$6.80 & 80.08$\pm$1.50 & 70.13$\pm$4.25 & 73.50$\pm$2.38 & 78.44$\pm$0.25 & \textbf{88.33$\pm$0.31} \\
\cellcolor{blue!10}GPA    & 71.07$\pm$1.48&3.00$\pm$0.79 & 36.60$\pm$0.96 & \textbf{46.48$\pm$1.01} & \underline{45.28$\pm$0.63} & 40.00$\pm$0.90 & 38.44$\pm$1.08 & 42.55$\pm$0.48 & 42.68$\pm$1.06 \\
\cellcolor{blue!10}Student & 0.73$\pm$0.00&-0.03$\pm$0.01  & \underline{0.56$\pm$0.01} & 0.53$\pm$0.01 & -0.99$\pm$0.19 & 0.50$\pm$0.01 & -0.60$\pm$0.13 & \textbf{0.57$\pm$0.03} & \underline{0.56$\pm$0.01} \\
\cellcolor{red!10}Adult &78.30$\pm$0.63 & 43.79$\pm$1.08 & 72.21$\pm$1.01 & 70.37$\pm$1.45 & \textbf{74.66$\pm$0.67} & 55.79$\pm$4.61 & \underline{73.97$\pm$3.10} & 63.70$\pm$0.91 & 64.20$\pm$0.80 \\
\cellcolor{red!10}Heart & 98.71$\pm$0.43& 27.68$\pm$0.00 & 77.05$\pm$1.77 & \textbf{85.16$\pm$0.96} & 77.13$\pm$1.30 & 75.56$\pm$2.72 & \underline{80.86$\pm$2.67} & 79.61$\pm$0.24 & 75.66$\pm$0.74 \\
\midrule
\textbf{Avg Rank} & - & 7.0 & 3.8 & 3.2 & 4.0 & 6.2 & 5.0 & 3.8 & \textbf{3.1} \\

\midrule
\textbf{Total Rank} & - & 7.5 & 4.8 & 5.1 & 4.0 &4.3  & 4.7 &\textbf{2.2 }& 3.5 \\
\bottomrule
\end{tabularx}
}
\end{adjustbox}
\end{table*}

\clearpage

\section{Privacy Evaluation via Distance to Closest Record (DCR)}
\label{app:dcr}
To assess whether synthetic samples inadvertently reproduce training records, we follow prior work on tabular data synthesis and report the \emph{Distance to Closest Record (DCR)}.  
For each generated instance $s_{\text{gen}}$, DCR computes its distance to the nearest neighbour in the original training set $T_{\text{train}}$:
\begin{equation}
    \mathrm{DCR}(s_{\text{gen}}) 
    = \min_{x \in T_{\text{train}}} d(s_{\text{gen}}, x),
\end{equation}
where the mixed-type distance $d(\cdot,\cdot)$ is defined as an $L_1$ distance on numerical features and a $0/1$ contribution for categorical features (0 if equal, 1 otherwise).  
Small DCR values, particularly those close to~0, would indicate potential memorization or replication of training instances.

\subsection{DCR Results and Analysis}

\begin{figure*}[b]
    \centering
    \begin{subfigure}{0.23\textwidth}
        \centering
        \includegraphics[width=\linewidth]{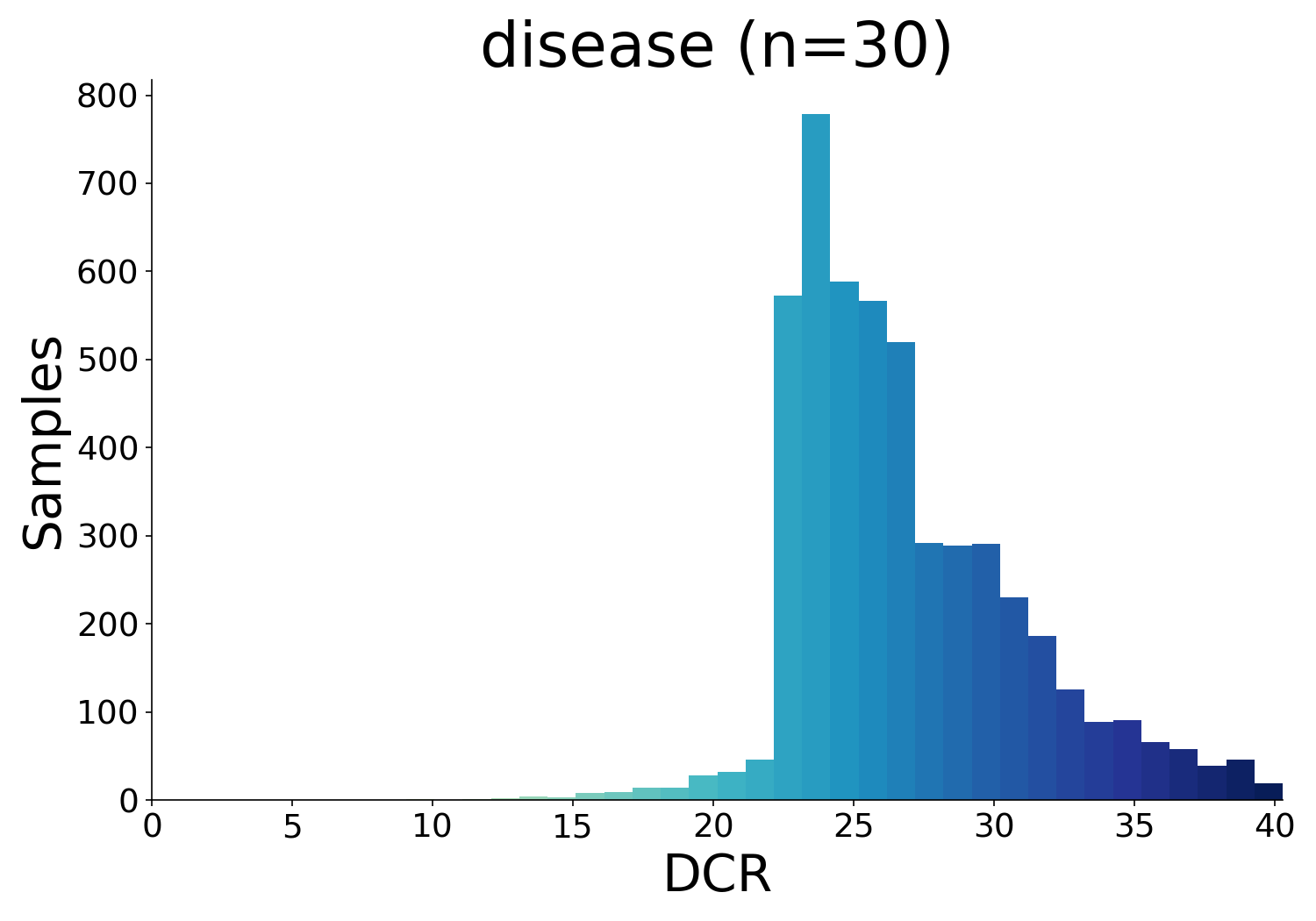}
    \end{subfigure}
    \begin{subfigure}{0.23\textwidth}
        \centering
        \includegraphics[width=\linewidth]{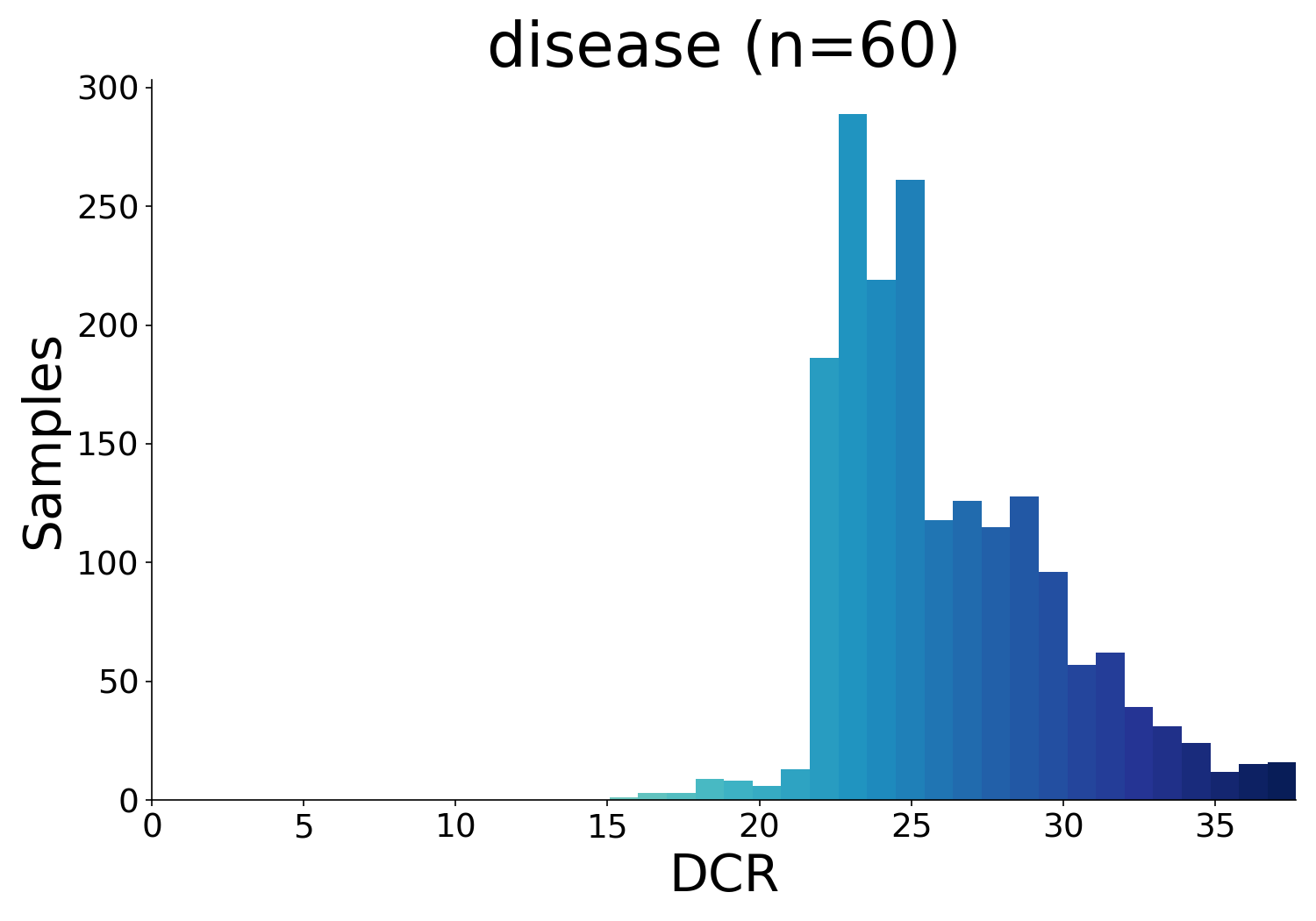}
    \end{subfigure}
    \begin{subfigure}{0.23\textwidth}
        \centering
        \includegraphics[width=\linewidth]{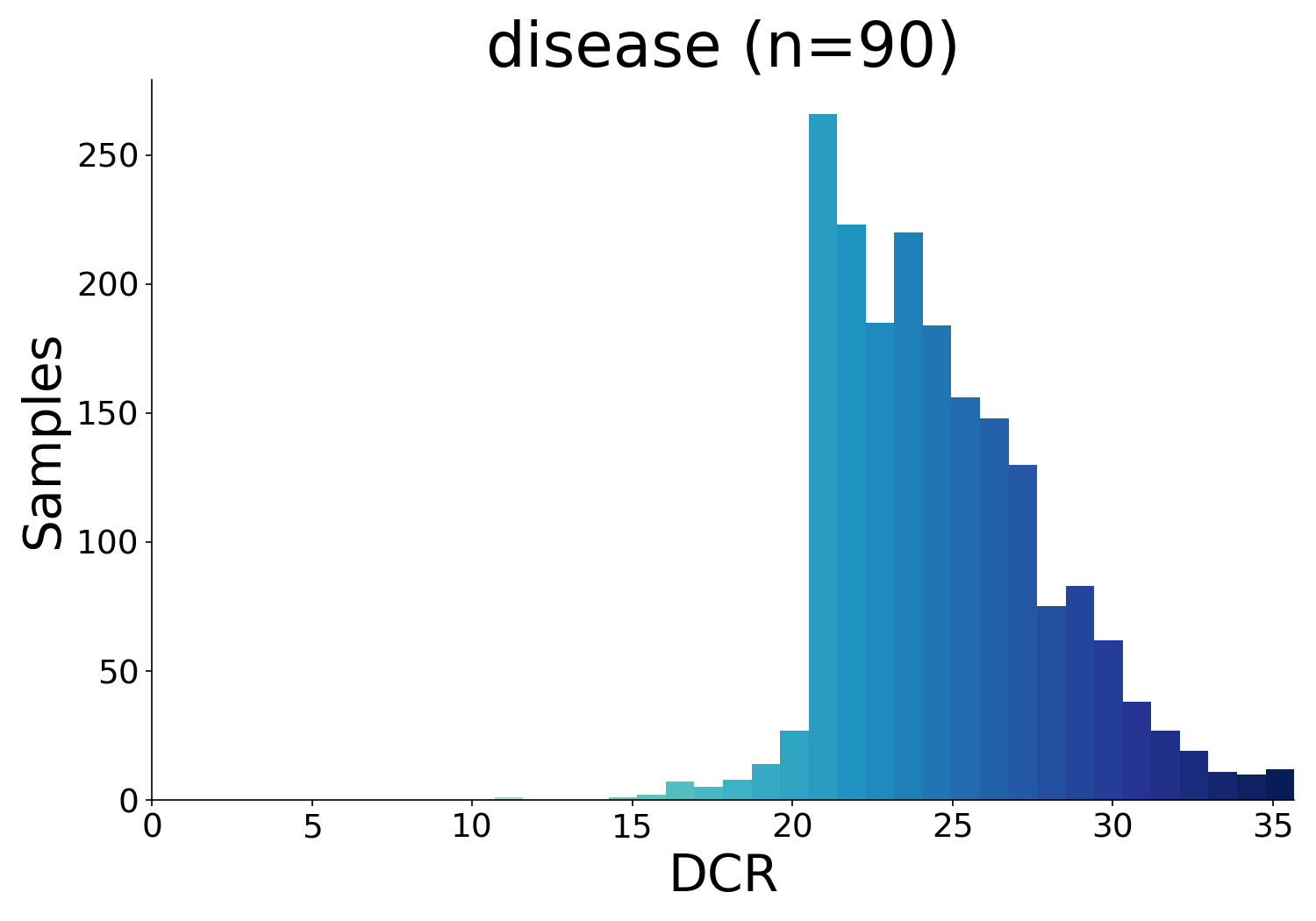}
    \end{subfigure}
    \begin{subfigure}{0.23\textwidth}
        \centering
        \includegraphics[width=\linewidth]{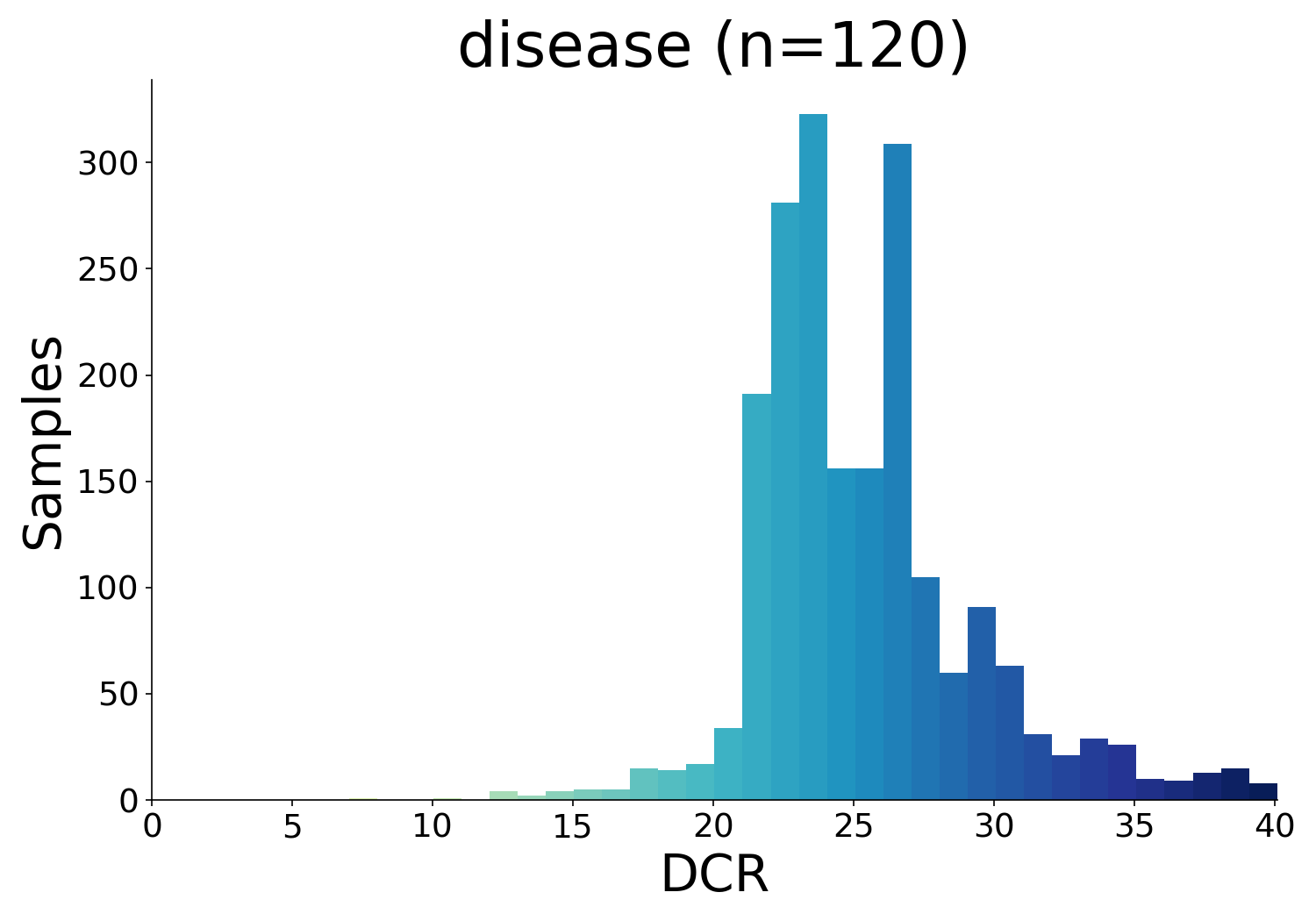}
    \end{subfigure}
    \begin{subfigure}{0.23\textwidth}
        \centering
        \includegraphics[width=\linewidth]{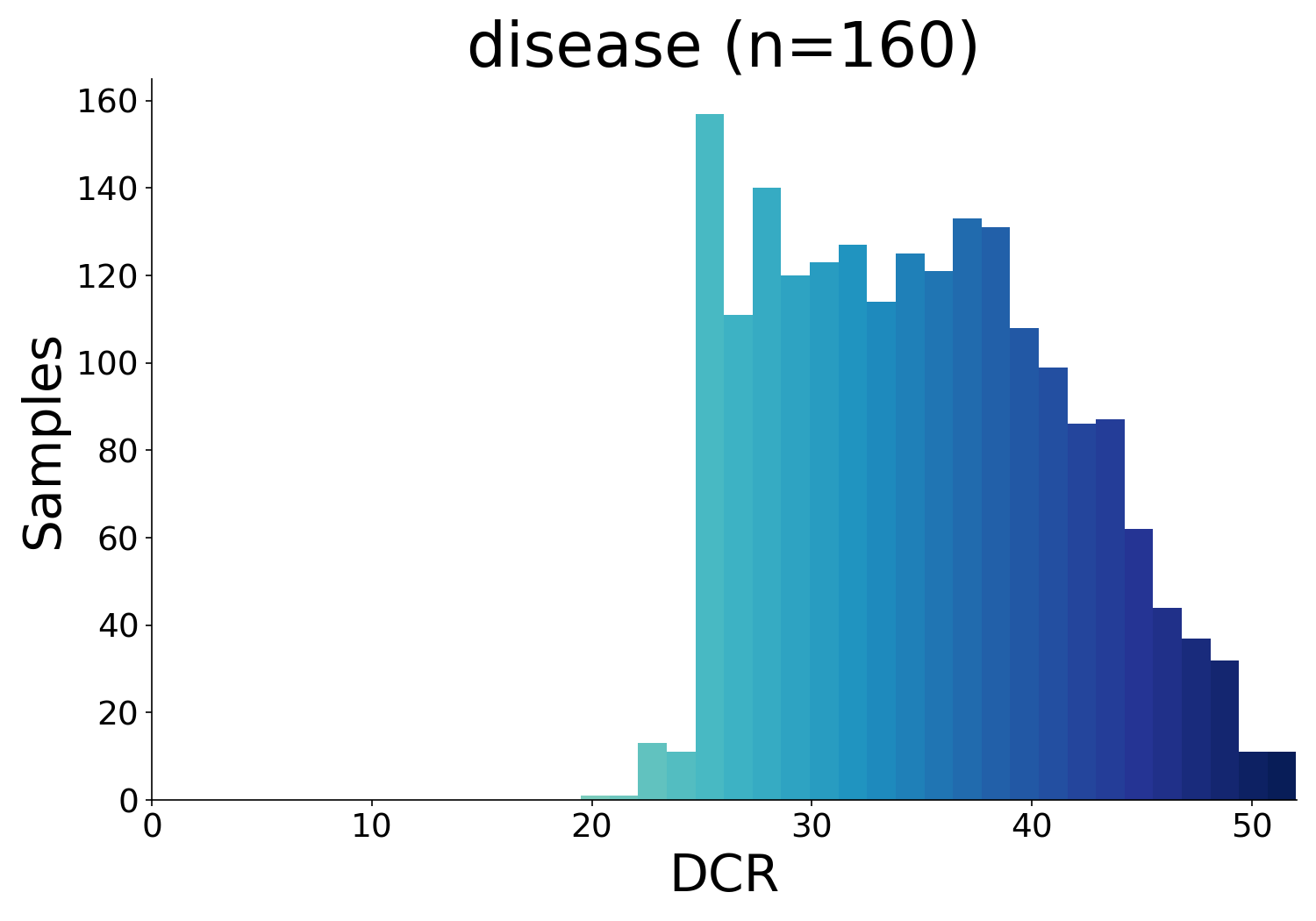}
    \end{subfigure}
    \begin{subfigure}{0.23\textwidth}
        \centering
        \includegraphics[width=\linewidth]{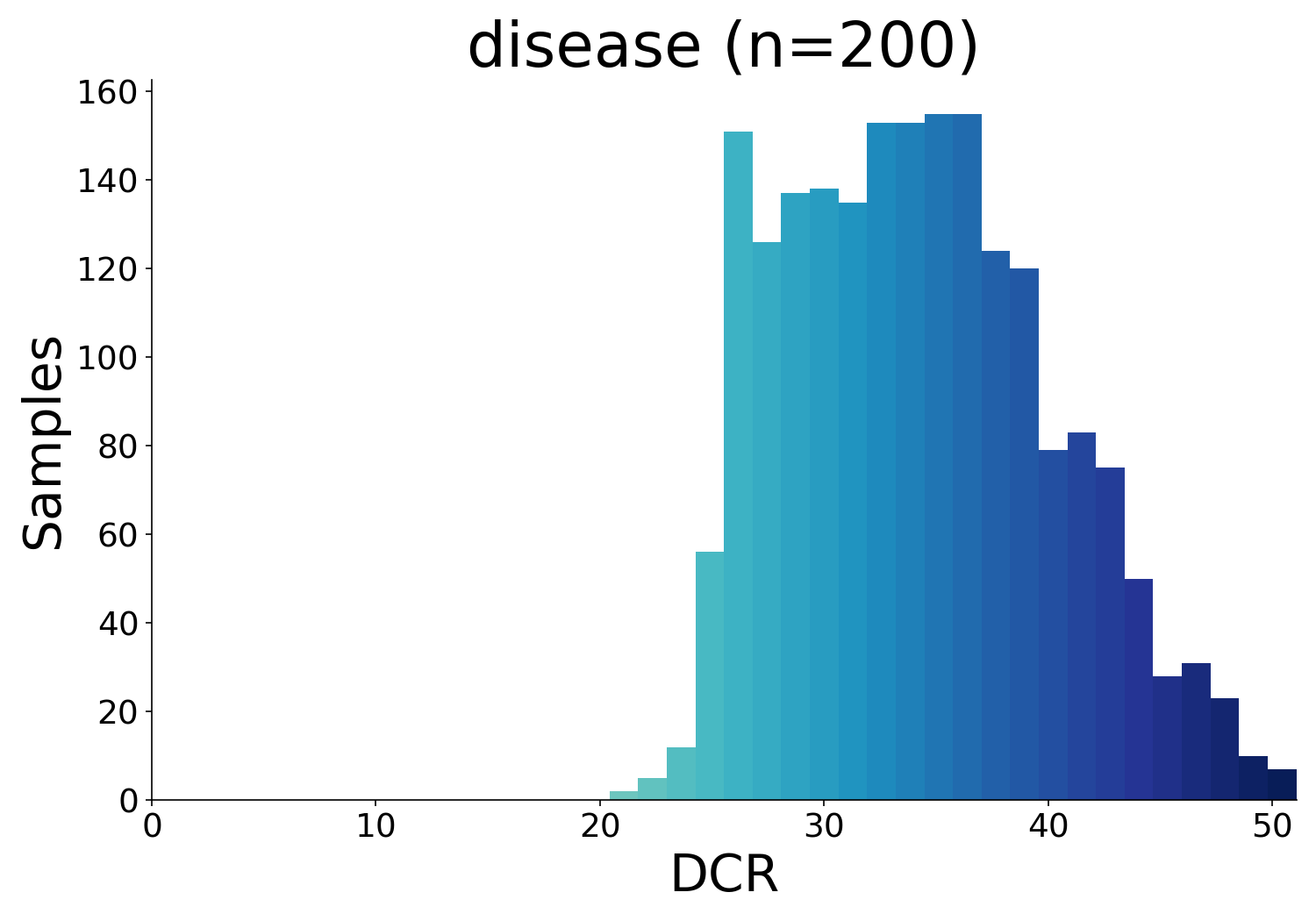}
    \end{subfigure}
    \begin{subfigure}{0.23\textwidth}
        \centering
        \includegraphics[width=\linewidth]{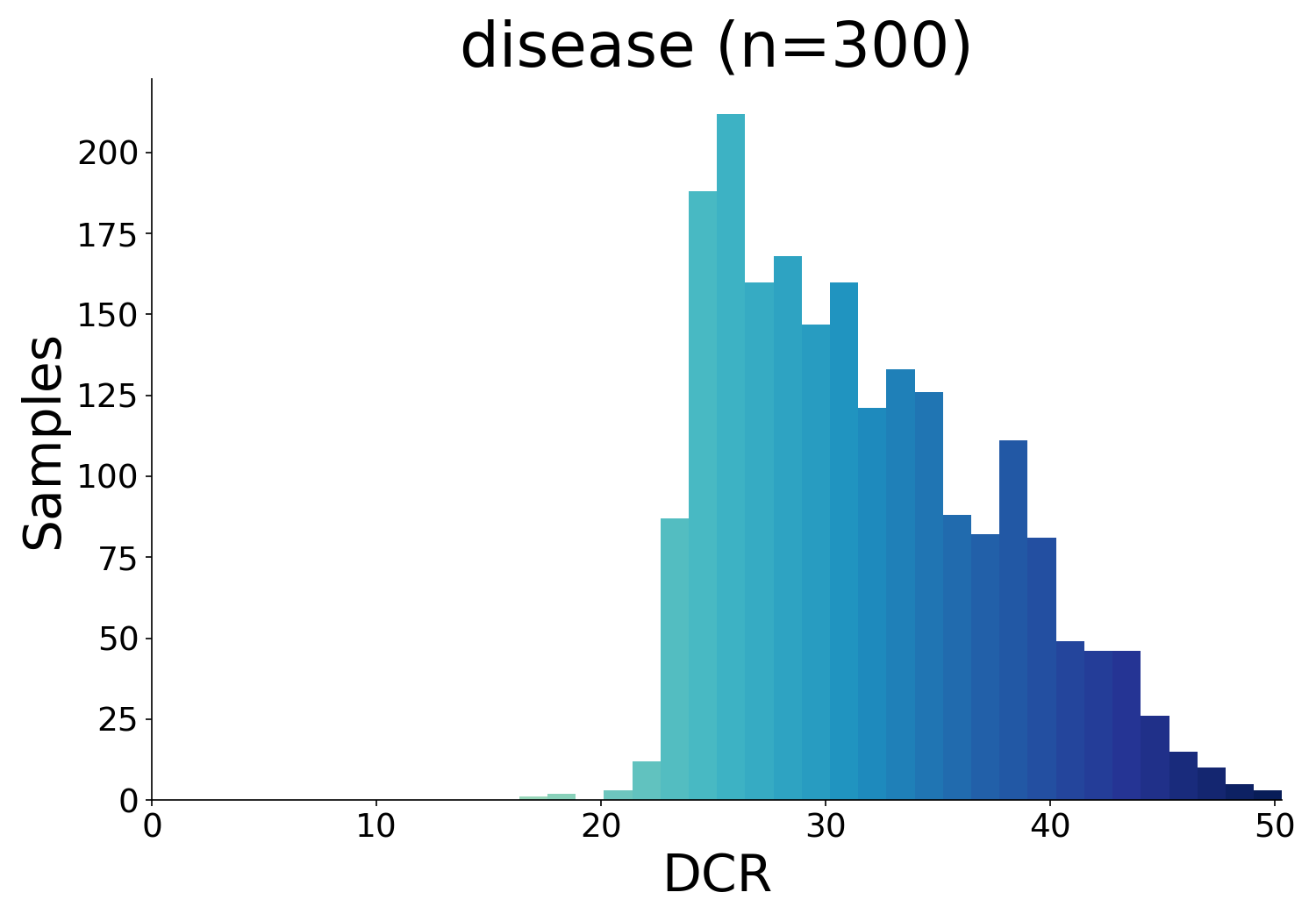}
    \end{subfigure}
        \begin{subfigure}{0.23\textwidth}
        \centering
        \includegraphics[width=\linewidth]{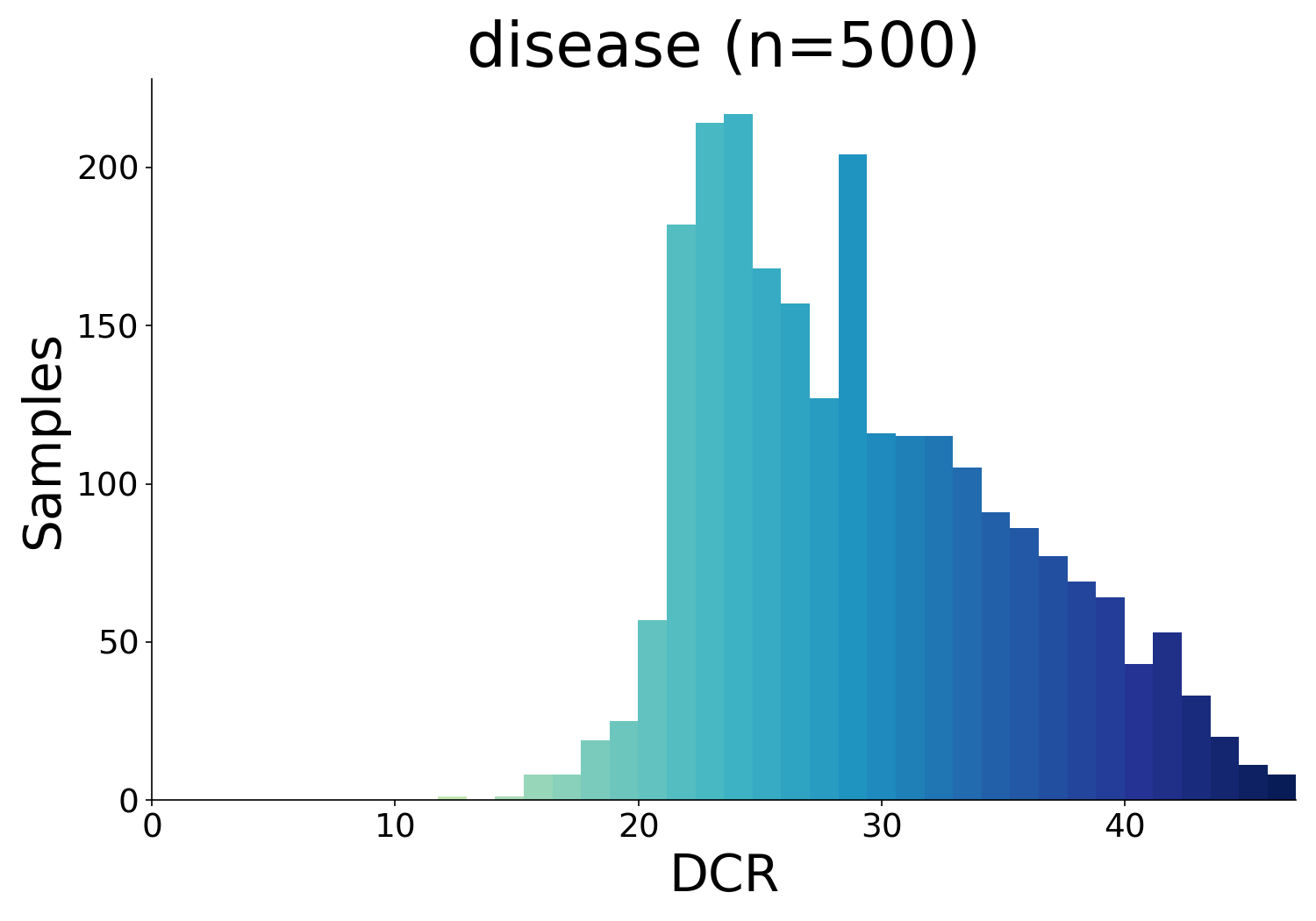}
    \end{subfigure}

    \caption{Distance to closest record (DCR) distributions for the \textsc{Disease} dataset, 
    comparing synthetic data generated by \textbf{ReFine} to the original train set.}
    \label{fig:dcr_disease}
\end{figure*}

\begin{figure*}[htbp]
    \centering
    \begin{subfigure}{0.23\textwidth}
        \centering
        \includegraphics[width=\linewidth]{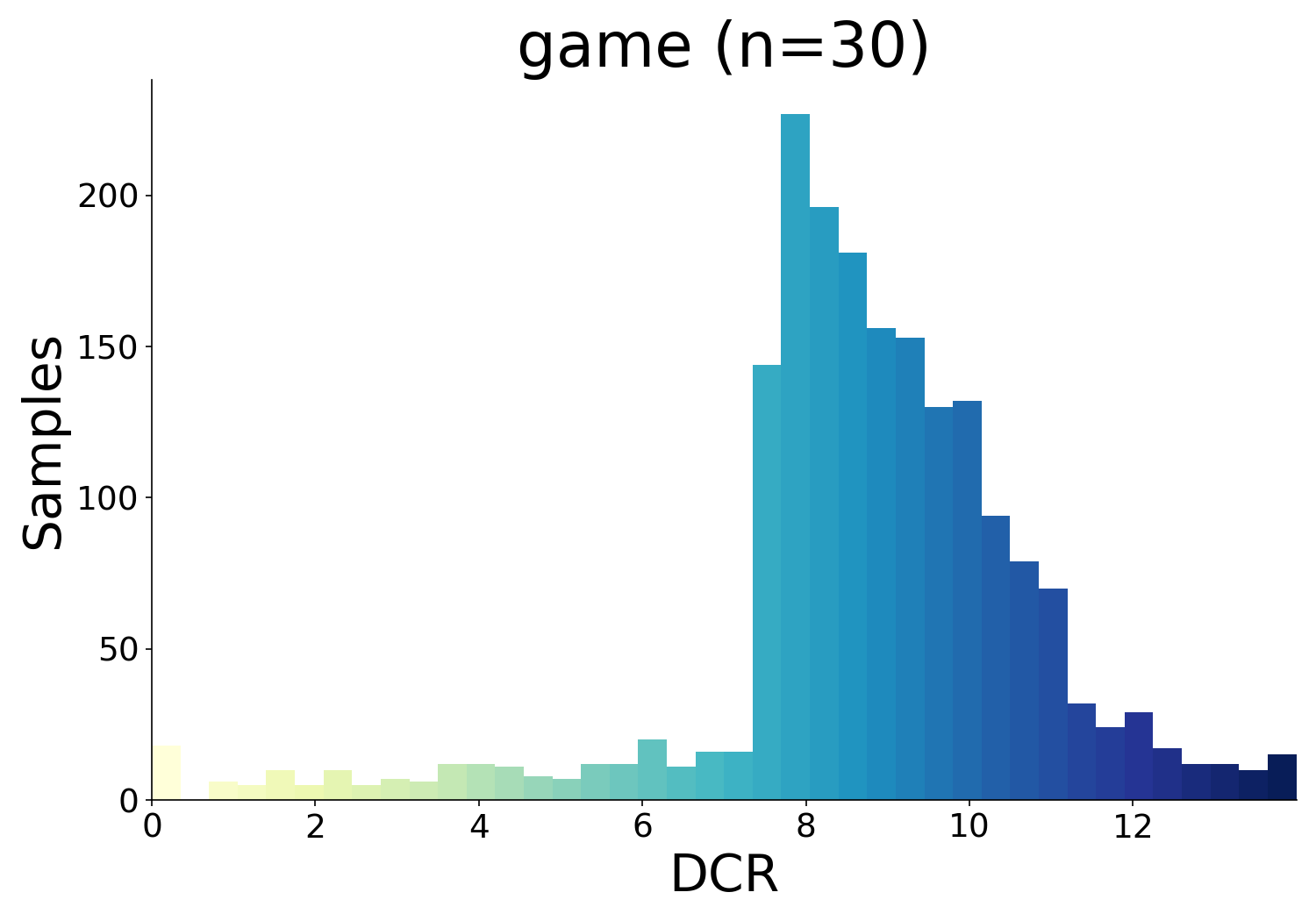}
    \end{subfigure}
    \begin{subfigure}{0.23\textwidth}
        \centering
        \includegraphics[width=\linewidth]{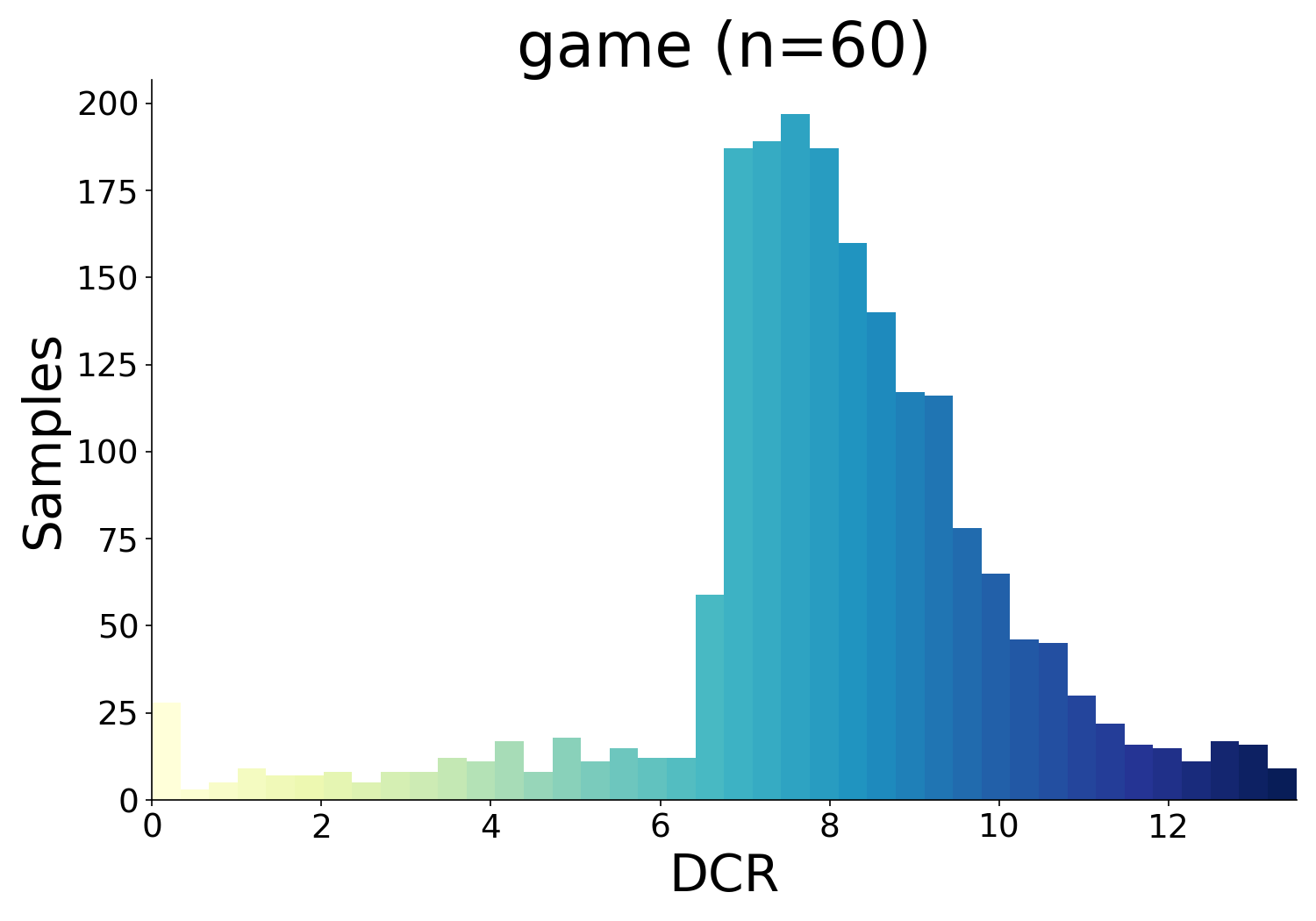}
    \end{subfigure}
    \begin{subfigure}{0.23\textwidth}
        \centering
        \includegraphics[width=\linewidth]{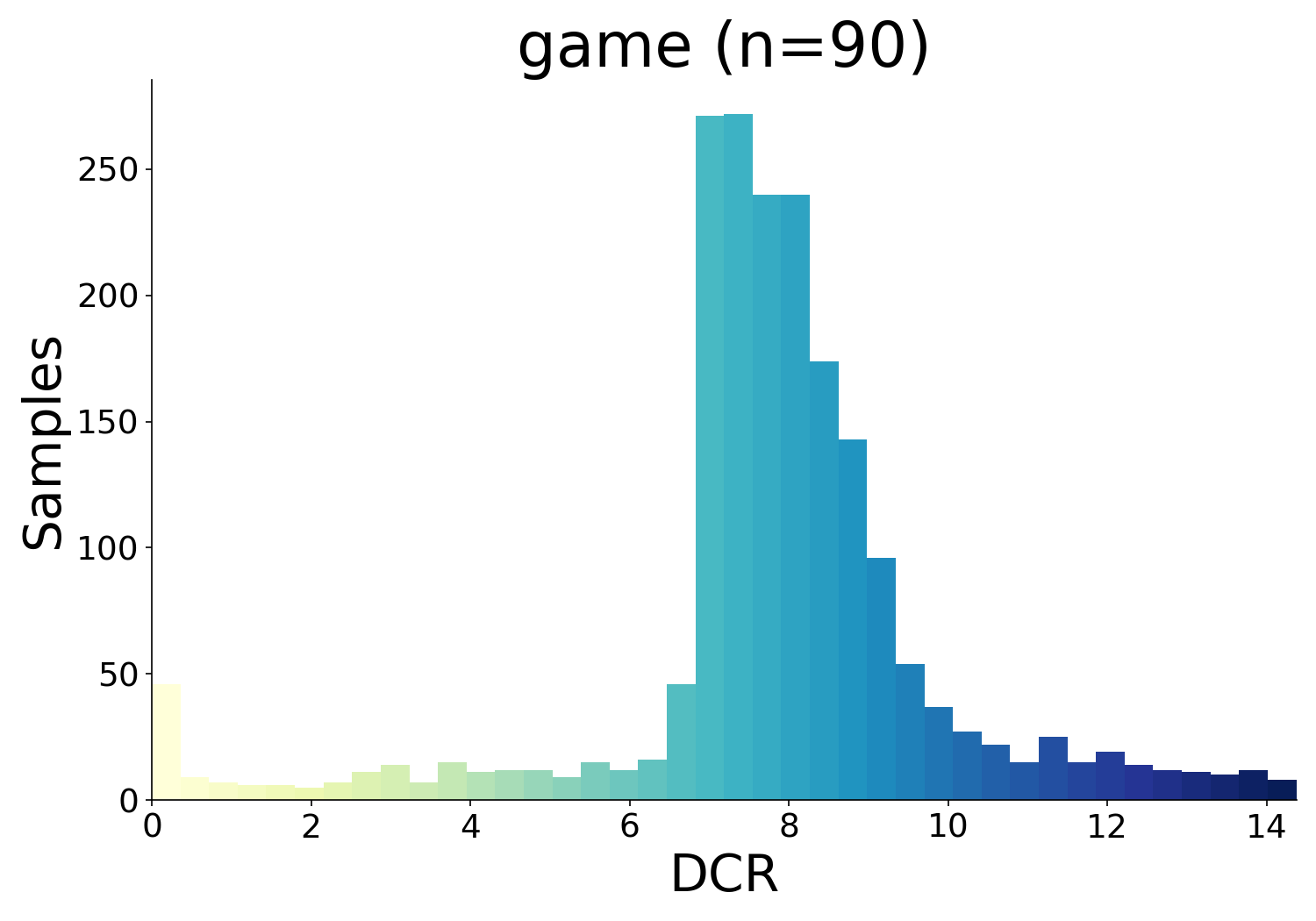}
    \end{subfigure}
    \begin{subfigure}{0.23\textwidth}
        \centering
        \includegraphics[width=\linewidth]{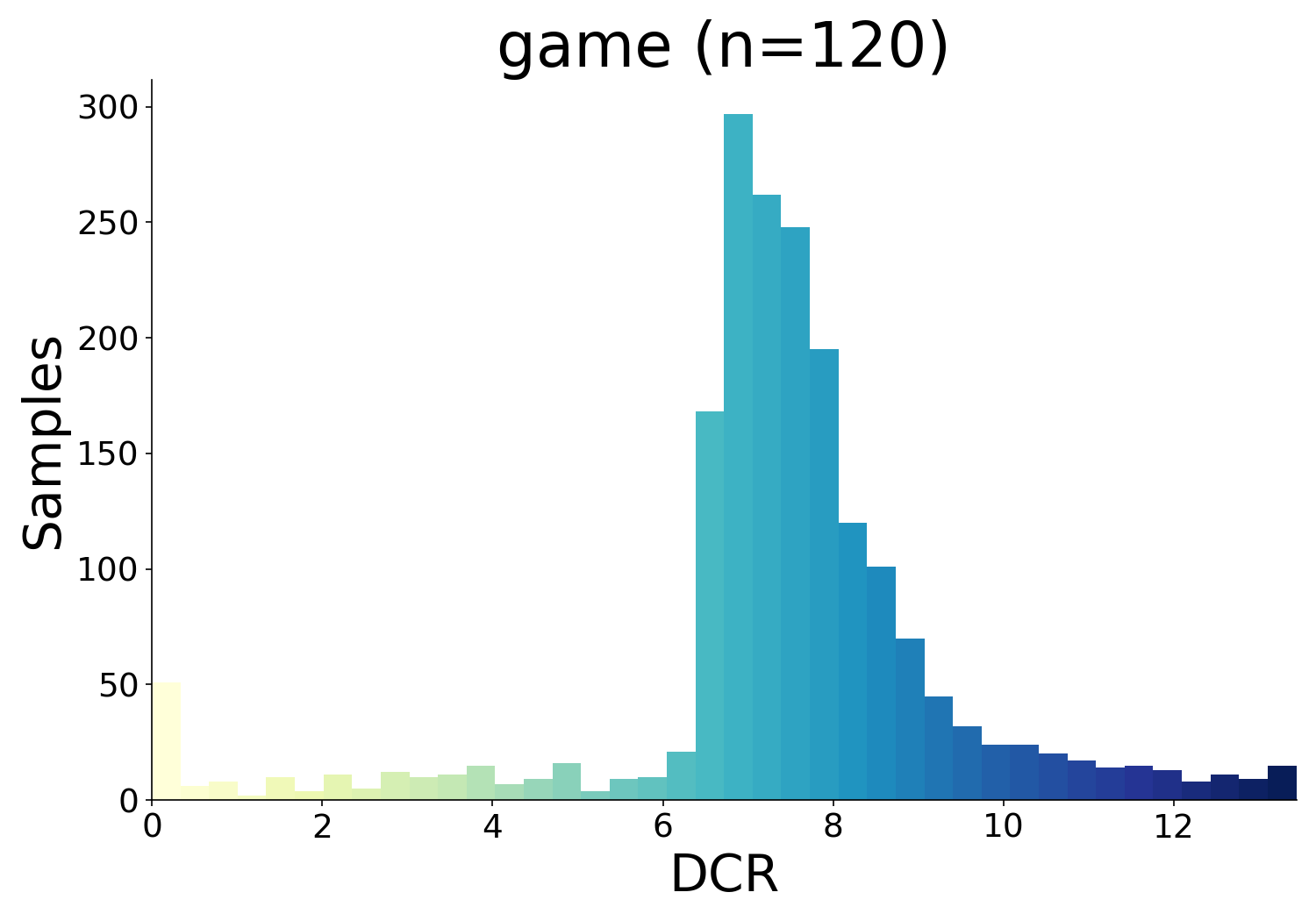}
    \end{subfigure}
    \begin{subfigure}{0.23\textwidth}
        \centering
        \includegraphics[width=\linewidth]{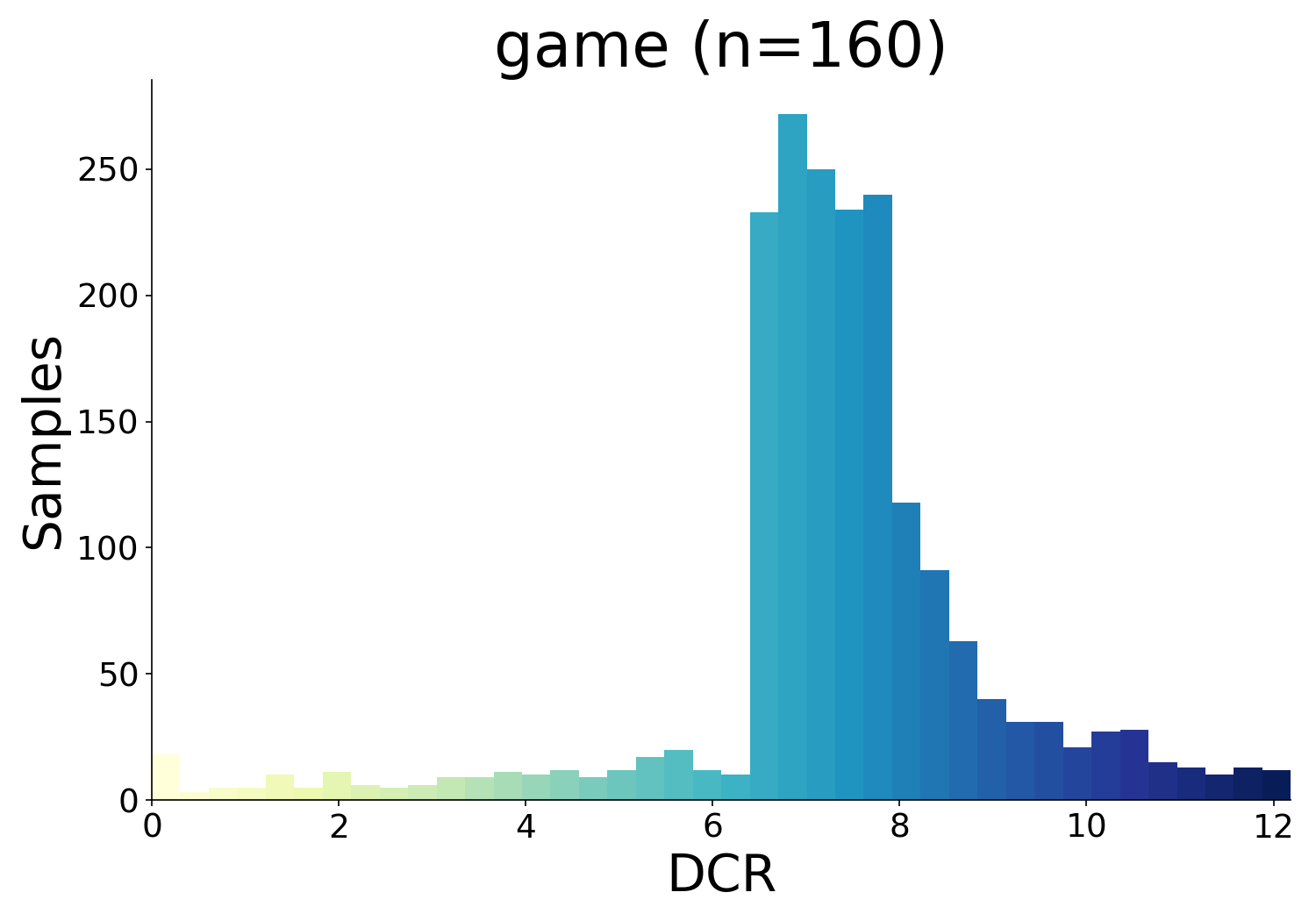}
    \end{subfigure}
    \begin{subfigure}{0.23\textwidth}
        \centering
        \includegraphics[width=\linewidth]{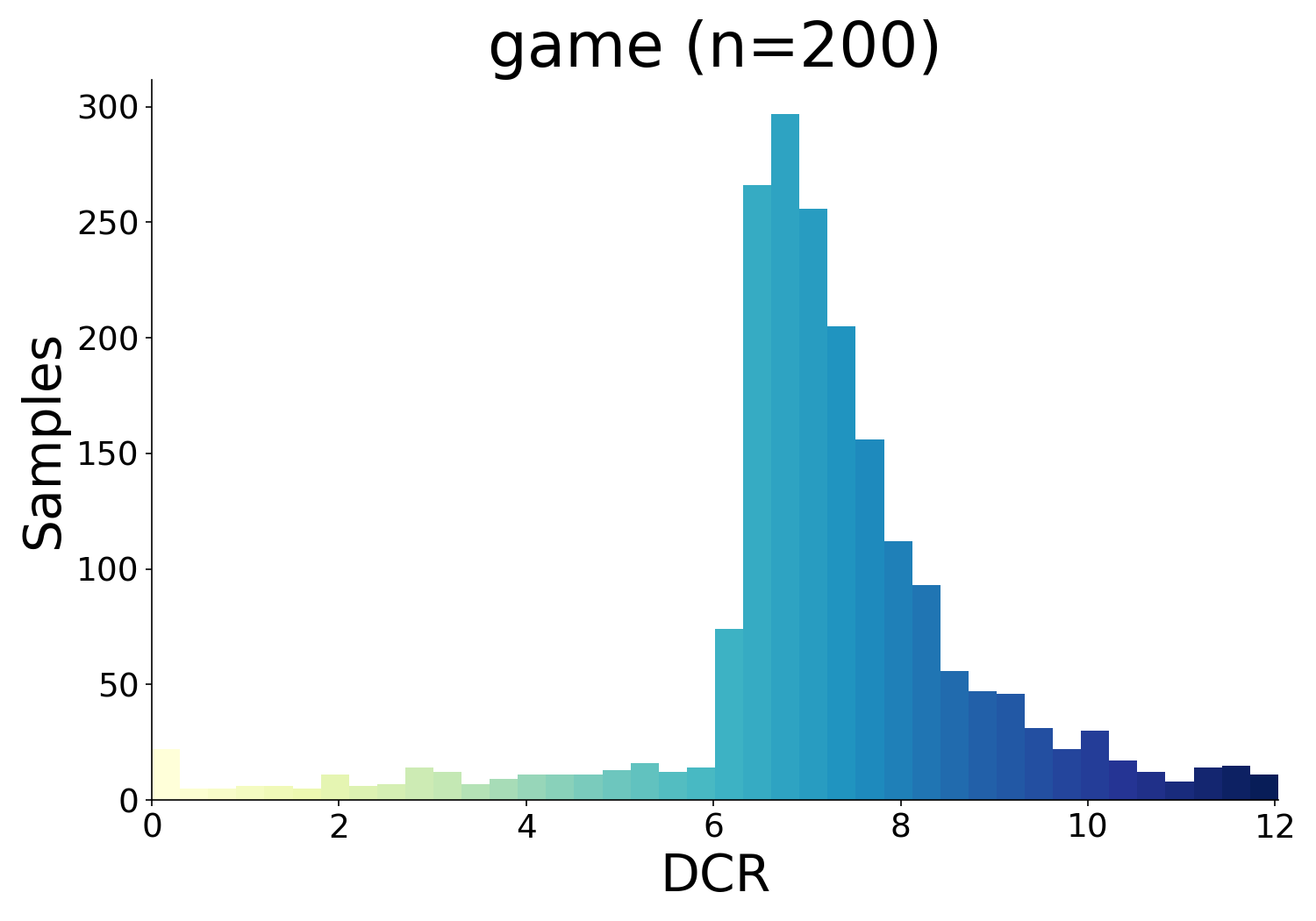}
    \end{subfigure}
    \begin{subfigure}{0.23\textwidth}
        \centering
        \includegraphics[width=\linewidth]{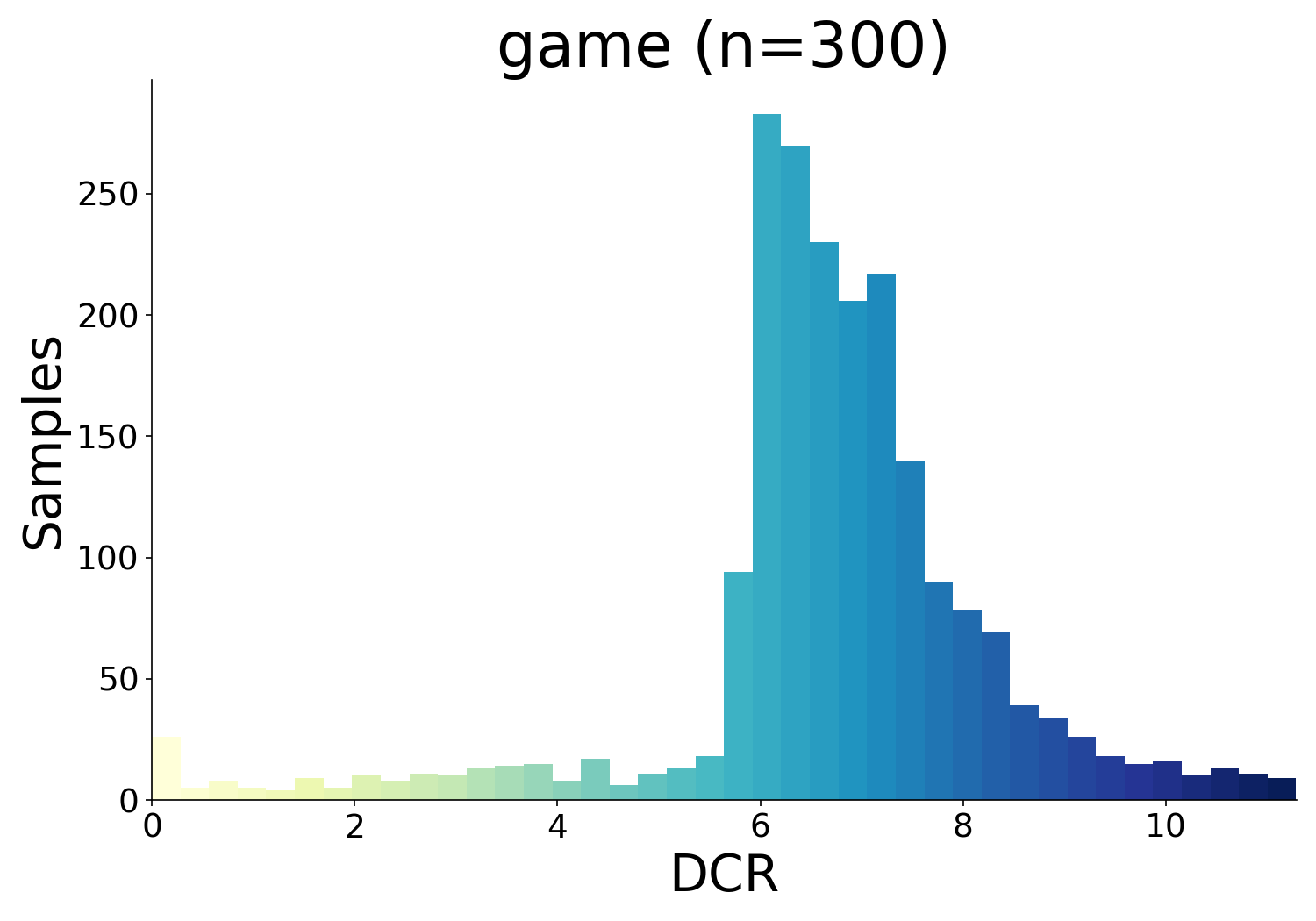}
    \end{subfigure}
    \begin{subfigure}{0.23\textwidth}
        \centering
        \includegraphics[width=\linewidth]{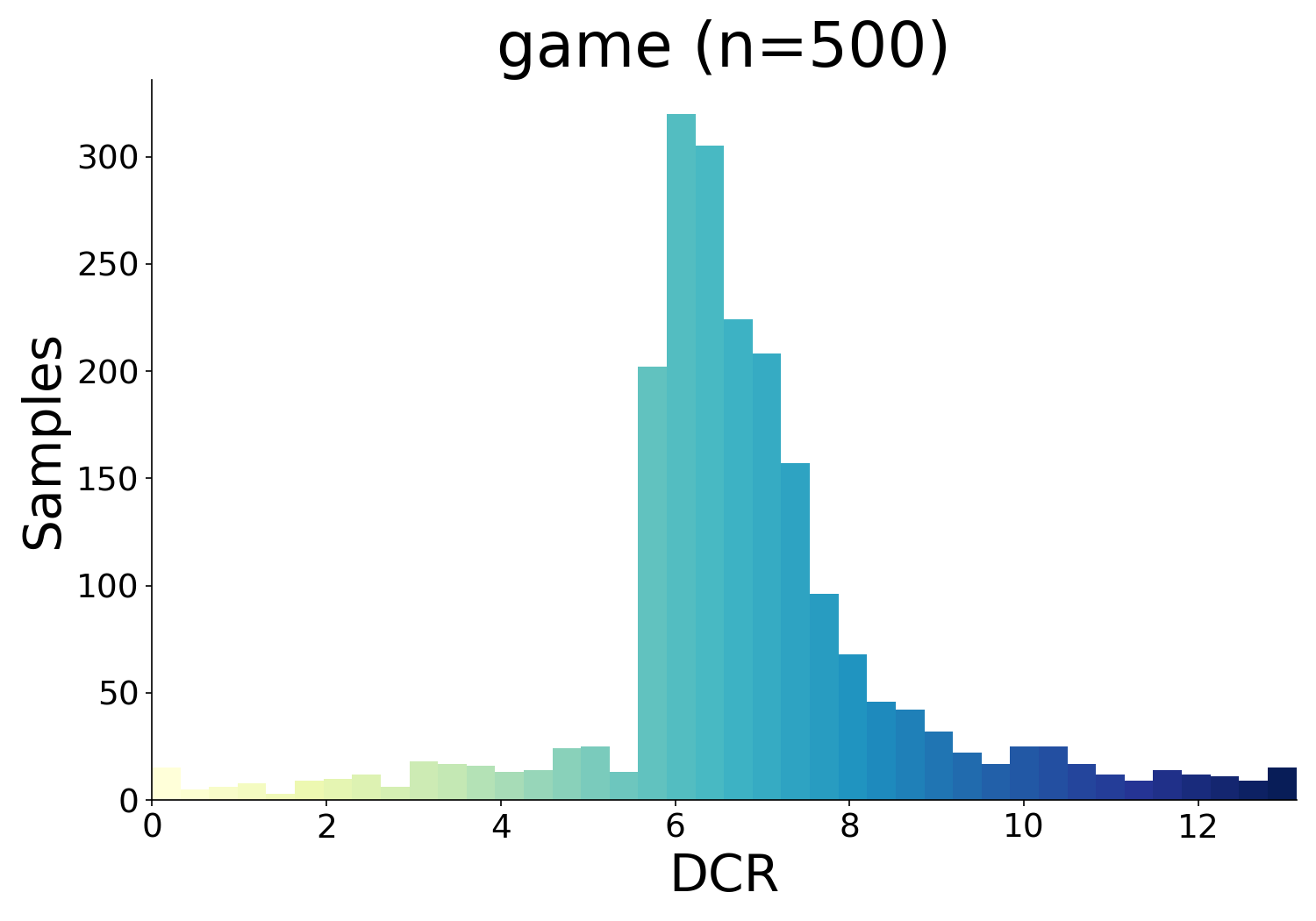}
    \end{subfigure}

    \caption{Distance to closest record (DCR) distributions for the \textsc{Game} dataset, 
    comparing synthetic data generated by \textbf{ReFine} to the original train set.}
    \label{fig:dcr_game}
\end{figure*}

\begin{figure*}[htbp]
    \centering
    \begin{subfigure}{0.23\textwidth}
        \centering
        \includegraphics[width=\linewidth]{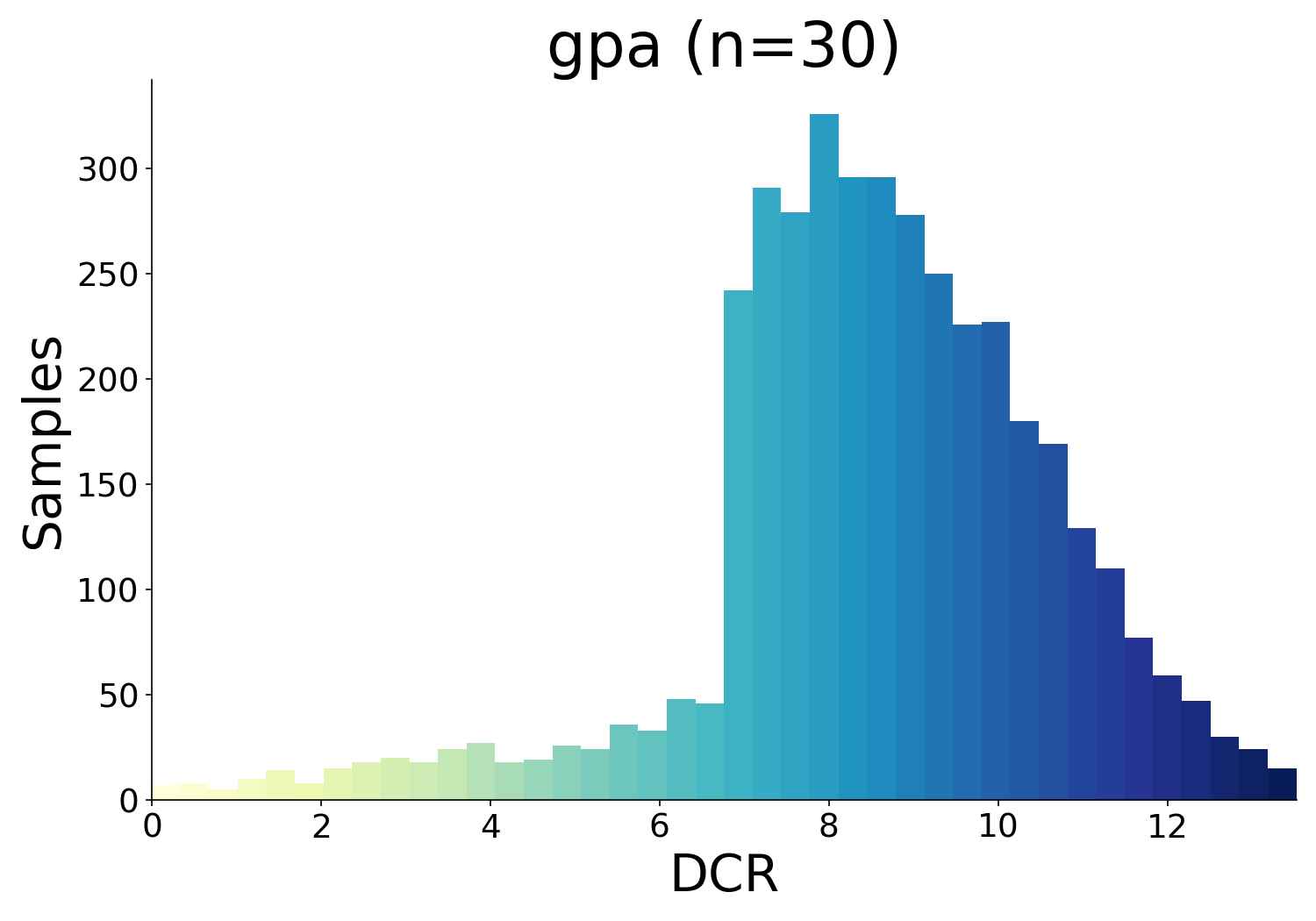}
    \end{subfigure}
    \begin{subfigure}{0.23\textwidth}
        \centering
        \includegraphics[width=\linewidth]{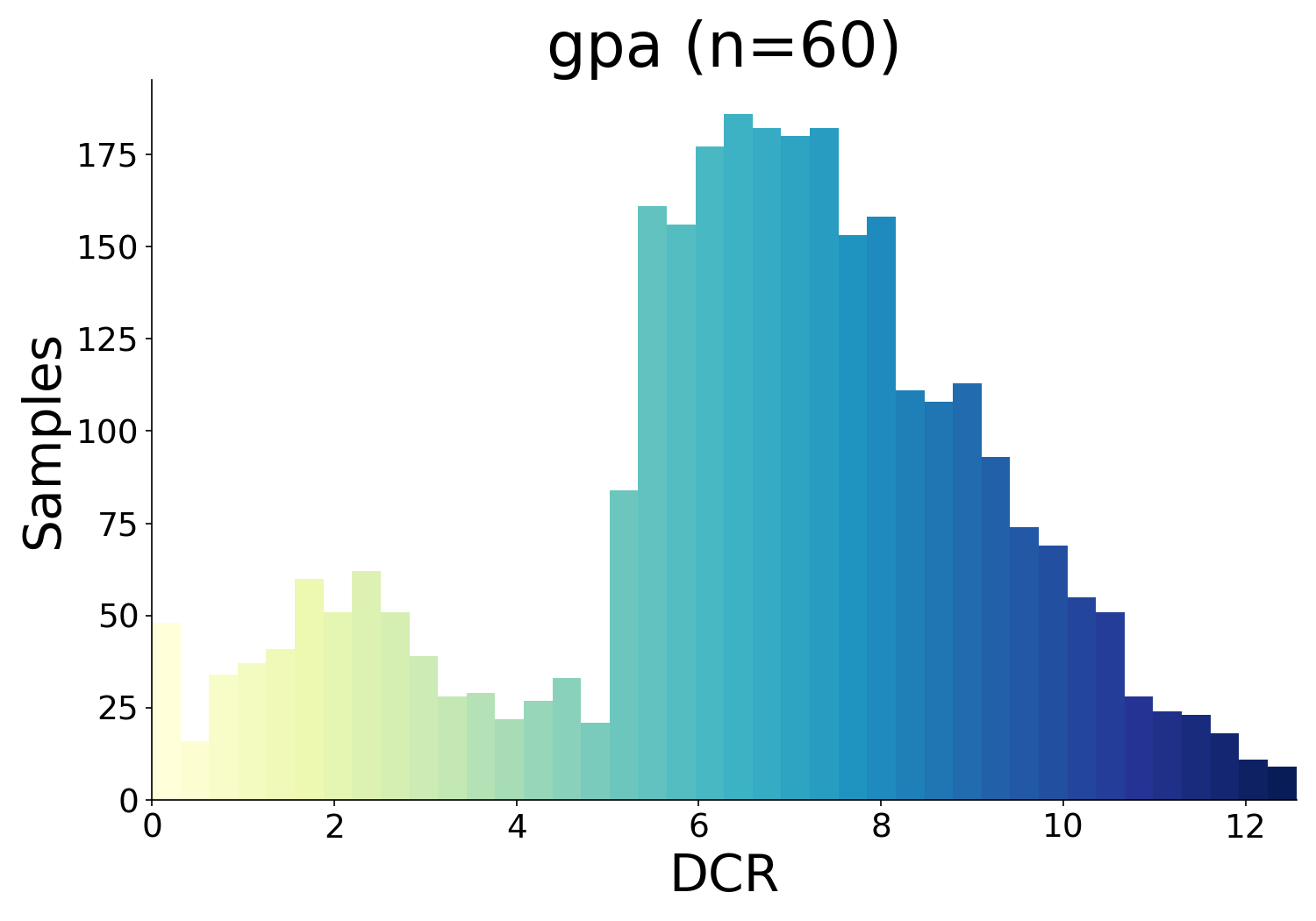}
    \end{subfigure}
    \begin{subfigure}{0.23\textwidth}
        \centering
        \includegraphics[width=\linewidth]{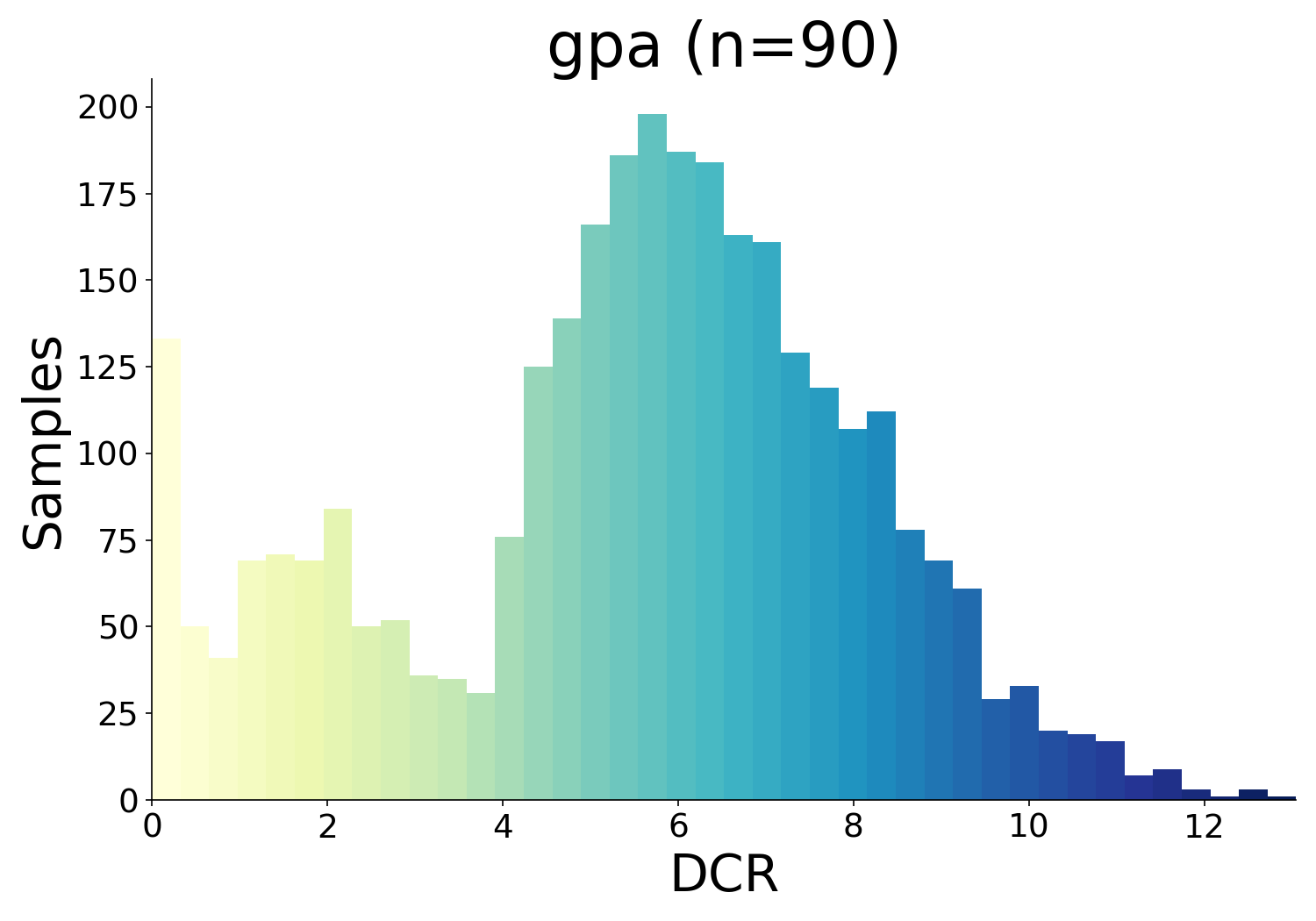}
    \end{subfigure}
    \begin{subfigure}{0.23\textwidth}
        \centering
        \includegraphics[width=\linewidth]{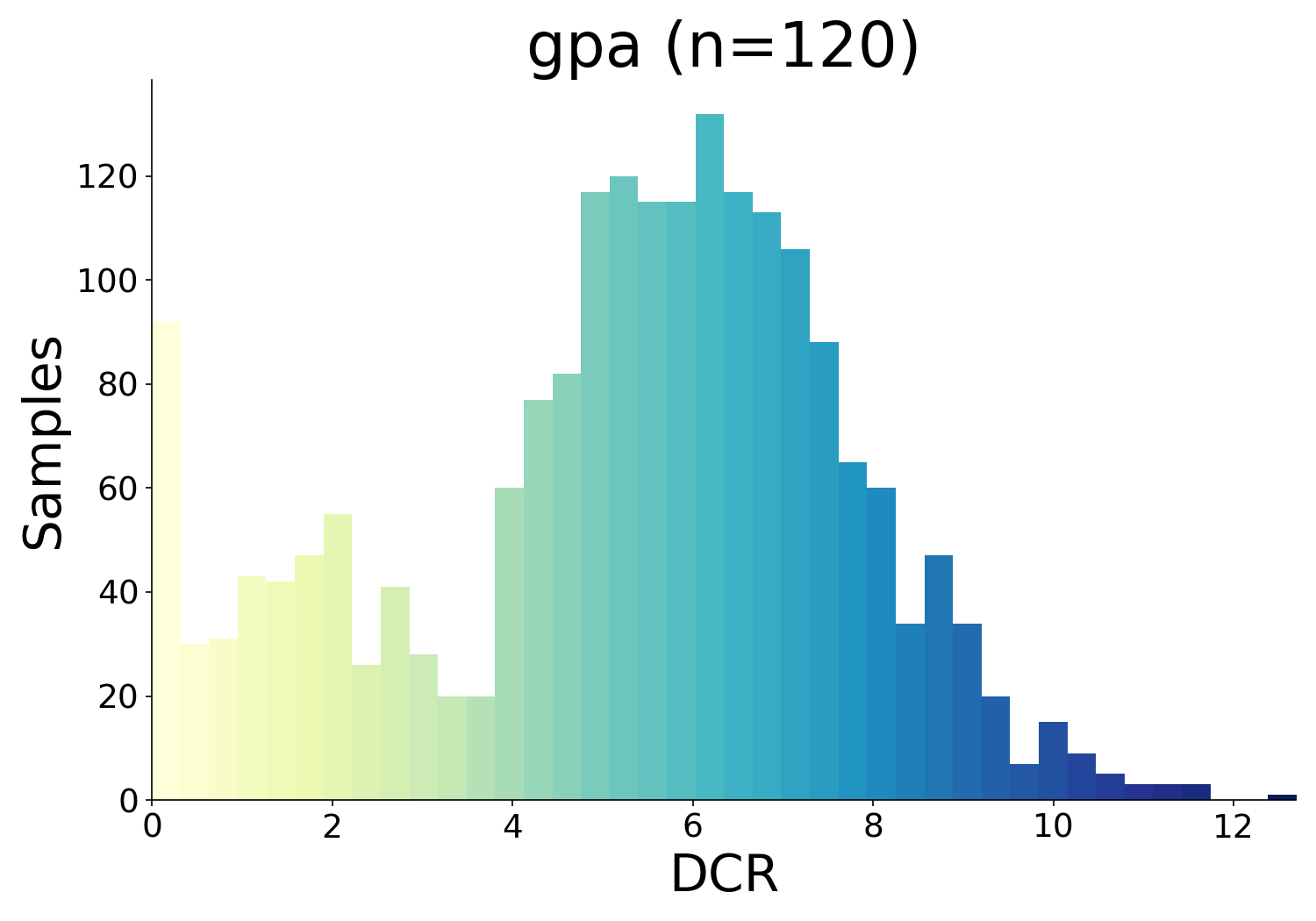}
    \end{subfigure}
    \begin{subfigure}{0.23\textwidth}
        \centering
        \includegraphics[width=\linewidth]{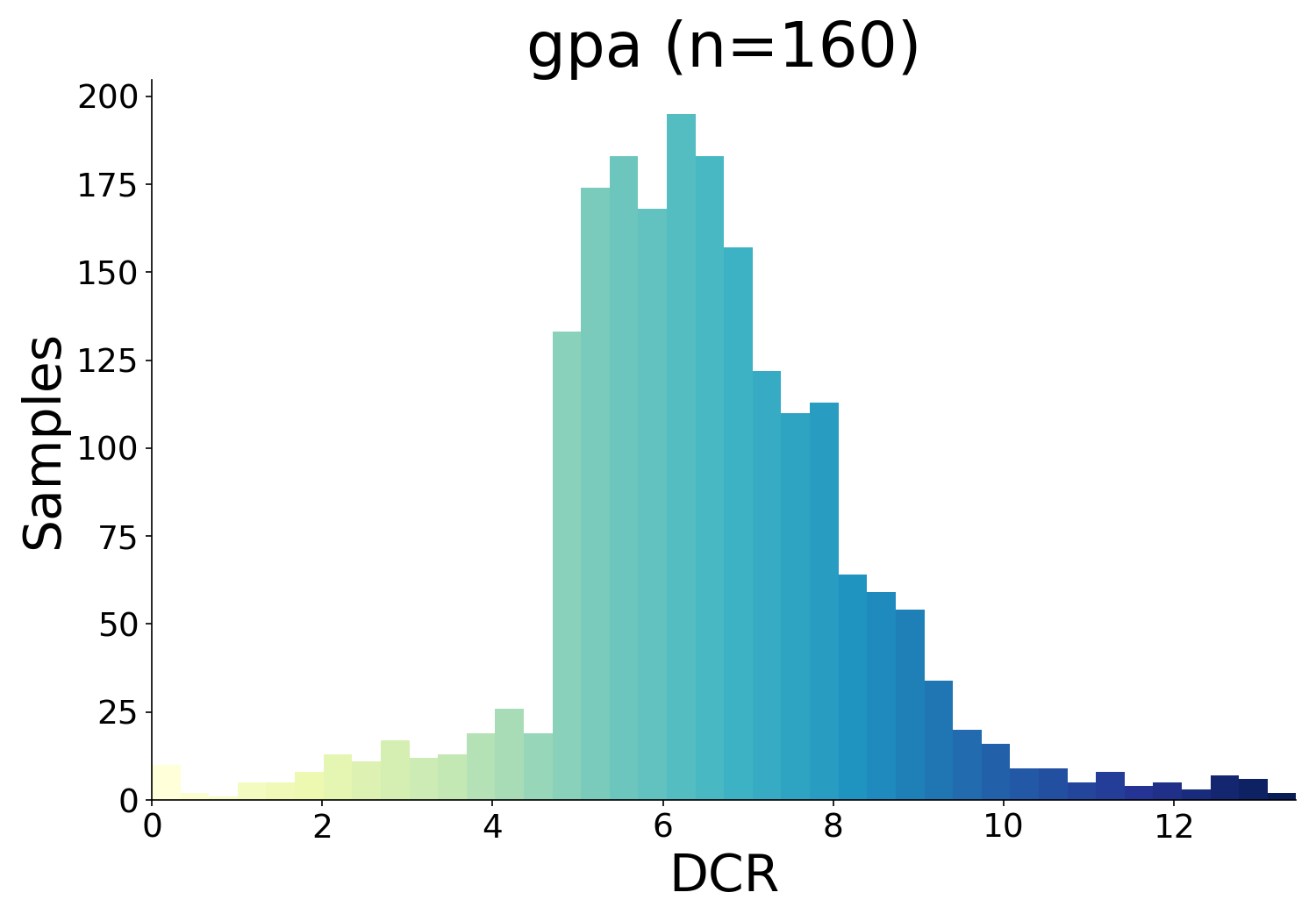}
    \end{subfigure}
    \begin{subfigure}{0.23\textwidth}
        \centering
        \includegraphics[width=\linewidth]{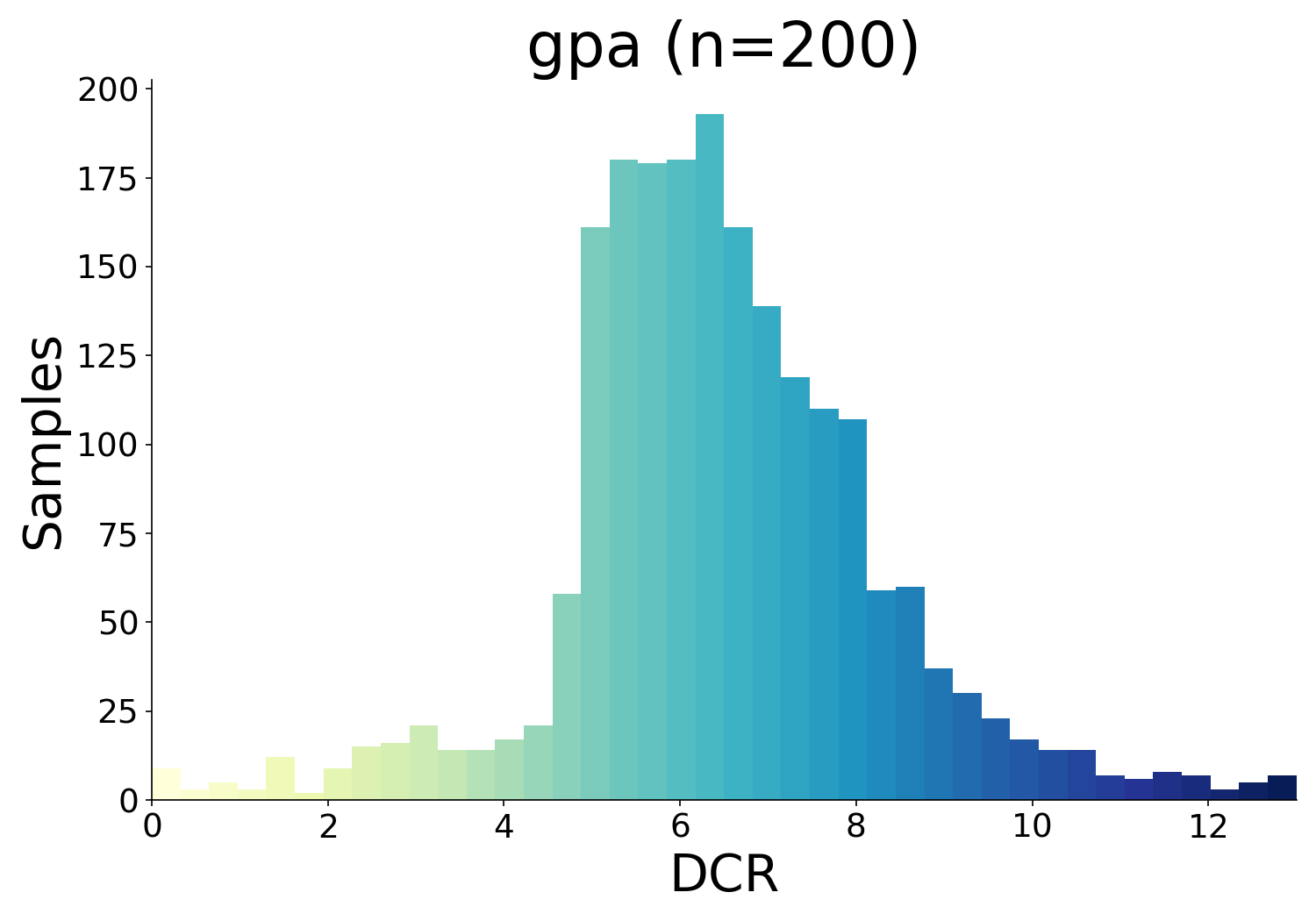}
    \end{subfigure}
    \begin{subfigure}{0.23\textwidth}
        \centering
        \includegraphics[width=\linewidth]{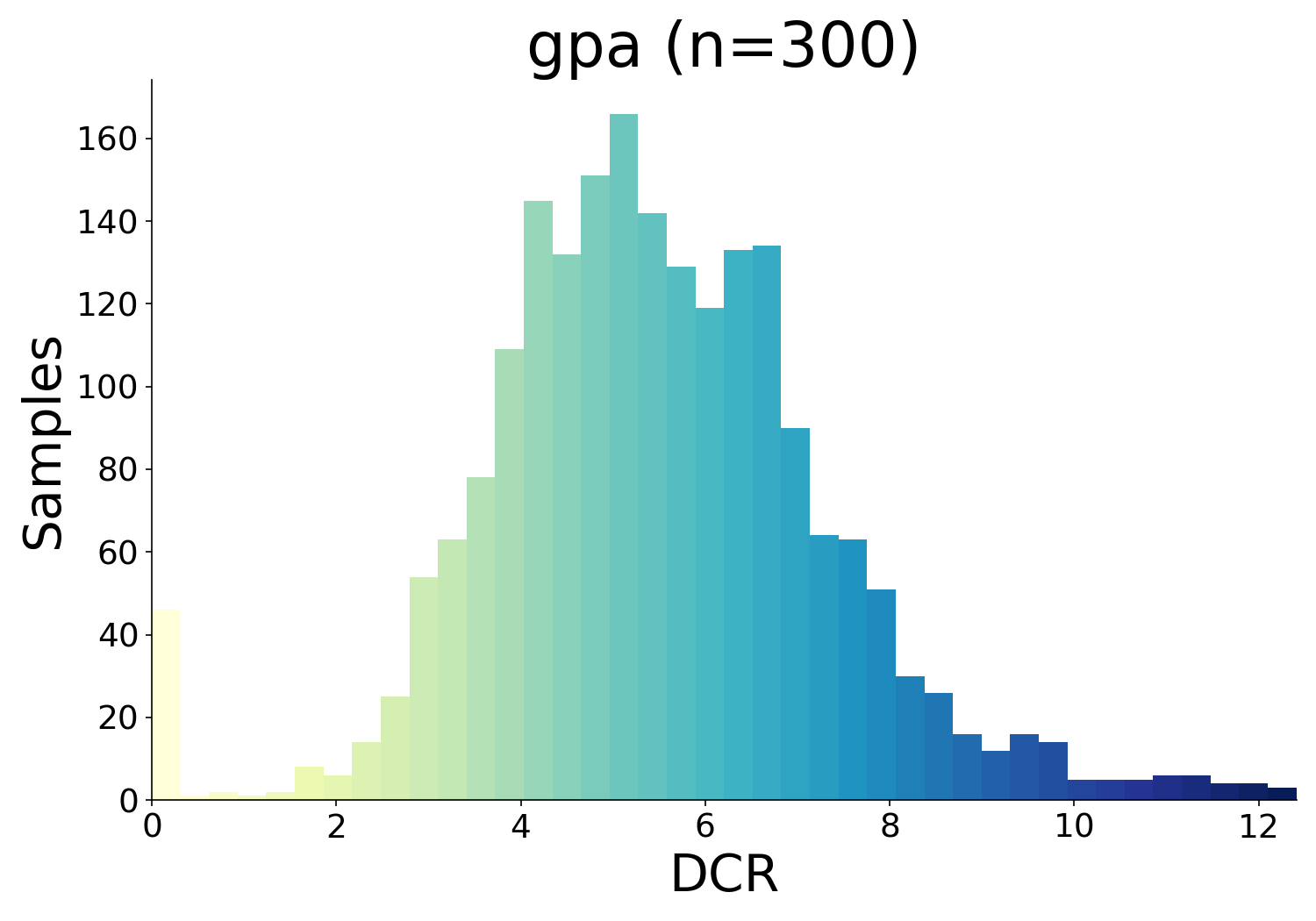}
    \end{subfigure}
    \begin{subfigure}{0.23\textwidth}
        \centering
        \includegraphics[width=\linewidth]{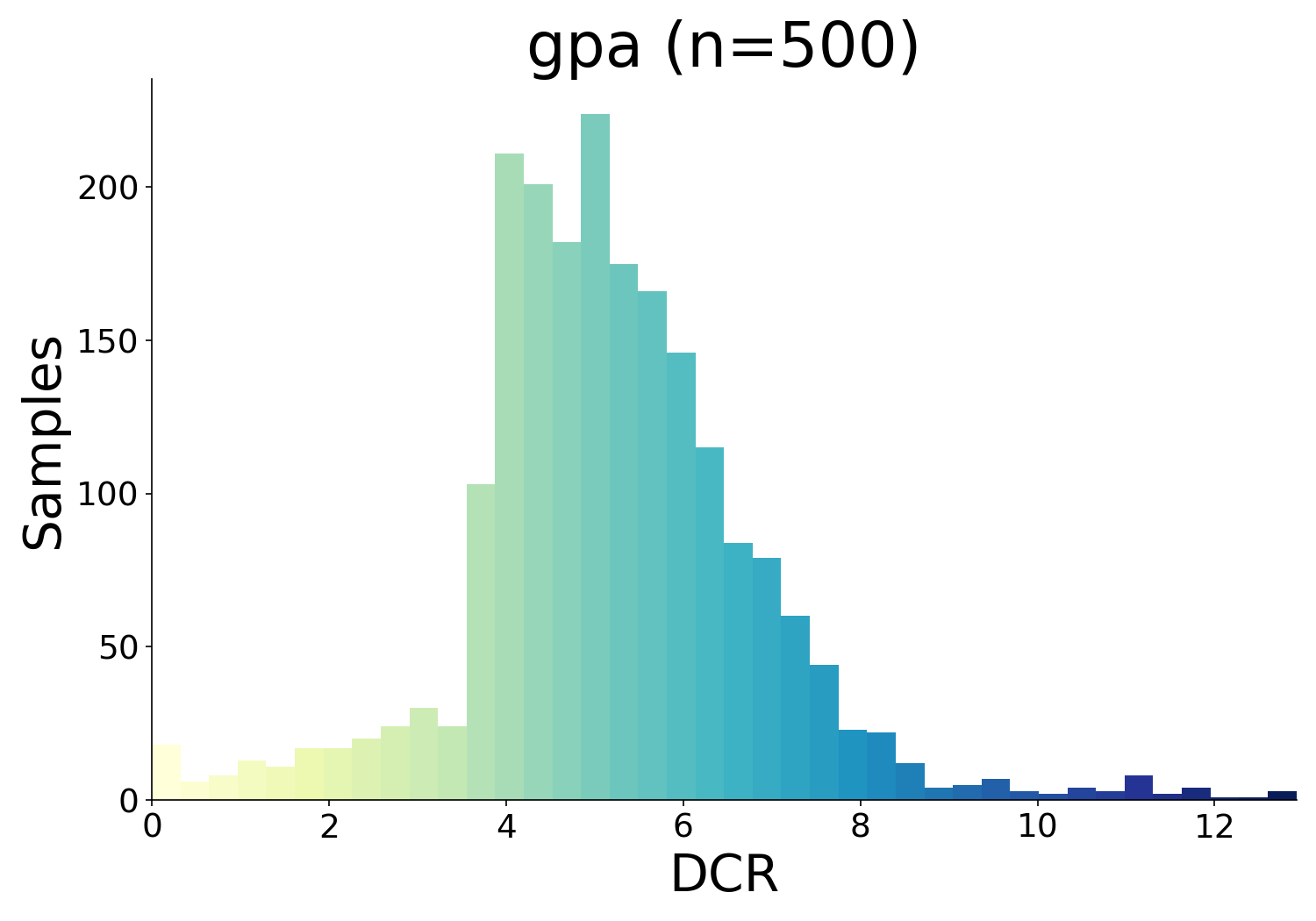}
    \end{subfigure}
    \caption{Distance to closest record (DCR) distributions for the \textsc{Disease} dataset, 
    comparing synthetic data generated by \textbf{ReFine} to the original train set.}
    \label{fig:dcr_gpa}
\end{figure*}

\begin{figure*}[h]
    \centering
    \begin{subfigure}{0.23\textwidth}
        \centering
        \includegraphics[width=\linewidth]{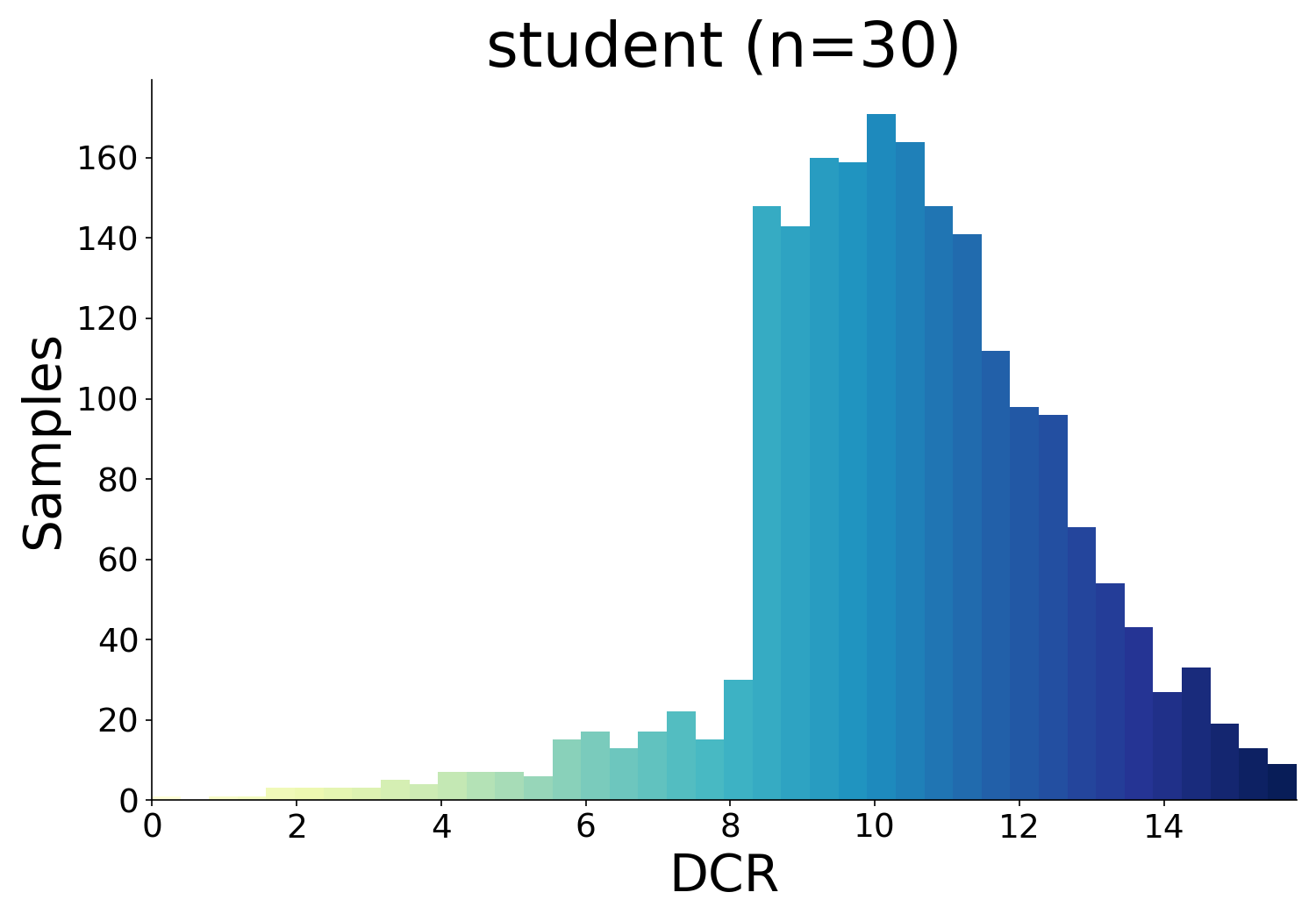}
    \end{subfigure}
    \begin{subfigure}{0.23\textwidth}
        \centering
        \includegraphics[width=\linewidth]{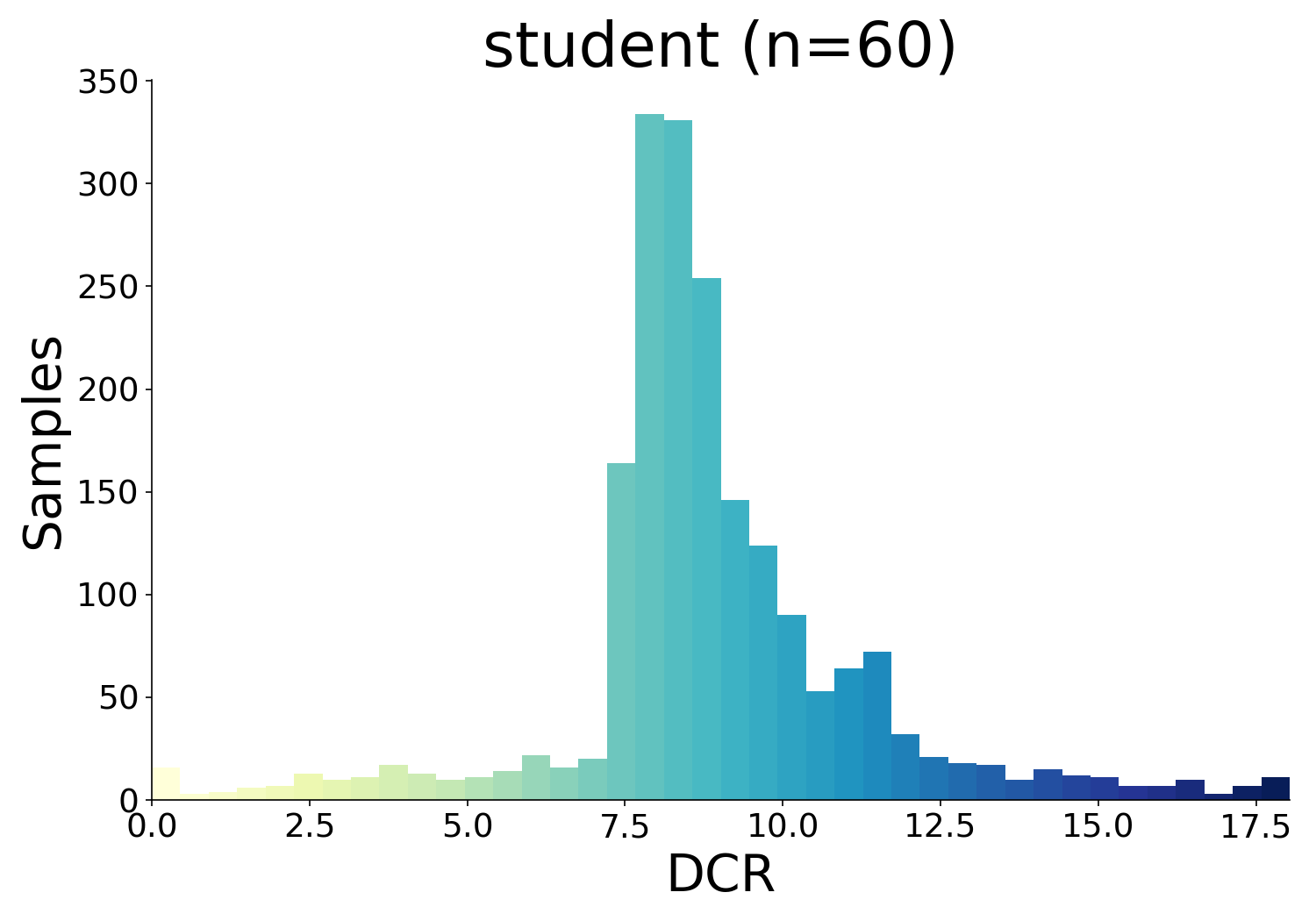}
    \end{subfigure}
    \begin{subfigure}{0.23\textwidth}
        \centering
        \includegraphics[width=\linewidth]{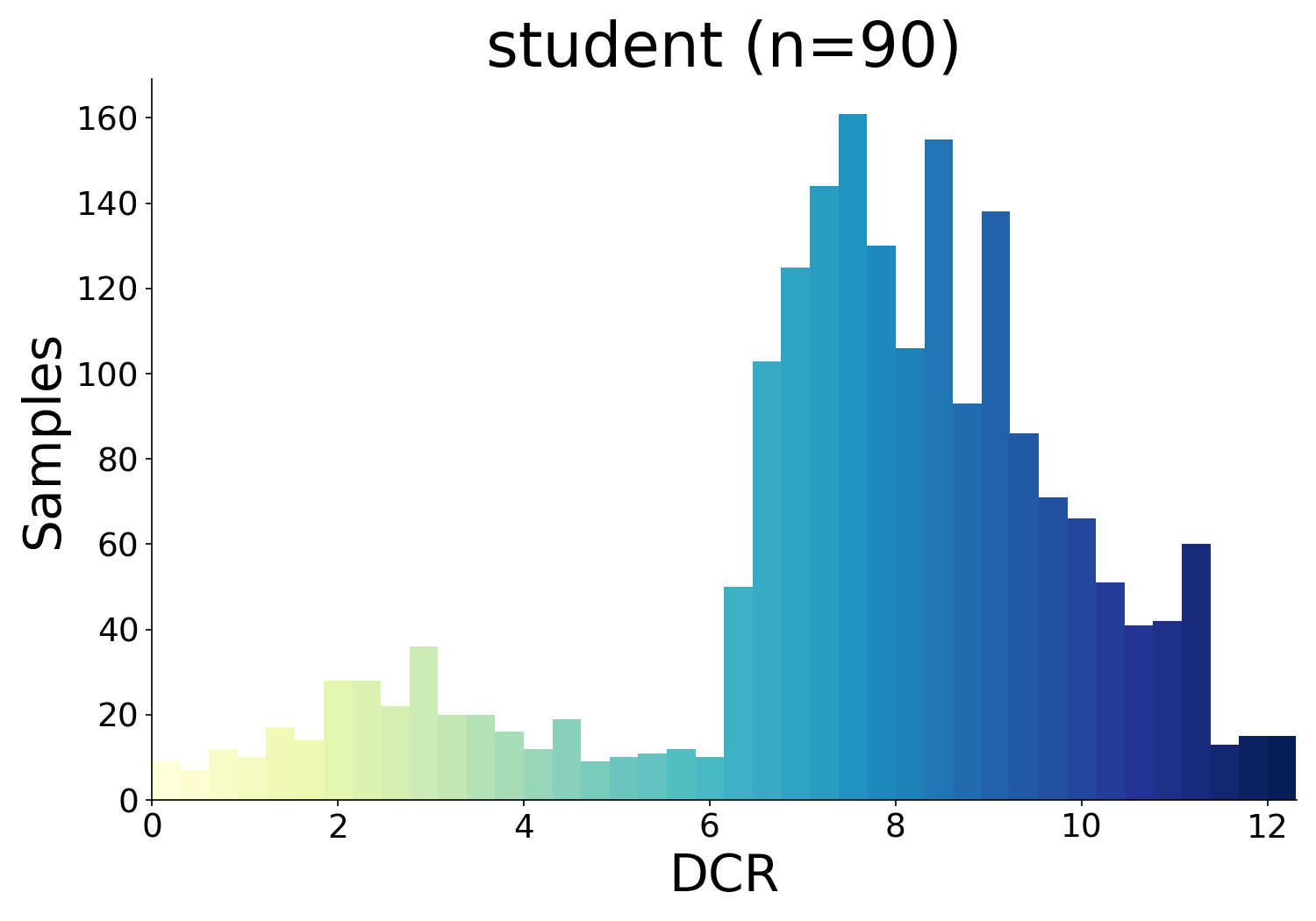}
    \end{subfigure}
    \begin{subfigure}{0.23\textwidth}
        \centering
        \includegraphics[width=\linewidth]{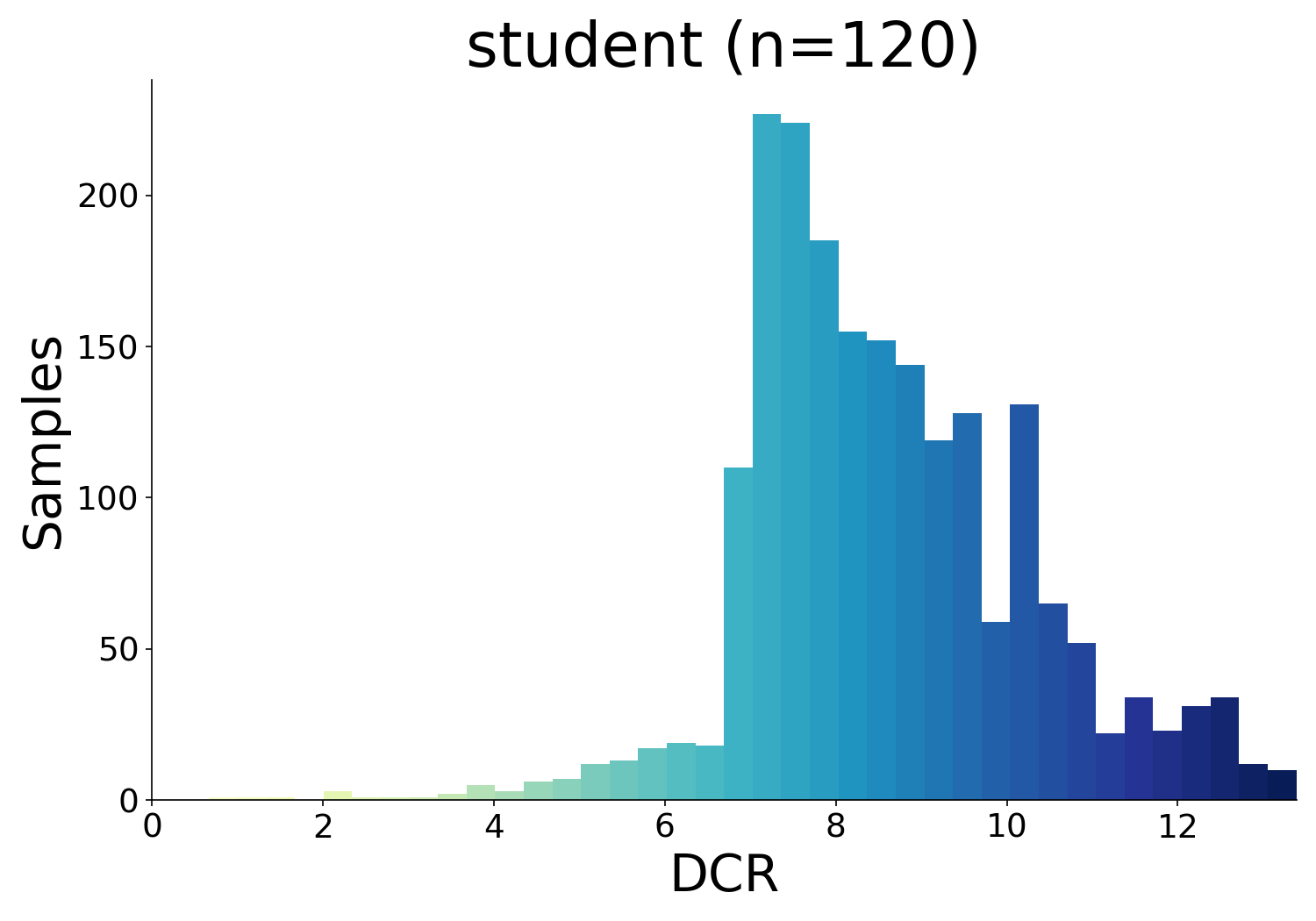}
    \end{subfigure}
    \begin{subfigure}{0.23\textwidth}
        \centering
        \includegraphics[width=\linewidth]{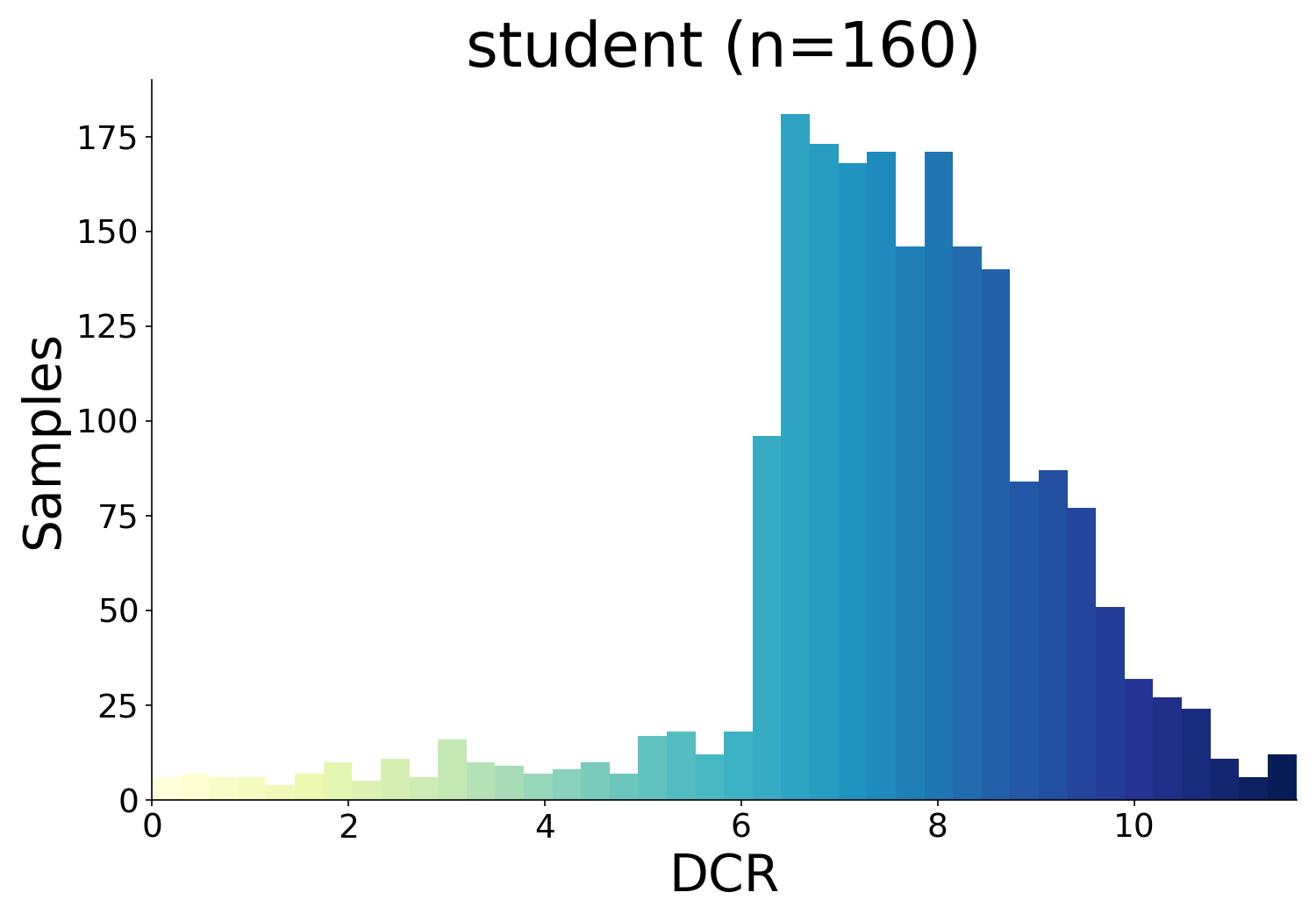}
    \end{subfigure}
    \begin{subfigure}{0.23\textwidth}
        \centering
        \includegraphics[width=\linewidth]{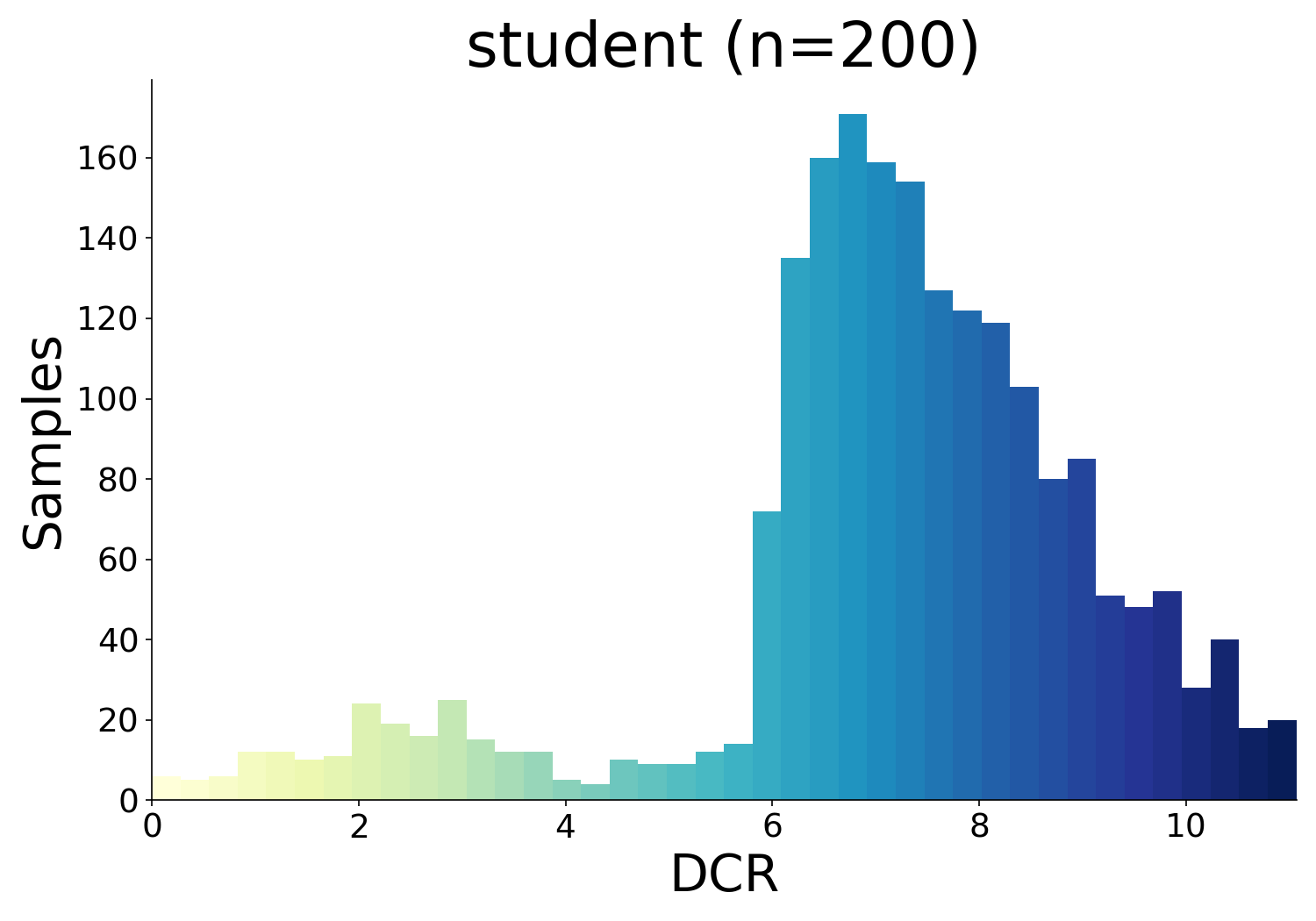}
    \end{subfigure}
    \begin{subfigure}{0.23\textwidth}
        \centering
        \includegraphics[width=\linewidth]{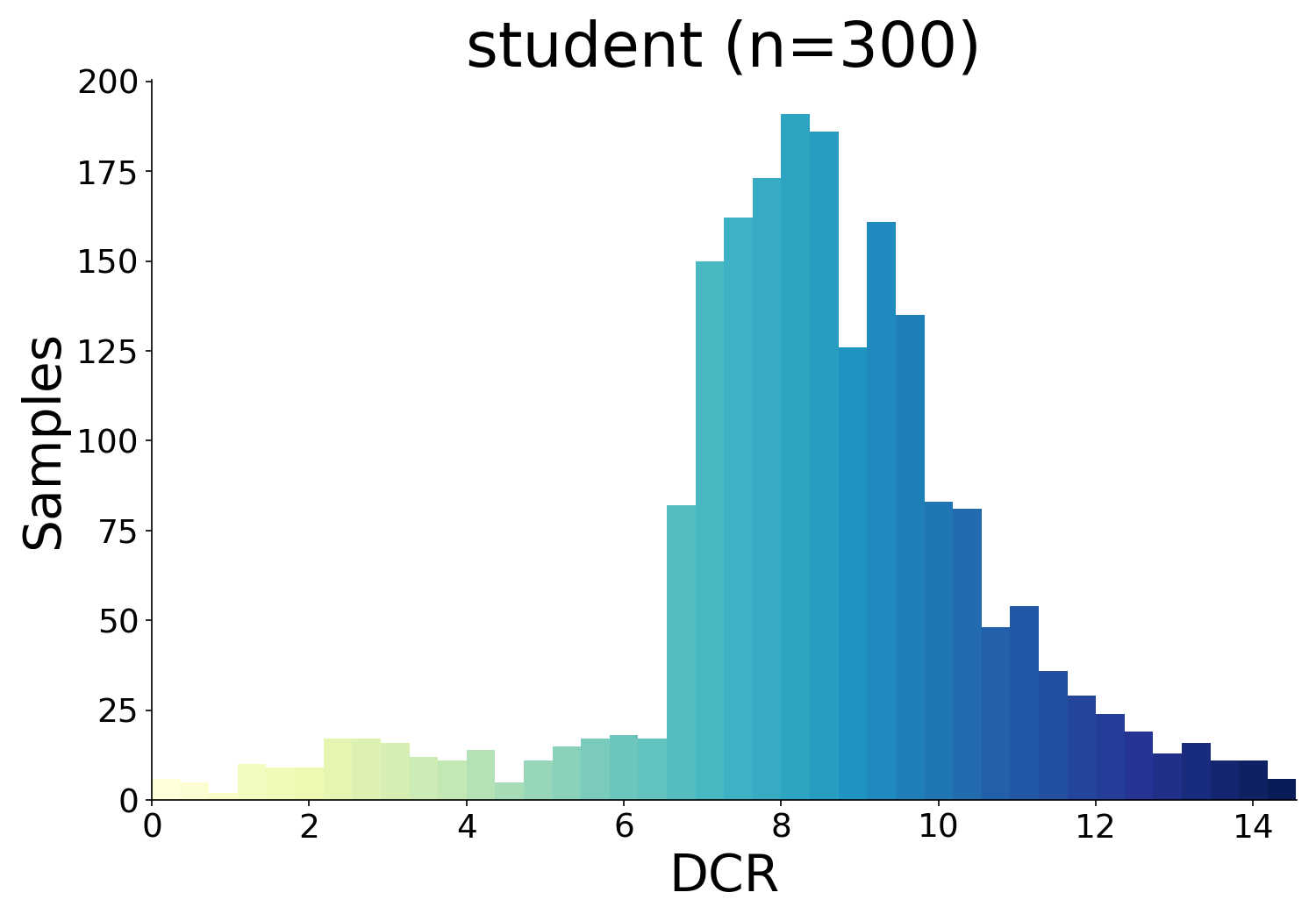}
    \end{subfigure}
    \begin{subfigure}{0.23\textwidth}
        \centering
        \includegraphics[width=\linewidth]{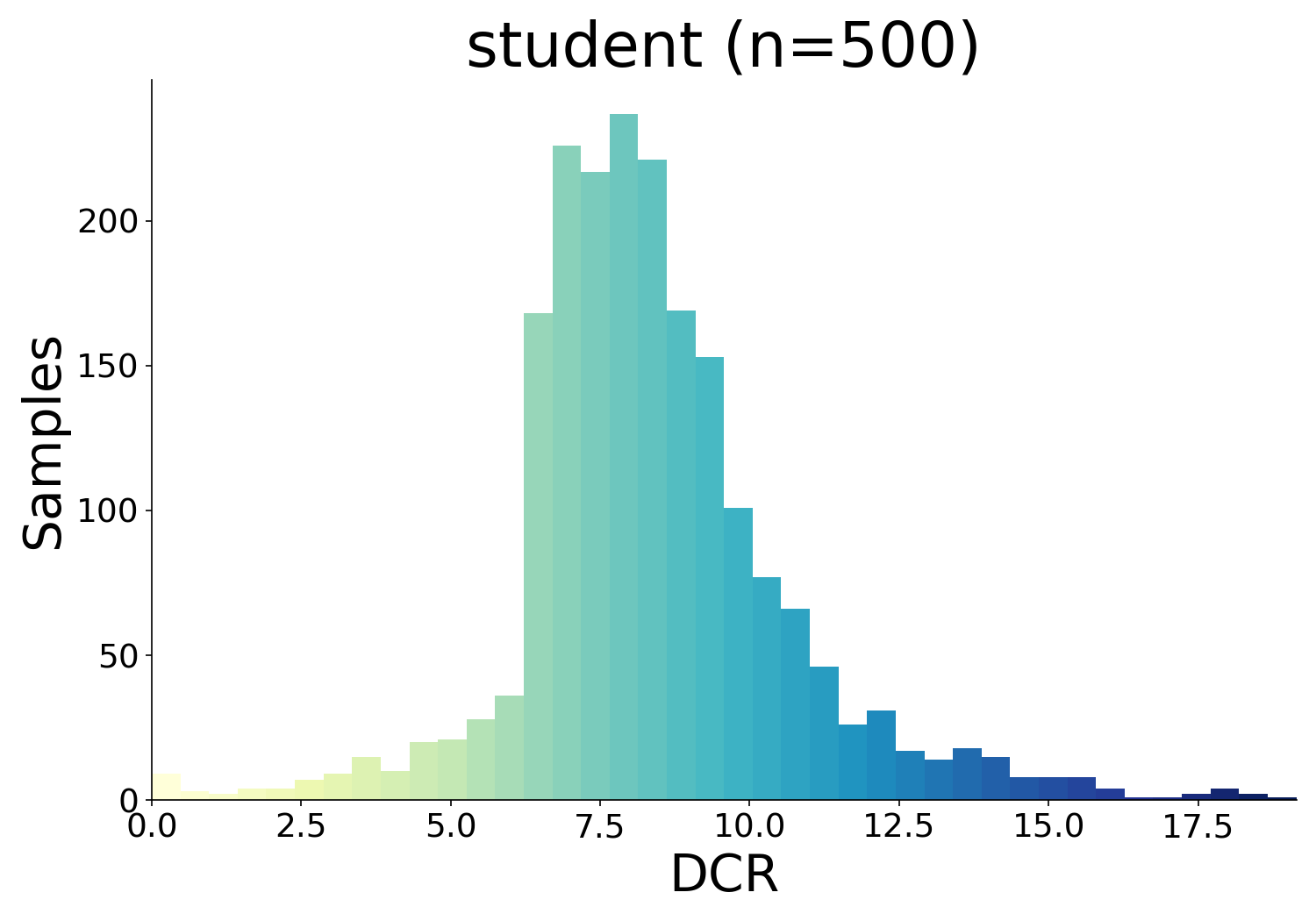}
    \end{subfigure}

    \caption{Distance to closest record (DCR) distributions for the \textsc{Disease} dataset, 
    comparing synthetic data generated by \textbf{ReFine} to the original train set.}
    \label{fig:dcr_student}
\end{figure*}

Figures~\ref{fig:dcr_disease}--\ref{fig:dcr_student} report DCR histograms for four unseen datasets (Disease, Game, GPA, Student) under training sizes 
$N \in \{30, 60, 90, 120, 160, 200, 300, 500\}$.  
Each histogram shows the distance between a generated sample and its closest training instance, computed using \textbf{DCR}.  

Across all datasets and all low-data regimes, two consistent observations emerge:

\paragraph{(i) No evidence of memorization.}
Even at the smallest data setting ($N=30$), the DCR mass remains well separated from zero:  
we do not observe spikes at distance~0 nor anomalous concentration near~0.  
This indicates that ReFine does not reproduce training records verbatim, despite the extreme data scarcity.

\paragraph{(ii) Stable privacy behaviour as $N$ increases.}
As $N$ grows from 30 to~500, the DCR distribution shifts slightly toward smaller values—reflecting the fact that more structure becomes identifiable—but it remains broad and non-degenerate.  
Importantly, the distributions never collapse toward zero, showing that increased data availability does not lead to sample-level leakage.

\medskip
In summary, ReFine maintains strong privacy behaviour across all low-data settings: it improves downstream learning without memorizing or leaking individual training records, even when only 30 examples are available.

\section{Prompt for Component I}
\label{app:prompts}
\begin{figure*}[b]
    \centering
    \includegraphics[width=0.9\linewidth]
    {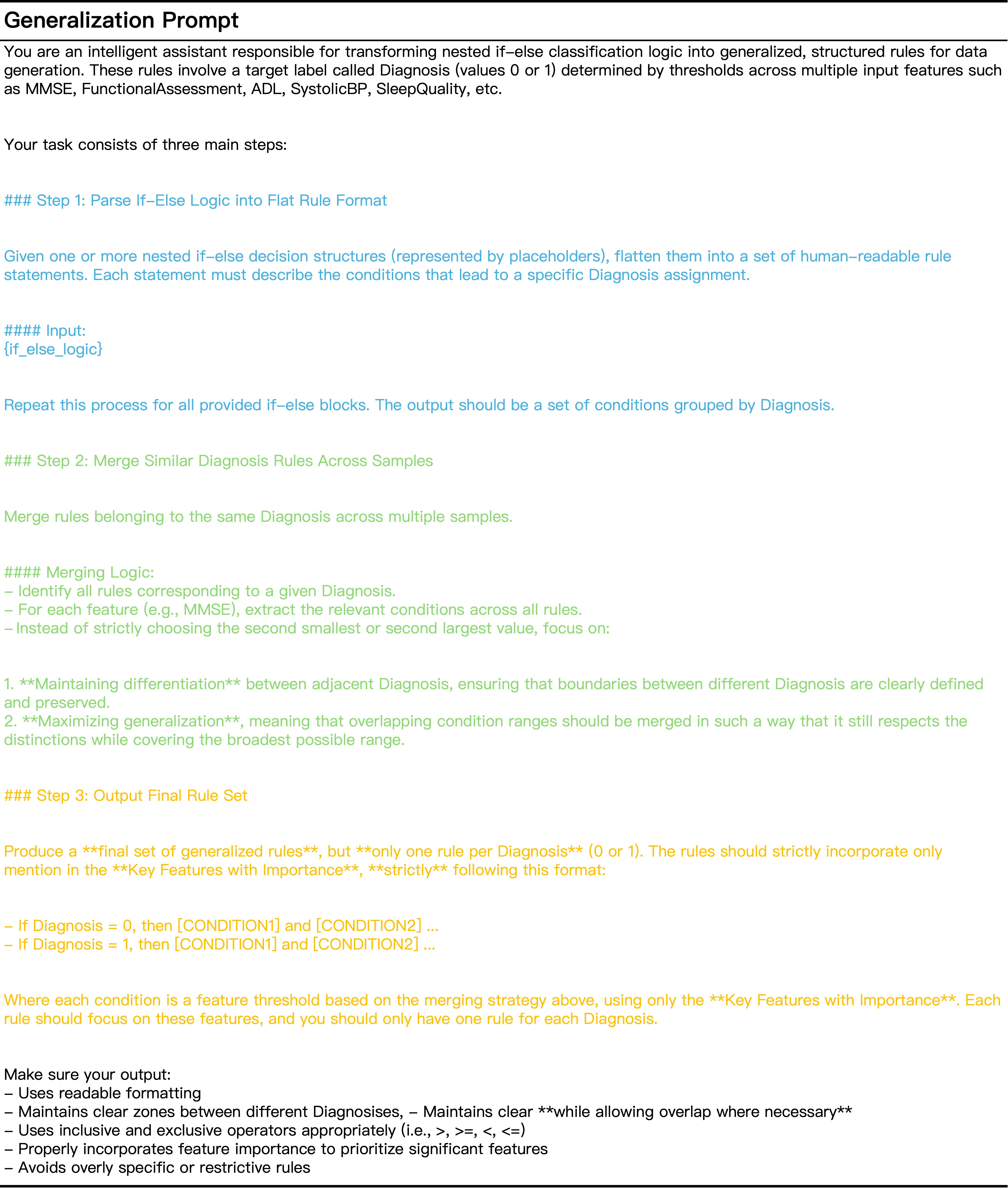}
    \caption{\textbf{Rule Generalization Prompt.} We prune and consolidate decision paths across top-performing trees,retaining only their core (i.e., high-support, low-depth) branches that reflect stable feature–label dependencies. This yields generalized
    \textit{if–then} rules that extend beyond specific training instances.
    Treated as conditional templates with the label as premise, these rules enable inverse reasoning and guide sample generation toward broader yet
    distributionally faithful regions of the data space.}
    \label{fig:rule_generalization_prompt}
\end{figure*}

\begin{figure*}[t]
    \centering
    \includegraphics[width=0.9\linewidth]{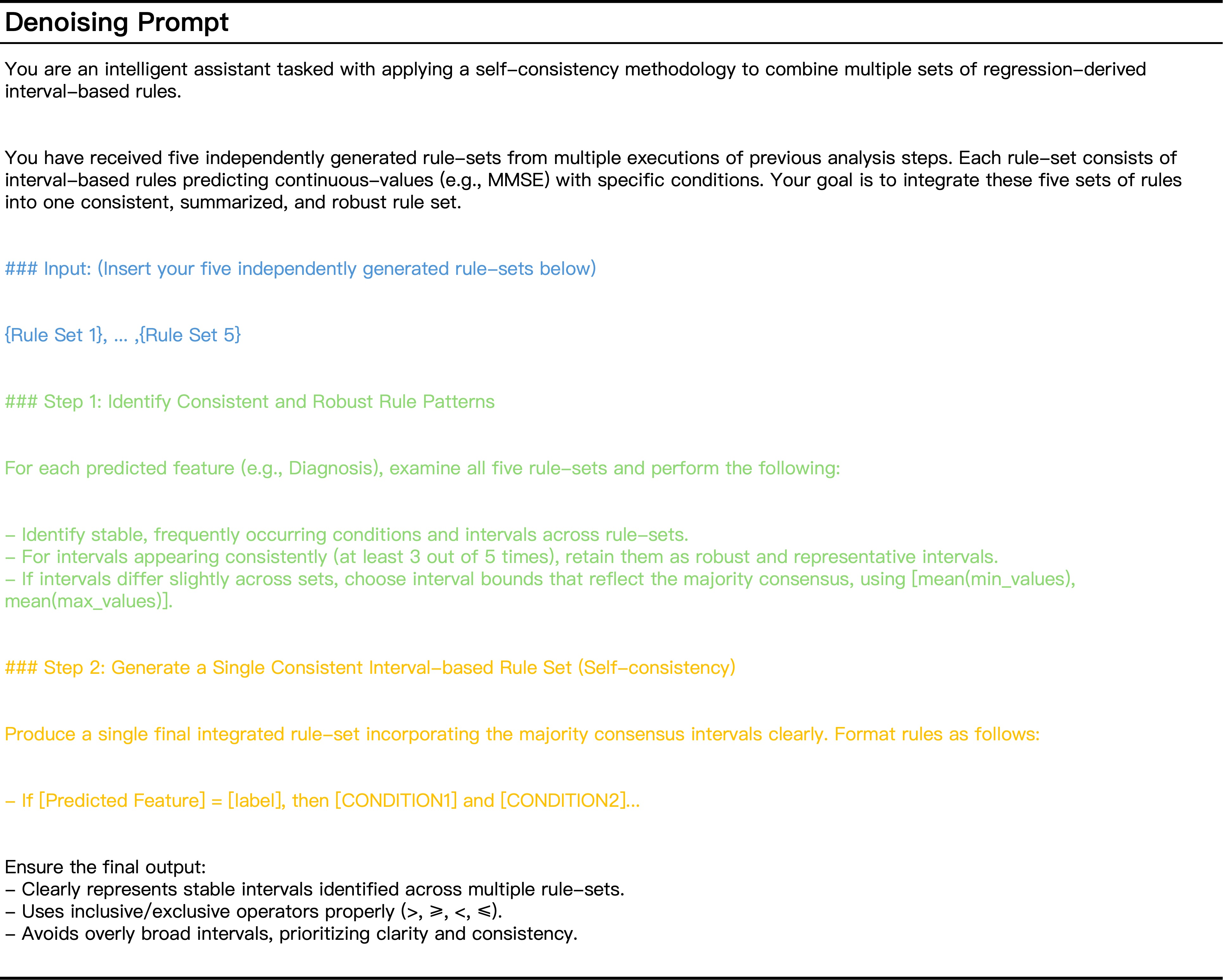}
    \caption[Rule Denoising Prompt.]{
    \textbf{Rule Denoising Prompt.}
    To mitigate variance introduced by symbolic extraction and decoding noise~\protect\cite{he2024advancing}, 
    we adopt self-consistency strategies commonly used in reasoning tasks~\protect\cite{wangself,lewkowycz2022solving}. 
    Multiple generations are sampled using different random seeds, and only the most frequently occurring rules are retained. 
    This procedure stabilizes the extracted rule set and ensures its reliability for downstream generation.
    }
    \label{fig:rule_denoising_prompt}
\end{figure*}

This section provides the prompts used in \textbf{Component I} (\emph{Rules-Guided Generation}).
Given decision paths extracted from top-performing trees, we employ two stages to obtain prompt-feasible and reliable symbolic guidance for generation.
First, we \emph{generalize} the raw decision paths by pruning overly specific branches and consolidating overlapping conditions into compact \textit{if--then} rules (Figure~\ref{fig:rule_generalization_prompt}).
Second, we \emph{denoise} the rule set via self-consistency, retaining only rules that persist across multiple independently sampled generations to reduce extraction and decoding variance (Figure~\ref{fig:rule_denoising_prompt}).

\end{document}